\documentclass[twoside]{article}
\usepackage[accepted]{aistats2026}
\usepackage{marvosym}
\usepackage[T1]{fontenc}
\usepackage{textcomp}

\usepackage{amsmath, amssymb, amsthm, bm, mathtools}
\usepackage{url}
\usepackage{natbib}
\usepackage{graphicx, subfigure}
\usepackage{booktabs, array, multirow, makecell}
\usepackage{algorithm, algorithmic}
\usepackage[T1]{fontenc}
\usepackage{dsfont, mathrsfs}
\usepackage{mathrsfs}
\usepackage{xcolor}
\usepackage[title]{appendix}

\usepackage{enumerate}

\usepackage[colorlinks=true, urlcolor=blue]{hyperref}

\usepackage{tikz}
\usetikzlibrary{arrows.meta, positioning, bayesnet}

\newcount\comments  
\newcommand{\genComment}[2]{\ifnum\comments=1{\textcolor{#1}{\textsf{\footnotesize #2}}}\fi}

\def\TV{\mathop{\text{TV}}}

\long\def\comment#1{}

\def\real{{\mathbb{R}}}
\def\R{{\real}}

\def\##1\#{\begin{align}#1\end{align}}
\def\$#1\${\begin{align*}#1\end{align*}}

\let\cite\citet

\let\hat\widehat
\let\tilde\widetilde

\newcommand{\eb}{\mathbf{e}}

\newcommand{\ob}{\mathbf{o}}
\newcommand{\pb}{\mathbf{p}}

\newcommand{\bb}{\bm{b}}

\newcommand{\bk}{\bm{k}}

\newcommand{\bv}{\bm{v}}

\newcommand{\bx}{\bm{x}}

\newcommand{\bI}{\bm{I}}

\newcommand{\cA}{\mathcal{A}}

\newcommand{\cG}{\mathcal{G}}

\newcommand{\cO}{\mathcal{O}}

\newcommand{\calA}{\mathcal{A}}
\newcommand{\calB}{\mathcal{B}}
\newcommand{\calC}{\mathcal{C}}
\newcommand{\calD}{\mathcal{D}}
\newcommand{\calE}{\mathcal{E}}
\newcommand{\calF}{\mathcal{F}}
\newcommand{\calG}{\mathcal{G}}
\newcommand{\calH}{\mathcal{H}}

\newcommand{\calK}{\mathcal{K}}

\newcommand{\calN}{\mathcal{N}}
\newcommand{\calO}{\mathcal{O}}

\newcommand{\calR}{\mathcal{R}}
\newcommand{\calS}{{\mathcal{S}}}

\newcommand{\EE}{\mathbb{E}}

\newcommand{\PP}{\mathbb{P}}
\newcommand{\QQ}{\mathbb{Q}}

\newcommand{\argmax}{\mathop{\mathrm{argmax}}}

\newcommand{\sign}{\mathop{\mathrm{sign}}}
\newcommand{\tr}{\mathop{\mathrm{tr}}}

\DeclareMathOperator{\Var}{{\rm Var}}

\newtheoremstyle{mytheoremstyle} 
    {\topsep}                    
    {\topsep}                    
    {\normalfont}                   
    {}                           
    {\bfseries}                   
    {.}                          
    {.5em}                       
    {}  

\theoremstyle{mytheoremstyle}

\ifx\BlackBox\undefined
\newcommand{\BlackBox}{\rule{1.5ex}{1.5ex}}  
\fi

\ifx\QED\undefined
\def\QED{~\rule[-1pt]{5pt}{5pt}\par\medskip}
\fi

\ifx\proof\undefined
\newenvironment{proof}{\par\noindent{\bf Proof\ }}{\hfill\BlackBox\\[2mm]}
\fi

\ifx\theorem\undefined
\newtheorem{theorem}{Theorem}[section]
\fi
\ifx\example\undefined
\newtheorem{example}{Example}[section]
\fi
\ifx\property\undefined
\newtheorem{property}{Property}[section]
\fi
\ifx\lemma\undefined
\newtheorem{lemma}{Lemma}[section]
\fi
\ifx\proposition\undefined
\newtheorem{proposition}{Proposition}[section]
\fi
\ifx\remark\undefined
\theoremstyle{plain}
\newtheorem{remark}{Remark}[section]
\fi
\ifx\corollary\undefined
\newtheorem{corollary}{Corollary}[section]
\fi
\ifx\definition\undefined
\newtheorem{definition}{Definition}[section]
\fi
\ifx\conjecture\undefined
\newtheorem{conjecture}{Conjecture}[section]
\fi
\ifx\fact\undefined
\newtheorem{fact}{Fact}[section]
\fi
\ifx\claim\undefined
\newtheorem{claim}{Claim}[section]
\fi
\ifx\assumption\undefined
\newtheorem{assumption}{Assumption}[section]
\fi
\ifx\cond\undefined
\newtheorem{cond}{Condition}[section]
\fi

\newtheoremstyle{anothertheoremstyle} 
    {\topsep}                    
    {\topsep}                    
    {\normalfont}                   
    {}                           
    {\bfseries}                   
    {.}                          
    {.5em}                       
    {\thmname{#1} \thmnumber{#2} \normalfont{#3}}  
\ifx\condition\undefined
\theoremstyle{anothertheoremstyle}
\newtheorem{condition}{Condition}
\fi
\numberwithin{equation}{section}
\numberwithin{theorem}{section}

\makeatletter
\newcommand{\raisemath}[1]{\mathpalette{\raisem@th{#1}}}
\newcommand{\raisem@th}[3]{\raisebox{#1}{$#2#3$}}
\makeatother

\let\oldabstract\abstract
\let\oldendabstract\endabstract
\makeatletter
\renewenvironment{abstract}
{%
               {\list{}{\addtolength{\leftmargin}{-0.5em} 
                       \listparindent 0em%
                        \itemindent    \listparindent%
                        \rightmargin   \leftmargin%
                        \parsep        \z@ \@plus\p@}%
                \item\relax}%
               {\endlist}%
\oldabstract}
{\oldendabstract}
\makeatother

\newcommand{\Astar}{A^{\star}}

\newcommand{\BALL}{\texttt{BALL}}
\newcommand{\regret}{\mathrm{reg}}
\newcommand{\LinUCB}{\texttt{LinUCB}}

\newcommand{\non}{\texttt{non}}
\newcommand{\lin}{\texttt{lin}}
\newcommand{\KL}{\operatorname{KL}}
\newcommand{\phihat}{\hat{\phi}}

\newcommand{\mA}{\boldsymbol{A}}

\newcommand{\mV}{\boldsymbol{V}}
\newcommand{\mX}{\boldsymbol{X}}
\newcommand{\mI}{\boldsymbol{I}}
\newcommand{\mQ}{\boldsymbol{Q}}
\newcommand{\mO}{\boldsymbol{O}}
\newcommand{\mS}{\boldsymbol{S}}

\newcommand{\mW}{\boldsymbol{W}}
\newcommand{\mY}{\boldsymbol{Y}}

\newcommand{\mPhi}{\boldsymbol{\Phi}}
\newcommand{\mSigma}{\boldsymbol{\Sigma}}

\newcommand{\vs}{\boldsymbol{s}}
\newcommand{\vx}{\boldsymbol{x}}
\newcommand{\vy}{\boldsymbol{y}}
\newcommand{\ve}{\boldsymbol{e}}

\newcommand{\vbeta}{\boldsymbol{\beta}}
\newcommand{\vbetahat}{\hat{\boldsymbol{\beta}}}
\newcommand{\vtheta}{\boldsymbol{\theta}}
\newcommand{\vthetahat}{\hat{\boldsymbol{\theta}}}
\newcommand{\vthetastar}{\boldsymbol{\theta}^\star}
\newcommand{\vthetastart}{\boldsymbol{\theta}^{\star\top}}
\newcommand{\vphi}{\boldsymbol{\phi}}
\newcommand{\vphihat}{\hat{\vphi}}

\newcommand{\vbhat}{\hat{\boldsymbol{b}}}
\newcommand{\vomega}{\boldsymbol{\omega}}

\newcommand{\What}{\hat{W}}

\newcommand{\muhat}{\hat{\mu}}

\newcommand{\Regret}{\mathcal{R}}

\usepackage{xcolor}
\usepackage{hyperref}
\usepackage{caption}
\hypersetup{
    colorlinks=true,
    linkcolor=blue,
    filecolor=blue,      
    urlcolor=blue,
    citecolor=blue,
}

\usepackage{caption}
\begin{document}

\newcommand{\YG}[1]{
  {\color{blue} [YG: {#1}]}
 }

\newcommand{\HY}[1]{
  {\color{gray} [HY: {#1}]}
 }

\newcommand{\hz}[1]{
  {\color{violet} [HZ: {#1}]}  
}
 
\twocolumn[
\aistatstitle{Learning with Incomplete Context: Linear Contextual Bandits with Pretrained Imputation}

\aistatsauthor{
  Hao Yan\textsuperscript{*} \hspace{3em}
  Heyan Zhang\textsuperscript{*} \hspace{3em}
  Yongyi Guo\textsuperscript{\Letter}
}
\runningauthor{Hao Yan, Heyan Zhang, Yongyi Guo}

\aistatsaddress{
  Department of Statistics,
  University of Wisconsin--Madison \\
}
]

\begingroup
  \renewcommand\thefootnote{}
  \renewcommand{\footnoterule}{} 
  \footnote{\scriptsize \textsuperscript{*} Equal contribution, listed in alphabetical order.}
  \footnote{\scriptsize \textsuperscript{\Letter} Corresponding author: \texttt{guo98@wisc.edu.}}
  \addtocounter{footnote}{-1}
\endgroup

\begin{abstract}
The rise of large-scale pretrained models has made it feasible to generate predictive or synthetic features at low cost, raising the question of how to incorporate such surrogate predictions into downstream decision-making. We study this problem in the setting of online linear contextual bandits, where contexts may be complex, nonstationary, and only partially observed. In addition to bandit data, we assume access to an auxiliary dataset containing fully observed contexts---common in practice since such data are collected without adaptive interventions. We propose PULSE-UCB, an algorithm that leverages pretrained models trained on the auxiliary data to impute missing features during online decision-making. We establish regret guarantees that decompose into a standard bandit term plus an additional component reflecting pretrained model quality. In the i.i.d.~context case with Hölder-smooth missing features, PULSE-UCB achieves near-optimal performance, supported by matching lower bounds. Our results quantify how uncertainty in predicted contexts affects decision quality and how much historical data is needed to improve downstream learning.
\end{abstract}

\section{INTRODUCTION}
\label{sec:intro}

Contextual bandits provide a powerful framework for sequential decision-making under uncertainty, where the learner repeatedly observes a context, chooses an action, and receives a reward. The key challenge is to balance exploration and exploitation while adapting decisions to the observed context. Owing to their simplicity and flexibility, contextual bandits have been widely applied in practice, including personalized recommendations \citep{li2010contextual}, mobile health \citep{nahum2016just}, and online education platforms \citep{cai2021bandit}. 

In many practical applications, the contexts required for decision-making may be missing or only partially observed during online interactions. For example, in the HeartSteps mobile health study \citep{liao2020personalized}, the full physiological state of a participant is unobserved, while only partial signals such as step counts, activity levels, or self-reports from wearables are available to guide intervention delivery. Similarly, in online education platforms \citep{lan2016contextual}, a learner’s complete knowledge state across multiple concepts is latent, and the system only observes partial signals such as responses to quiz items or practice problems. At the same time, large offline datasets with substantially more complete contexts are often accessible, since they can be collected without interventions or adaptive decision-making. Such datasets have been shown to reveal richer contextual information than what is available in online interaction \citep{kausik2025leveraging}, raising the question of how these auxiliary resources can be effectively leveraged to improve sequential decision-making when online contexts are missing.

In this work, we address the problem of linear contextual bandits when contexts are only partially observed during online interaction, while offline auxiliary data provide full context information. The key idea is to use predictive models trained on auxiliary data to impute the missing contexts for online decisions. Even with access to auxiliary data, it is often reasonable in practice to combine pretrained imputations with simple policies such as linear bandits, since they yield stable and interpretable rules, and enable valid post-hoc statistical inference, which are crucial in applications such as healthcare and education \citep{rafferty2019statistical, zhang2024replicable, guo2025statistical}. Fundamental questions arise: how can predictive models trained on auxiliary data improve such decision rules, and how does imputation quality affect regret?


\textbf{Our contributions.} We propose \texttt{PULSE-UCB}, an online algorithm that uses auxiliary data to impute missing contexts and guide decision-making in linear contextual bandits. Under general context sequence distributions, we establish a regret bound of $\tilde\cO(dT^{1/2} + \delta_0d^{3/2}T)$, where where $T$ is the time horizon, $d$ is the dimension of the full context, and $\delta_0$ captures the quality of the predictive model learned from auxiliary data. In the special case of i.i.d. contexts with $\beta$-Hölder smooth missing features, we further show that $\delta_0 \lesssim N^{-{\beta}/{(2\beta + d_S)}}$, where $N$ is the auxiliary sample size and $d_S$ is the dimension of the observed contexts, and we complement this with a matching lower bound, establishing near-optimality in both the time horizon and the auxiliary data size.

\subsection{Related works}

\textbf{Bandits with partially observed contexts.}
Given its importance, a substantial literature has studied contextual bandits with partially observed contexts. Many works impose parametric assumptions on the full context, such as i.i.d. Gaussian contexts or linear dynamical systems with additive Gaussian noise \citep{kim2023contextual, park2022regret, park2024thompson, zeng2025partially, xu2021generalized}. In contrast, we do not impose parametric assumptions on the context distribution, which
can be misspecified in complex applications. Other works allow more general distributions but with restrictions, such as fixed, time-invariant contexts \citep{kim2025linear}, or contexts missing completely at random (MAR) \citep{jang2022high}. Compared to these works, we allow the context process to be dependent, nonstationary, and missing not at random (MNAR). 
A closely related work is \citet{hu2025pre}, which considers nonlinear bandits with i.i.d. partially observed contexts (including both MAR and MNAR scenarios) and leverages pretrained models with orthogonal statistical learning to derive regret upper bounds. Compared to \citet{hu2025pre}, we focus on linear contextual bandits with potentially dependent and nonstationary contexts and provide both upper and matching lower bounds, establishing near-optimal regret in this setting.
Finally, another related line of work analyzes corrupted contexts and benchmarks against a mixture of contextual and multi-armed bandits \citep{bouneffouf2020online}, whereas our auxiliary data enable comparison to the stronger benchmark of the optimal full-context policy.


\textbf{Connections to broader areas.}
Our work also relates to AI-assisted decision-making, where pretrained models support online policies \citep{tianhui2024active, zhang2025contextual, chen2021decision, janner2021offline, lin2023transformers, lee2023supervised, ye2025lola, cao2024hr}, and to the broad literature on imputation-based methods in statistics and machine learning, from the classical EM algorithm \citep{dempster1977maximum} to modern ML-based approaches \citep{xia2024prediction, angelopoulos2023prediction}. We differ by focusing specifically on the missing-context issue in online bandits, providing regret bounds that guide the principled use of pretrained imputation for sequential decisions. A more complete literature review is deferred to the Appendix~\ref{litrev}.

\paragraph{Notation}
For a positive integer $n$, we write $[n] = \{1,2,\dots,n\}$.
For a vector $\bv=\left(v_1,v_2, \dots, v_n\right)^\top \in \R^n$, $\|\bv\|_2=\sqrt{\sum_{i=1}^n v_i^2}$ and $\|\bv\|_{\infty}=\max_i\left|v_i\right|$.
$\bI_{n} \in \R^{n\times n}$ denotes the $n$-by-$n$ identity matrix.
For positive functions $f(n)$ and $g(n)$, we write $f(n) \gtrsim g(n)$, $f(n) = \Omega(g(n))$ or $g(n) = \cO(f(n))$ if for some constant $C > 0$, we have $f(n)/g(n) \ge C$ for all sufficiently large $n$.
We write $f(n) = \tilde{\cO}(g(n))$ if $f(n) = \cO(g(n)\mathrm{polylog}(n))$, that is, there exist constants $C,k>0$ such that $f(n) \le C g(n) (\log n)^k$ for all sufficiently large $n$.

\section{PROBLEM SETUP}
\label{sec:setup}
We consider a sequential decision-making process in contextual bandits with partially observed contexts. Given a time horizon $T$, for each $t = 1, 2, \dots, T$:
\begin{itemize}  
    \item[(a)] \textbf{Context generation.} A latent context $\mY_t\in \mathbb R^{d_Y}$ is generated from an unknown probability distribution $p_\star(\cdot \mid \mY_{1:t-1})$. 
    At each time step $t$, the agent observes a partial context $\mS_t \in \mathbb{R}^{d_{S,t}}$. Note that we use the subscript $t$ in the dimension $d_{S,t}$ to explicitly indicate that the dimension of the observed feature is allowed to vary over time.
    
    \item[(b)] \textbf{Action and reward.} Based on the observed history, the agent selects an action  $A_t \!\in \!\cA$ and receives a reward  $R_t = R(t, A_t)$. Here we define the potential reward as
        \begin{equation} \label{eq:Rta:define}
            R(t, a)\!:=\!\langle\vthetastar\!, \mPhi(\mY_t, a)\rangle \!+ \!\eta_t ,\quad \forall t \!\in\! [T], a\!\in\!\mathcal A,
        \end{equation}
    where $\vthetastar \in \R^d$ is an unknown parameter, $\mPhi$ is a known feature mapping, and $\eta_t$ is mean-zero condition on past history.
    We assume that 
        \begin{equation}\label{eq:Rta:bounded}
            R(t,a) \in [-1,1].
        \end{equation}
\end{itemize}  

    
The feature map $\mPhi$ satisfies the following assumption: 
\begin{assumption}\label{assump:phi:bound}
For any $a \in \cA$, there exists $B > 0$
$$
    \sup_{\vy \in \R^{d_Y}} \left\|\mPhi(\vy, a)\right\|_{\infty} \leq 1, \quad \sup_{\vy \in \R^{d_Y}} \left\|\mPhi(\vy, a)\right\|_{2} \leq B. 
$$
\end{assumption}
In addition, we impose a standard assumption on the noise sequence $\{\eta_t\}_{t=1}^T$.
\begin{assumption}\label{assump:sub-G}
    Suppose that $\{\eta_t\}_{t=1}^T$ is a $\sigma^2_{\eta}$-sub-Gaussian martingale difference sequence with respect to $\{\calF_t\}_{t=1}^T$. 
    Here
\begin{equation} \label{eq:calF_t:define}
    \calF_t := \sigma\left(\mY_{1:t}, A_{1:t-1}, R_{1:t-1}\right),
\end{equation}
where $\sigma(\cdot)$ denotes the generated $\sigma$-algebra.
\end{assumption}


Note that in this bandit setting, both the full context $\mY_t$ and its partial observation $\mS_t$ are assumed to be exogenous and do not depend on the action sequence $(A_1, \dots, A_t)$. This asymmetric context availability, where offline auxiliary data provides richer, higher-fidelity information than what can be feasibly observed during real-time online decision-making, is strongly motivated by practical constraints and naturally arises in many real-world applications. For instance, in digital health interventions, the full state of a patient $\mY_t$ may include comprehensive physiological and psychological factors such as stress level and sleep quality. However, because online systems often face strict privacy regulations, local computational limits, or frequent sensor failures, only a partial subset $\mS_t$ (such as step counts or heart rate) is observed in real-time. Similarly, in online education platforms, a learner's true knowledge state across multiple concepts ($\mY_t$) is unobservable during real-time interactions, and the system only receives partial signals like answers to specific quiz items or homework questions ($\mS_t$). Fortunately, archival offline data in these domains can safely store the comprehensive state variables ($\mY_t$) to aid online learning. We provide a detailed discussion of these practical motivations in Appendix~\ref{app:motivation}.

Our goal is to sequentially select actions $\{A_t\}_{t=1}^T$, where each $A_t$ is chosen based only on the observed history $\{(\mS_\tau, A_\tau, R_\tau)\}_{\tau=1}^{t-1}$ and the current observation $\mS_t$, so as to maximize the cumulative reward. This is equivalent to minimizing the cumulative regret  
\begin{equation*}
    \sum_{t=1}^T \EE \left[R(t,\Astar_t) - R(t,A_t)\right],
\end{equation*}
where $\Astar_t$ is the optimal action that maximizes the expected reward, assuming the full context $\mY_t$ is observed:
\begin{equation} \label{eq:Astart:def}
    \Astar_t := \arg\max_{a \in \calA} \left\langle \vthetastar, \mPhi(\mY_t, a)\right\rangle.
\end{equation}

The key challenge is that the latent contexts are only observed indirectly through the partial information $\mS_{1:T}$. In general, good decision-making is impossible without adequate knowledge of the underlying contexts. In practice, however, it is often possible to obtain auxiliary historical data from related populations that include both partial observations and richer measurements of the underlying state. In the digital health example, historical studies often collect both wearable sensor streams and survey or clinical assessments. In online education, large-scale platforms frequently link fine-grained interaction logs (e.g., quiz responses, practice problems) with standardized test scores or comprehensive assessments, providing aligned data on both partial signals and richer proxies of the true knowledge state. Motivated by these settings, we assume access to an auxiliary dataset $\calD$ consisting of i.i.d.~trajectories
\begin{equation*}  
    \calD = \left\{ \left(\mY_{i, 1:T_0}^{(0)}, \mS_{i, 1:T_0}^{(0)}\right) : i = 1, \dots, N \right\},
\end{equation*}  
where $T_0 \geq 1$ denotes the time horizon of the historical data, and each trajectory $(\mY_{i,1:T_0}^{(0)}, \mS_{i,1:T_0}^{(0)})$ is drawn from the same joint distribution as the bandit contexts $(\mY_{1:T_0}, \mS_{1:T_0})$. This dataset is assumed to reasonably capture the joint distribution of $\mY_{1:T}$ and $\mS_{1:T}$. For instance, if the dependence structure between $\mY_{1:T}$ and $\mS_{1:T}$ is complex, one may require $T_0$ to be of the same order as the bandit horizon $T$ in order to accurately recover this relation. In general, however, $T_0$ is flexible and need not be greater than $T$, and most of our results impose no explicit relation between them.


\section{THE PULSE-UCB ALGORITHM}
\label{sec:method}

Under the setting introduced above, we propose  \textit{Pretrained Unobserved Latent State Estimation UCB} (\texttt{PULSE-UCB}), an algorithm that leverages auxiliary data to “fill in the blanks” of the missing contexts before making decisions. The main idea is as follows. We first pretrain a model $\hat p$ on $\mathcal{D}$ that learns to predict the full context $\mY_t$ from the observed sequence $\mS_{1:t}$\footnote{Here $\hat p$ can be any pretrained model that provides a conditional distribution of $\mY_t$ given $\mS_{1:t}$. If $\hat p$ provides a deterministic prediction, one can convert it into a probabilistic model by viewing it as the mean of a suitably chosen distribution. }. Then, during online interaction, whenever we only see the partial context $\mS_{1:t}$, we use $\hat p$ to impute the missing parts and obtain complete feature vectors $\vphihat_{t,a}$ for each action. With these surrogate features in hand, the problem reduces to a standard linear contextual bandit, and we apply OFUL \citep{abbasi2011improved}: the algorithm maintains a confidence set for the unknown parameter $\vtheta^\star$, chooses the action that maximizes an optimistic reward estimate, observes the payoff, and updates its estimates accordingly. In this way, the pretrained model provides the missing information, while OFUL handles the exploration-exploitation trade-off. 

To formalize the imputation step, at each time $t$, for any action $a \in \mathcal{A}$, we define the imputed features as the conditional expectation of $\mPhi(\mY_t, a)$ under the pretrained model $\hat p$:
\begin{equation} \label{eq:phihat:define}
    \vphihat_{t,a} := \EE_{\hat{p}} \left[\mPhi(\mY_t, a) \mid \mS_{1:t}\right], \quad \text{for all } t \in [T].  
\end{equation}
In practice, this conditional expectation may not admit a closed-form expression. However, a natural approximation is to draw samples $\vy^{(b)} \sim \hat p(\cdot \mid \mS_{1:t})$ for $b \in [B]$ and compute the Monte Carlo average
\begin{equation*}
   \vphihat_{t,a} \approx \frac{1}{B} \sum_{b=1}^B \mPhi\left(\vy^{(b)}, a\right). 
\end{equation*}
Such approximation can be made arbitrarily accurate, given sufficient computational resources, and we therefore assume direct access to $\vphihat_{t,a}$ in later analysis.  

A full description is given in Algorithm~\ref{alg:linucb}.

\begin{algorithm}[htbp]
\begin{algorithmic}[1]
\REQUIRE Pretrained distribution $\hat{p}$, tuning parameters $\lambda, \{\gamma_t\}_{t=1}^T$.
\STATE Initialize $\mSigma_0 = \lambda \mI$, $\BALL_0 \gets \left\{ \vtheta \mid \!\lambda \|\vtheta\|^2_2 \leq \gamma_0\right\}$.
\FOR{$t = 1$ to $T$}
\STATE Observe context $\mS_t$, compute $\vphihat_{t,a}$ according to Equation~\eqref{eq:phihat:define}. 
\STATE Choose action 
\begin{equation} \label{eq:At:define}
    A_t=\operatorname{argmax}_{a \in \cA} \max _{\vtheta \in \BALL_{t-1}} \vtheta^\top \vphihat_{t,a},
\end{equation}
with ties broken arbitrarily.
\STATE Receive payoff $R_t$.
\STATE Update
\vspace{-3mm}
\begin{equation}\label{eq:Sigt:define}
\mSigma_t \gets \lambda \mI + \sum_{\tau=1}^{t} \vphihat_{\tau, A_\tau}  \vphihat_{\tau, A_\tau}^\top,
\end{equation}
\vspace{-3mm}
\begin{equation}\label{eq:thetahat:define}
    \vthetahat_t \gets \mSigma_t^{-1} \sum_{\tau=1}^{t} R_{\tau} \vphihat_{\tau, A_\tau}.
\end{equation}
\begin{equation}\label{eq:ball:define}
    \BALL_t \!\gets \!\left\{ \vtheta \mid \left(\vthetahat_t - \vtheta\right)^\top \!\mSigma_t \left(\vthetahat_t - \vtheta\right) \leq \gamma_t\right\}.
\end{equation}
\ENDFOR
\end{algorithmic}
\caption{\texttt{PULSE-UCB}}
\label{alg:linucb}
\end{algorithm}

\section{REGRET ANALYSIS}
\label{sec:main-result}

In this section, we analyze the regret of Algorithm~\ref{alg:linucb}. Section~\ref{sec::theory-imputation-err} characterizes the imputation error of the context from the pretrained model, which serves as a key ingredient in the analysis. Section~\ref{sec::regret-general} then establishes a general regret bound under arbitrary context distributions, and Section~\ref{subsec:nonparam:ex} specializes the result to some specific distributional settings.

\subsection{Characterizing imputation error}
\label{sec::theory-imputation-err}

Our first step in the regret analysis is to quantify the quality of the imputed contexts---that is, how far the predicted $\mY_t$ can deviate from the true $\mY_t$ given the current partial context $\mS_{1:t}$. We capture this discrepancy through how well the pretrained model $\hat p$ approximates the ground-truth distribution. Formally, let $\hat\PP$ denote the distributions of $(\mY_{1:T}, \mS_{1:T})$ under $\hat{p}$. For any $t \in [T]$, we measure the divergence between $\hat{\PP}$ and the ground truth $\PP$ by
\begin{equation} \label{eq:Dt:def0}
    D_t \!=\! \tfrac{1}{2} \!\KL\!\left(
\PP\!\left(\mY_{t} \!\!\mid\! \mS_{1:t} \!=\! \vs_{1:t}\right)\!\|
\hat\PP\!\left(\mY_{t} \!\!\mid\! \mS_{1:t} \!=\! \vs_{1:t}\right)
\right).
\end{equation}

The next lemma establishes the theoretical basis that a small $D_t$ ensures the imputed contexts remain close to the true contexts, enabling reliable downstream decision-making. The proof is in the Appendix~\ref{lem1}.

\begin{lemma} \label{lem:conditional-kld}
For any time step $t \in [T]$ and any measurable scalar function $g: \R^{d_Y} \to \R$ with $\|g\|_\infty \leq 1$, 
\begin{equation} \label{eq:KLDbound}
     \EE\!\left[g(\mY_{t}) \!\!\mid\! \mS_{1:t} \!=\! \vs_{1:t}\right] 
          \!-\! \EE_{\hat{p}}\!\left[g(\mY_{t}) \!\!\mid\! \mS_{1:t} \!=\! \vs_{1:t}\right] \!\leq \!\!\sqrt{D_t}. 
\end{equation}
Here $\EE[\cdot]$ denotes the expectations with respect to $\PP$.
\end{lemma}
Lemma~\ref{lem:conditional-kld} can be applied to obtain an error bound for the imputed contexts used in bandit decisions.
Specifically, considering Equation~\eqref{eq:KLDbound} and Assumption~\ref{assump:phi:bound}, for any action $a \in \calA$ we have 
\begin{equation} \label{eq:Dt:def}
\begin{aligned}
        \left\|\EE \left[\mPhi(\mY_t, a) \mid \mS_{1:t}\right] - \vphihat_{t,a}\right\|_{\infty} &\leq 
        \sqrt{D_t}.
\end{aligned} 
\end{equation}
\begin{remark}
The choice of KL divergence in defining $D_t$ in (\ref{eq:Dt:def0}) is for convenience. In fact, the proof bounds the left-hand side of~\eqref{eq:KLDbound} via total variation and Pinsker's inequality. Thus, any $f$-divergence with a Pinsker-type bound (e.g., Hellinger distance) yields an analogous result. If the left-hand side of~\eqref{eq:KLDbound} can be controlled directly, no divergence measure is needed at all. 
\end{remark}

\subsection{Regret analysis under general context distributions}\label{sec::regret-general}
To establish the regret bound of Algorithm \ref{alg:linucb}, we begin by decomposing the reward at round $t$. Specifically,
\begin{align}
R_t & = \vthetastart \mPhi(\mY_t, A_t) + \eta_t\nonumber\\
& = \vthetastart\EE\big[\mPhi(\mY_t, A_t)\mid\mS_{1:t}, A_t\big] + \varepsilon_t + \eta_t,\label{eq::reward-decomposition}
\end{align}
where in the first term, 
$\EE\big[\mPhi(\mY_t, A_t)\mid\mS_{1:t}, A_t\big]$
can be viewed as a new effective context---the part we could recover if the underlying distribution $\mathbb P$ were known. The second term,
\begin{equation} \label{eq:vareps:define}
    \varepsilon_t := \vthetastart \!\left(\mPhi(\mY_t, A_t) - \EE \left[\mPhi(\mY_t,A_t) \!\mid\! \mS_{1:t}, A_t\right]\right),
\end{equation}
captures the error introduced by the unobserved portion of the true context $\mY_t$. Intuitively, if $\varepsilon_t$ has mean zero conditioned on the history, then $\varepsilon_t + \eta_t$ forms a martingale difference sequence. This structure allows us to invoke martingale self-normalized concentration techniques to analyze the regret of the resulting linear contextual bandit if $\mathbb P$ is known. To ensure that the error term $\varepsilon_t$ can be properly controlled, we impose the following assumption. 
\begin{assumption}\label{assump:martingale:difference}
    For all $a \in \calA$ and $t \in [T]$, 
    \begin{equation*}
        \EE \!\left[\mPhi(\mY_t,a) \mid \mY_{1:t-1}, \eta_{1:t-1}, \mS_{1:t}\right] 
        = \EE \!\left[\mPhi(\mY_t,a) \mid \mS_{1:t}\right].
    \end{equation*}
\end{assumption}

Assumption~\ref{assump:martingale:difference} is quite general. It holds whenever $\mY_t$ is conditionally independent of $(\mY_{1:t-1}, \eta_{1:t-1})$ given $\mS_{1:t}$, i.e., whenever $\mS_{1:t}$ provides sufficient information for predicting $\mY_t$. 
In particular, it is satisfied whenever $\mY_t$ is generated as an arbitrary function of $\mS_{1:t}$ together with independent randomness. 
This already covers a broader class of models than many existing works on stochastic contextual bandits, which typically assume i.i.d.~contexts \citep{li2021regret, kim2023contextual, hu2025pre}. Informally, the key requirement is that the new randomness injected at time $t$ does not depend on past randomness; whenever this holds, our algorithm and analysis apply.
\begin{remark}
An exception arises when $\mY_t$ depends on $\mS_{1:t}$ but is driven by autoregressive (AR) rather than independent noise. In such cases, latent variables exhibit temporal dependence that propagates into the rewards, and different techniques are required. Moreover, different forms of AR structure yield fundamentally different learning dynamics: action-driven AR models \citep{bacchiocchi2024autoregressive} admit modified UCB algorithms with $\cO(\sqrt{T})$ regret, whereas latent AR models \citep{trella2024non} require Kalman-filter-type estimators and may incur regret exceeding $\cO(\sqrt{T})$. Even without Assumption~\ref{assump:martingale:difference}, however, $\varepsilon_t$ can sometimes be controlled by alternative means: for example, if $\mW_t$ is generated by a stationary process with geometrically decaying dependence (e.g., a stationary AR process), then concentration inequalities for mixing sequences may be applied, though such analysis would typically require additional structural assumptions on the feature map $\mPhi$. These distinctions underscore that extending beyond Assumption~\ref{assump:martingale:difference} to AR settings generally requires case-by-case, model-specific treatment; a detailed discussion is provided in Appendix~\ref{app:ar-discussion}.
\end{remark}


\begin{lemma}\label{lem:eps:mds}
    Under Assumptions~\ref{assump:sub-G} and~\ref{assump:martingale:difference}, the random variables $\{\varepsilon_t\}_{t=1}^T$ defined in Equation~\eqref{eq:vareps:define} is a martingale difference sequence with respect to the filtration $\{\calG_t\}_{t=1}^T$, which is given by 
    \begin{equation*} 
        \cG_{t-1} := \sigma\left(\mS_{1:t}, \mY_{1:t-1}, \eta_{1:t-1}, U_{1:t}\right) 
    \end{equation*}
    where $U_{1:t}$ are independent auxiliary random variables in selecting $A_{1:t}$ under randomized algorithm. 
\end{lemma}
As a proof sketch, the main goal is to show that 
\begin{equation*}
    \EE \left[\mPhi(\mY_t, A_t) \mid \calG_{t-1}\right] = \EE \left[\mPhi(\mY_t, A_t) \mid \mS_{1:t}, A_t\right].
\end{equation*}
Ignoring the auxiliary randomness $U_{1:t}$ and considering a simplified setting where $(A_{1:t-1}, R_{1:t-1})$, the action and reward prior to round $t$, can be expressed via $(\mY_{1:t-1}, \eta_{1:t-1}, \mS_{1:t})$. The term $\EE \left[\mPhi(\mY_t, a) \mid \calG_t\right]$ can then be converted to a conditional expectation over $(\mY_{1:t-1}, \eta_{1:t-1}, \mS_{1:t})$.
Under Assumption~\ref{assump:martingale:difference}, such conditional expectation only depends on $\mS_{1:t}$, rendering Equation~\eqref{eq:vareps:define} a martingale difference sequence. 
The complete proof is given in the Appendix~\ref{lem2}.

With $\varepsilon_t$ forming a martingale difference sequence, and considering the reward decomposition~\eqref{eq::reward-decomposition}, we can then adapt self-normalized concentration techniques from linear contextual bandits \citep{abbasi2011improved}. However, an important caveat arises: the effective context 
$
\EE\big[\mPhi(\mY_t, A_t)|\mS_{1:t}, A_t\big]
$
is unknown because the true context distribution $\PP$ is unobserved. Instead, it can only be approximated by 
$
\EE_{\hat p}\big[\mPhi(\mY_t, A_t)|\mS_{1:t}, A_t\big].
$
Our analysis therefore requires an additional sensitivity argument that quantifies how inaccuracies in the imputed contexts affect the cumulative regret, leading to regret bounds that explicitly depend on the approximation error between $\hat \PP$ and $\PP$.

At each time $t$, define the conditional instantaneous regret between $A_t$ and $\Astar_t$ given the observed context $\mS_{1:t}$ as
\begin{equation} \label{eq:gap-t:def}
    \regret_t = \EE \left[R(t, \Astar_t) - R(t, A_t) \mid \mS_{1:t}\right],
\end{equation}
and define the cumulative conditional regret up to horizon $T$ as  
\[
    \Regret_T := \sum_{t=1}^T \regret_t.
\]

\begin{remark}[Alternative Regret Formulation]
\label{rem:alternative_regret}
Our regret is defined against an oracle observing the full context $Y_t$. Alternatively, one might consider a baseline that acts on the conditional expected context: $A_t^\dagger = \arg\max_a \mathbb{E}[\vthetastart \mPhi(\mY_t,a) \mid \mS_{1:t}]$. By Jensen's inequality, the expected reward of $A_t^\dagger$ is bounded by that of our full-context oracle:
$$
\max_a \mathbb{E}[\vthetastart\mPhi(\mY_t,a) | \mS_{1:t}] \le \mathbb{E}\Big[\max_a \vthetastart\mPhi(\mY_t,a) | \mS_{1:t}\Big].
$$
Therefore, bounding the regret against the full-context oracle establishes a strictly stronger theoretical guarantee that naturally upper-bounds the regret against the conditional-expectation baseline.
\end{remark}

We now state the main result. The next theorem provides a high-probability upper bound on the cumulative regret of Algorithm~\ref{alg:linucb} under general context distributions. The proof is provided in the Appendix~\ref{thm1proof}.
\begin{theorem}\label{thm:regret:upper:bound}
    Suppose that $\|\vthetastar\|_2 \!\leq \!1$, and let Assumptions~\ref{assump:sub-G} and~\ref{assump:martingale:difference} hold.
    For a given $\delta\in(0, 1)$, in Algorithm \ref{alg:linucb} choose
    \begin{equation} \label{eq:gamma_t:define}
    \gamma_t := \gamma^{(0)}_t + 3d^2 \sum_{\tau=1}^t D_t,
    \end{equation}
    where
    \begin{equation*}
        \gamma_t^{(0)} := 3\lambda + 6(\sigma_{\eta} + 2)^2 \log \left[\frac{4 t^2}{\delta} \left(1 + \frac{t B^2}{d \lambda}\right)^d\right],
    \end{equation*}
    and $D_t$ is defined in \eqref{eq:Dt:def0}. Then, with probability at least $1-\delta$, the regret of Algorithm \ref{alg:linucb} satisfies 
    \begin{equation*}
        \Regret_T \leq \Regret^{(\texttt{imp})}_T + \Regret^{(\texttt{lin})}_T.  
    \end{equation*}
    Here,
    \begin{equation*}
        \Regret^{\texttt{(lin)}}_T := 2\sqrt{2 \gamma^{(0)}_T dT \log\left(1 + \frac{T B^2}{d \lambda}\right)}
    \end{equation*}
    denotes the standard $\tilde{\cO}(d\sqrt{T})$ cumulative regret achieved by the vanilla \texttt{OFUL} algorithm, and
    \begin{equation*}
    \begin{aligned}
        \Regret^{\texttt{(imp)}}_T =  4\sqrt{6 d^3\left(\sum_{t=1}^T D_t\right)T \log \left(1 \!+\! \frac{T B^2}{d \lambda}\right)} 
    \end{aligned}
    \end{equation*}
    captures the additional cumulative regret due to imputing missing context with the pretrained model.
\end{theorem}

Note that both Lemma~\ref{lem:conditional-kld} and Theorem \ref{thm:regret:upper:bound} remain valid regardless of how $\hat p$ is trained or whether it is correctly specified. Thus, our theory is applicable to a broad range of modern machine learning models.

\begin{remark}\label{rem:Dt}
In Theorem \ref{thm:regret:upper:bound}, the choice of hyperparameters $\gamma_t$ depends on $D_t$, which may not be directly known. In many practical settings, however, $D_t$ or its order can be reasonably estimated. For example, if the dimensions of $\mPhi(\mY_t, a)$ and $\mY_t$ are bounded and the dependence of $\mY_t$ on $\mS_{1:t}$ is parametric within a fixed window (i.e., depending only on the most recent few $\mS_{\tau}$'s), then $D_t$ is typically of order $\tilde{\cO}((NT_0)^{-1/2})$, leading to $\Regret^{\texttt{(imp)}}_T=\tilde{\cO}(T(NT_0)^{-1/2})$. More generally, when the dependence is nonparametric but smooth, the order of $D_t$ can also be derived (see Section~\ref{subsec:nonparam:ex}). In both parametric and nonparametric settings---and in more general cases without structural assumptions---$D_t$ and $\gamma_t$ may also be chosen in a data-driven manner. We defer a detailed discussion to the Appendix~\ref{datadriven}.
\end{remark}

\begin{remark}
We also provide a variant of Algorithm \ref{alg:linucb}, which combines the imputations with an alternative online exploration algorithm similar to the idea from \citet{he2022nearly}. We show that an improved regret guarantee is possible when the order of total imputation error is known and the error varies across time. See Appendix \ref{app:robust_variant} for details.
\end{remark}

\begin{remark}[Time-Varying Context Dimensions and Structures]
\label{rem:varying_dimension}
We explicitly emphasize that our framework does not restrict the dimension of the observed state $\mS_t$ to be fixed across all time steps. The notation $\mS_t \in \mathbb{R}^{d_{S,t}}$ accommodates highly dynamic real-world environments where the availability of sensors or recorded features fluctuates over time (e.g., Missing-Not-At-Random scenarios). Our theoretical guarantees (Theorem \ref{thm:regret:upper:bound}) hold even when the online learner does not share the exact same set of observed features at every round. 
To empirically substantiate this, we demonstrate \texttt{PULSE-UCB}'s robustness to dynamically changing observed feature sets via a Random Feature Masking experiment in Appendix~\ref{app:random_feature_masking}.
\end{remark}

\subsection{Application of Theorem \ref{thm:regret:upper:bound}}
\label{subsec:nonparam:ex}

We now provide several examples that yield explicit rates for $D_t$ in Theorem~\ref{thm:regret:upper:bound} and the resulting cumulative regret of Algorithm~\ref{alg:linucb}. These examples are intentionally simplified for clarity but remain representative, and the ideas extend to more general settings.

Suppose that the full context $\mY_t$ can be written as $(\mS_t, W_t) \in \R^{d_S} \times \R$, where $\mS_t$ denotes the partially observed features in the bandit period and $W_t$ denotes the features that are missing. 

\paragraph{Linear Model} Consider a linear model where
\begin{equation}\label{eq:W:linear}
    W_t = \sum_{j=0}^{m} \vbeta_j^\top \mS_{t-j} + \xi_t, \; \xi_t \sim \calN(0,1), \; \forall t \in [T] 
\end{equation}
and $\mS_{-j} = \boldsymbol{0}$ for $j \in [m]$.
The historical data contains $N$ i.i.d.~observations of length $T_0$ from Equation~\eqref{eq:W:linear}: 
\begin{equation}\label{eq:calD:lin}
    \calD = \left\{\mY^{(0)}_{i,1:T_0}\right\}_{i=1}^N = \left\{ \left( \mS^{(0)}_{i,1:T_0}, W^{(0)}_{i,1:T_0}\right) \right\}_{i=1}^N
\end{equation}

\begin{proposition} \label{cor:exp:lin}
    Suppose that the historical data $\calD$ follows Equation~\eqref{eq:calD:lin} and the missing feature $W_t$ follows Equation~\eqref{eq:W:linear}.
    Assume that $\mS^{(0)}_{i,t} \stackrel{i.i.d.}{\sim} \calN(\boldsymbol{0}, \mI_{d_S})$ for all $i \in [N]$ and $t \in [T_0]$, and that $T_0\geq 2m$. 
    There exists a pretrained model $\hat{p}$ such that 
    \begin{equation*}
        \EE \sqrt{D_t} \lesssim \sqrt{\frac{m d_S}{N T_0}}.
    \end{equation*}
    Thus, the expected cumulative regret of Algorithm~\ref{alg:linucb}, taken over $\mS_{1:T}$ and $\calD$, satisfies
    \begin{equation*}
    \begin{aligned}
        \EE \left[\Regret_T\right]=\tilde{\cO}\left(T\sqrt{\frac{m d_S d^3}{N T_0}} + d\sqrt{T}\right).
    \end{aligned}
    \end{equation*}
\end{proposition}

\paragraph{Nonparametric Model} As another example, consider the case where $T_0 = 1$, so that the historical dataset $\calD$ contains $N$ i.i.d.~samples 
\begin{equation}\label{eq:calD:def:main}
    \calD = \left\{\mY^{(0)}_i\right\}_{i=1}^N = \left\{(\mS^{(0)}_i, W^{(0)}_i)\right\}_{i=1}^N
\end{equation}
where each missing feature $W_i\in\mathbb R$.
For simplicity, assume $\mS^{(0)}_i \sim \operatorname{Unif}\left([0,1]^{d_S}\right)$.
Consider a nonparametric regression model where for all $i \in [N]$, 
\begin{equation} \label{eq:W:distribution}
    W^{(0)}_i = f\left(\mS^{(0)}_i\right) + \xi_i, \quad \xi_i \sim \calN(0,1).
\end{equation}
Here $f$ is a scalar-value function that satisfies the Hölder smoothness condition with parameters $(\beta, L)$. 
\begin{assumption}[Hölder Smoothness]\label{assump:smooth}
    A function $f: \R^{d_S}\to \R$ satisfies the Hölder condition with parameters $(\beta, L)$ if for all $\vs, \vs' \in \R^{d_S}$, 
    \begin{equation} \label{eq:beta-smooth}
        \left|f(\vs) - f(\vs')\right| \leq L \left\|\vs - \vs'\right\|_2^{\beta}
    \end{equation}
    for some $\beta \in (0, 1]$ and $L > 0$. Denote the class of such functions as $\calF_{\beta, L}$. 
\end{assumption}

\begin{proposition} \label{cor:exp:non}
    Suppose that the historical data $\calD$ is specified by Equation~\eqref{eq:calD:def:main} and the missing feature $W$ follows Equation~\eqref{eq:W:distribution} with $f$ satisfying Assumption~\ref{assump:smooth}. 
    Then there exists $\hat{p}$ such that 
    \begin{equation*}
        \EE \sqrt{D_t} \lesssim N^{-\frac{\beta}{2\beta + d_S}}
    \end{equation*}
    and thus, the expected cumulative regret of Algorithm~\ref{alg:linucb}, taken over $\mS_{1:T}$ and $\calD$, satisfies
    \begin{equation*}
    \begin{aligned}
        \EE \left[\Regret_T\right]=\tilde{\cO}\left(Td^{\frac{3}{2}} N^{-\frac{\beta}{2\beta + d_S}} + d\sqrt{T}\right).
    \end{aligned}
    \end{equation*}
\end{proposition}
In Section~\ref{sec:lower}, we show that this upper bound is near-minimax optimal in both the time horizon and the auxiliary sample size, provided the auxiliary dataset is sufficiently large.

The proof of both propositions, as well as the explicit choice of $\hat p$ is included in the Appendix~\ref{propoproof}. 
As a remark, the examples above focus on a one-dimensional missing covariate. Extending to a general $d_W$-dimensional missing context is straightforward and leads to a similar result, except for an additional $\sqrt{d_W}$ factor from handling coordinates separately (e.g., via a union bound or vector-valued concentration).
Furthermore, the rates in Propositions~\ref{cor:exp:lin} and~\ref{cor:exp:non} are robust to mild covariate shift between the historical and online distributions; see Appendix~\ref{app:covariate-shift-theory} for details.

\section{LOWER BOUNDS}
\label{sec:lower}
To demonstrate the optimality of our algorithm, we establish a minimax lower bound by analyzing a carefully constructed two-arm contextual bandit instance with partially observed contexts. The exact data-generating process, including feature construction and verification of technical conditions, is provided in the Appendix~\ref{lowersetup}\&\ref{lowerproof}. We give a high-level overview below.  

Our construction highlights two fundamental sources of difficulty. First, the reward of one arm depends on an unobserved scalar $W$, whose conditional mean is determined by an unknown function $f \in \calF_{\beta,L}$ defined on a $d_{\non}$-dimensional subset of the observed context $\mS$. When historical data are limited, the challenge of estimating $f$ dominates the regret. Second, the rewards of both arms involve a linear parameter $\vtheta^\star$ acting on the complementary $d_{\lin}$-dimensional subset of $\mS$. The two subsets together form a partition of $\mS$, so that $d_{\lin} + d_{\non} = d_S$. Once $f$ can be accurately estimated from pretraining data, the remaining difficulty reduces to online learning of $\vtheta^\star$, which contributes a $\sqrt{T}$ regret term.

By alternating between these two regimes, the construction forces both sources of error to matter: historical samples provide noisy information about $f$, while online bandit interaction governs the estimation of $\vtheta^\star$. As a result, the minimax regret necessarily includes two additive components: one tied to the nonparametric rate for learning $f$, and the other to the linear rate for estimating $\vtheta^\star$. Full details 
are deferred to the Appendix~\ref{lowersetup} \& \ref{lowerproof}.

\begin{theorem}[Informal Lower Bound] \label{thm:lower:informal}
Consider the two-arm contextual bandit problem with partially observed contexts $\mS_t \in \R^{d_S}$. 
Under suitable regularity conditions, there exists a construction 
such that the minimax expected cumulative regret satisfies the rate of 
\[
    \Omega\!\left( T N^{-\frac{\beta}{2\beta + d_\non}} + \sqrt{d_\lin T} \right),
\]
where $d_{\non}, d_{\lin} > 0$ denote the nonparametric and linear dimensions of the observed context, respectively, and $d_{\non} + d_{\lin} = d_S$. 
\end{theorem}


\begin{remark}  
When $N=\Omega\!\big(T^{\frac{2\beta+d_S}{2\beta}}\big)$, both the upper bound in Proposition~\ref{cor:exp:non} and the lower bound in Theorem~\ref{thm:lower:informal} reduce to $\sqrt{T}$ (up to logarithmic factors). Consequently, for sufficiently large $N$, Algorithm~\ref{alg:linucb} attains near-minimax optimality. Notably, this matches the oracle rate when the context is fully observed, indicating that with ample data there is no efficiency loss when leveraging a well-suited pretrained model.

For small $N$ (taking $d_{\non}=d_S-1$ in Theorem~\ref{thm:lower:informal}), we observe a slight difference in the $N$-dependence relative to Proposition~\ref{cor:exp:non}. This stems from our proof’s partition of $\mS$ into complementary subsets to decouple the nonparametric and linear components. Allowing the linear part to also depend on the nonparametric coordinates would likely shift the dependence toward $d_S$, but entails substantially complicated analysis. Sharpening the small-$N$ dependence is an appealing direction for future work.
\end{remark}

\section{NUMERICAL EXPERIMENTS}
\label{sec:numeric}
In this section, we validate our theory and algorithm with simulations on synthetic data and the real Taobao Ad Display/Click dataset \citep{taobao}. To facilitate reproducibility, the complete source code is publicly available at \url{https://github.com/Heyan-Zhang/PULSE-UCB}.

\subsection{Synthetic Experiments}
\label{subsec:synthetic}

In the synthetic experiments, the full context $\mY_t = (\mS_t, W_t)$, where $\mS_t$ denotes the observed context and $W_t$ the unobserved part. The observed context $\{\mS_t\}_{t\geq 1}$ follows a stationary $\mathrm{ARMA}(2,2)$ process: $\mS_t = \phi_1 \mS_{t-1} + \phi_2 \mS_{t-2} + \varepsilon_t + \theta_1 \varepsilon_{t-1} + \theta_2 \varepsilon_{t-2},$
with $(\phi_1, \phi_2, \theta_1, \theta_2) = (0.75, -0.25, 0.65, 0.35)$ and $\varepsilon_t\sim \mathcal{N}(0, 0.1^2)$. 
The unobserved context $W_t$ depends on a feature vector $\bx_{t} \in \mathbb{R}^2$ summarizing recent context history: $\bx_{t}=(1, x_{t,2})^\top$, where $x_{t, 2} = (\mS_t + \mS_{t-1} + \mS_{t-2})/3$.
We consider two cases of how $W_t$ depends on $\bx_t$: (a) \textbf{Linear}: $W_t = \boldsymbol{\beta}_{*}^\top \bx_{t} + \xi_{t}$; (b) \textbf{Nonlinear}: $W_t = \boldsymbol{\beta}_{*}^\top \bx_{t} + \sin(\rho \cdot x_{t,2}) + \xi_{t}$, where $\rho=4$. In both settings, we choose $\boldsymbol{\beta}_{*} = (0.50, -0.14)$ and $\xi_{t} \sim \mathcal{N}(0, 0.1^2)$.
Finally, the reward $R_t$ follows (\ref{eq:Rta:define}) with
$\mPhi(\mY_t, a_t) = (1, \ \mS_t, \ W_t, \ \mS_t \cdot a_t)^\top$, 
$\vthetastar = (0.65, 1.52, -0.23, -0.23)$, and $\eta_t \sim \mathcal{N}(0, 0.05^2)$.

\begin{figure}[htbp]
    \centering
    \includegraphics[width=0.94\linewidth]{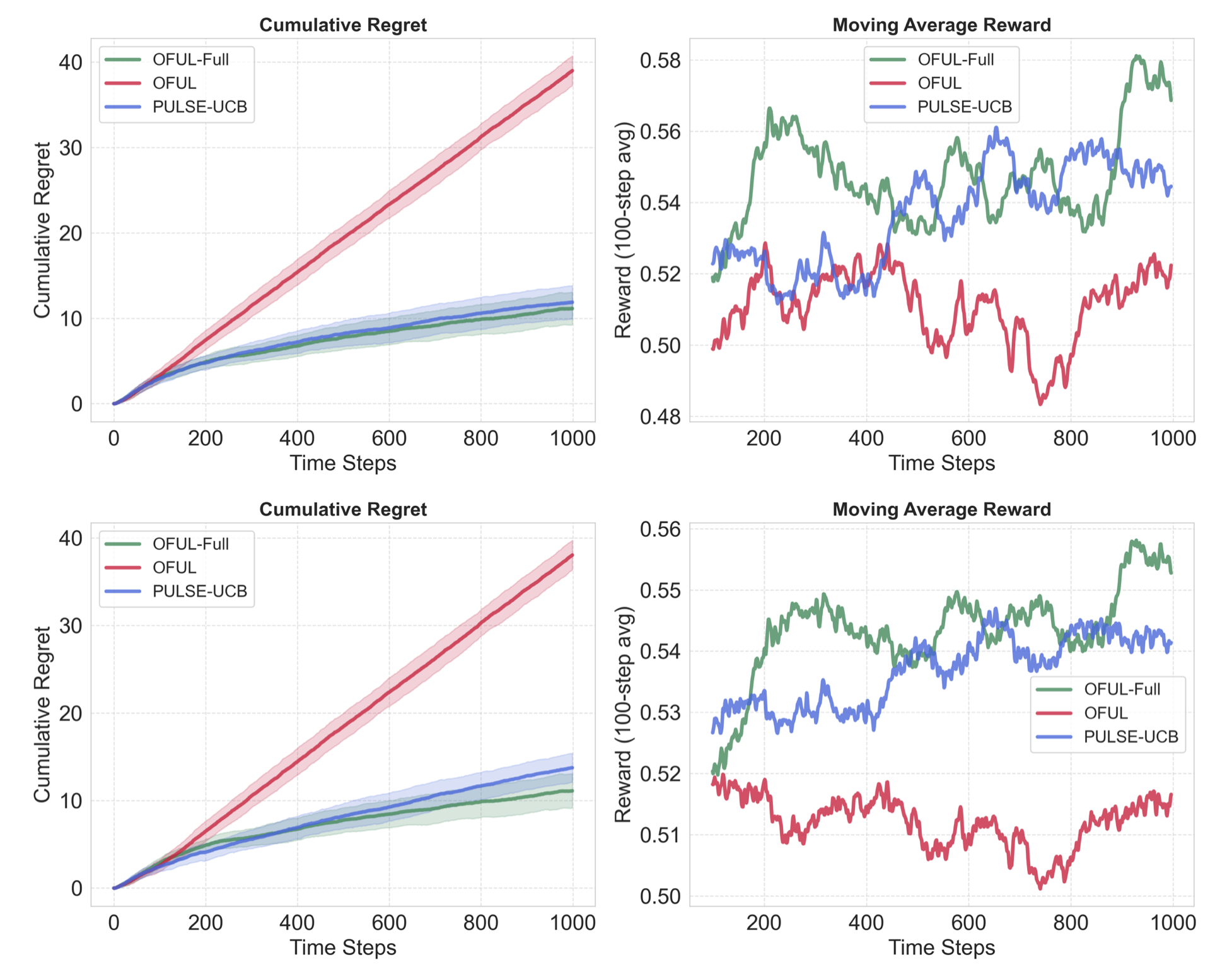}
    \captionsetup{font=scriptsize}
    \caption{
    Comparison of algorithms in synthetic experiments. Left: cumulative regret. Right: 100-step moving average reward. Top: linear missing-feature setting (a). Bottom: nonlinear setting (b). Shaded areas denote $\pm$ one standard error over 30 trials.}
    \label{fig:comparison_results}
\end{figure}

\texttt{PULSE-UCB} consists of two phases. In pretraining, a context transition model is learned from $N=1000$ historical time series of length $T_0=100$ to predict the latent context $W_t$. In the online evaluation, the agent runs for $T=1000$ steps. We compare against two benchmarks: (i) \texttt{OFUL}, a naive agent that ignores $W_t$ and uses only $\mS_t$; (ii) \texttt{OFUL-Full}, an idealized agent with access to the full context $\mY_t=(\mS_t, W_t)$. The cumulative regret, averaged over 30 independent trials, is shown in Figure~\ref{fig:comparison_results}. As expected, \texttt{OFUL-Full} achieves the lowest regret since it observes the full context, while \texttt{OFUL} performs worst by ignoring the missing component. In both the linear and nonlinear settings for the missing context, \texttt{PULSE-UCB} performs nearly as well as \texttt{OFUL-Full}, demonstrating the clear benefit of leveraging a predictive model for the unobserved context. 
To further demonstrate the performance of our approach, we also compared \texttt{PULSE-UCB} with \texttt{CLBBF}~\citep{kim2023contextual}. However, because \texttt{CLBBF} intrinsically relies on a Missing-At-Random (MAR) assumption, it struggles to handle the more general Missing-Not-At-Random (MNAR) structures present in our setting. Consequently, \texttt{PULSE-UCB} consistently outperforms \texttt{CLBBF} and achieves near-oracle performance. Additional experimental details, including sensitivity analyses on pre-training data scales ($N$ and $T_0$), are deferred to Appendix \ref{app:additional_baseline}.

\setlength{\parskip}{0.2em}

\subsection{Real-World Experiments}
\label{subsec:real-world}
To evaluate our method in a practical setting, we use the public Taobao Ad Display/Click dataset~\citep{taobao}, which contains 186,730 advertisement display/click records from Taobao.com. Each record includes 83 features describing user and ad attributes such as gender, age, consumption grade, brand, and category. We embed the features into a 32-dimensional space and partition them into 16 observed features ($\mS_t$) and 16 unobserved features (${\mW}_t$). All algorithms are evaluated on $80\%$ of the data. For \texttt{PULSE-UCB}, we additionally use the remaining $20\%$ for pretraining the context transition model. The action corresponds to selecting an ad (adgroup ID), and the reward is the binary click feedback (1 if clicked, 0 otherwise). Further preprocessing details are deferred to the Appendix~\ref{details_real}.

\begin{figure}[htbp]
    \centering
    \includegraphics[width=0.75\linewidth]{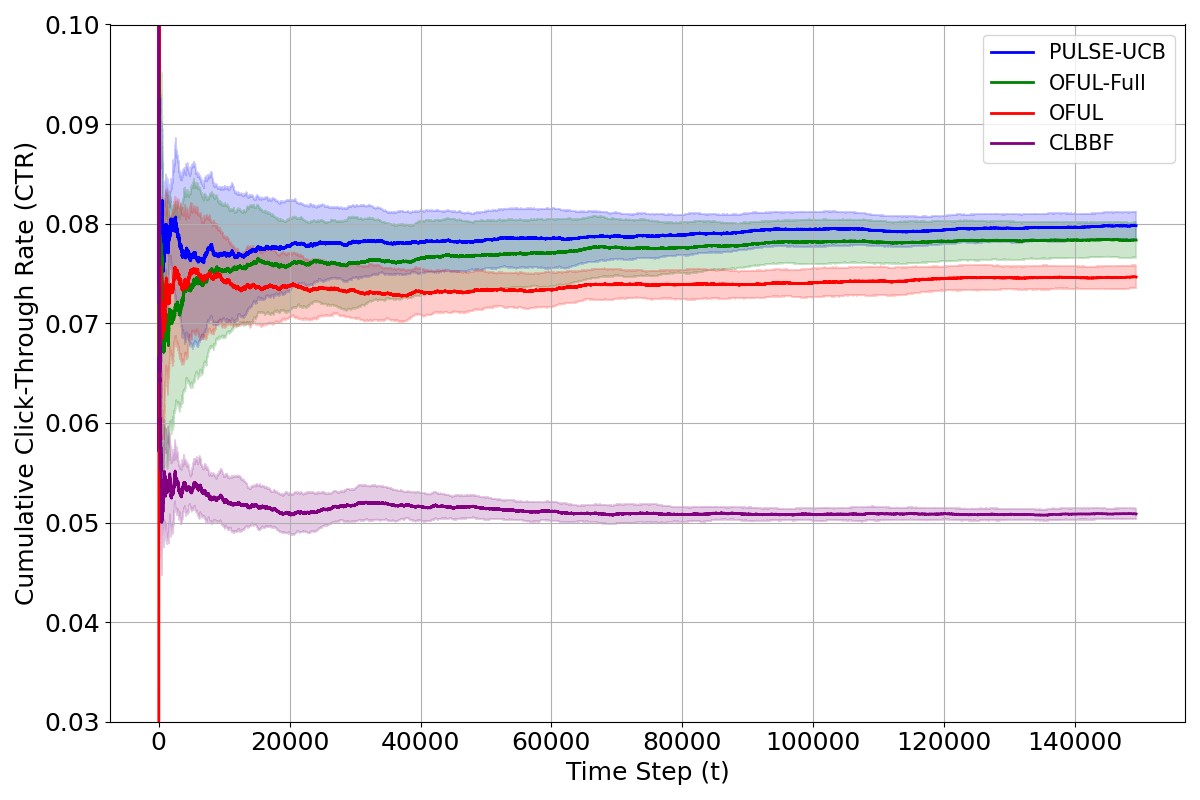}
    \captionsetup{font=scriptsize}
    \caption{
    Algorithm comparison on the Taobao dataset. Shaded areas denote $\pm$ one standard error over 5 runs.}
    \label{fig:taobao_result}
\end{figure}

We compare \texttt{PULSE-UCB} with three baselines: \texttt{OFUL}, which ignores the missing context; \texttt{OFUL-Full}, which has access to the full context; and \texttt{CLBBF}~\citep{kim2023contextual}, designed for bandits with stochastically missing features. We compare these algorithms over $T \approx 1.5 \times 10^5$ steps, with $K=20$ arms per step, averaging results over 5 runs. Figure~\ref{fig:taobao_result} shows that \texttt{PULSE-UCB} greatly outperforms \texttt{OFUL}, highlighting the benefit of context reconstruction, and also surpasses \texttt{CLBBF}, whose mechanism struggles under structural missingness. Notably, \texttt{PULSE-UCB} achieves performance nearly indistinguishable from the ideal \texttt{OFUL-Full}, indicating that the pretraining step not only imputes the missing context but also produces a feature representation well suited for linear bandit learning.

\textbf{Discussion and future directions.} 
We proposed a new algorithm, \texttt{PULSE-UCB} (Algorithm~\ref{alg:linucb}), which leverages imputation models pretrained on historical data to address linear contextual bandits with missing covariates. We established regret guarantees in Theorem~\ref{thm:regret:upper:bound} and showed near-optimality via the lower bound in Theorem~\ref{thm:lower:informal}. Empirical results in Section~\ref{sec:numeric} demonstrate strong performance across both synthetic and real-world datasets. Future directions include extending our framework to more general decision-making problems (e.g., Markov decision processes), accommodating more complex missing data mechanisms, and developing adaptive strategies to update the pretrained model during bandit interactions.
\newpage

\bibliographystyle{apalike}
\bibliography{main}

\section*{Checklist}



\begin{enumerate}

  \item For all models and algorithms presented, check if you include:
  \begin{enumerate}
    \item A clear description of the mathematical setting, assumptions, algorithm, and/or model.
    
    \textbf{Yes}: Our problem formulation, theoretical assumptions, and the proposed \texttt{PULSE-UCB} algorithm are detailed in Sections~\ref{sec:setup} and~\ref{sec:method}.
    
    \item An analysis of the properties and complexity (time, space, sample size) of any algorithm.
    
    \textbf{Yes}: We provide a theoretical analysis of the regret bound in Section~\ref{sec:main-result} and~\ref{sec:lower}.
    
    \item (Optional) Anonymized source code, with specification of all dependencies, including external libraries.
    
    \textbf{Yes}: We provide anonymized source code for all our experiments in the supplementary material. 
  \end{enumerate}

  \item For any theoretical claim, check if you include:
  \begin{enumerate}
    \item Statements of the full set of assumptions of all theoretical results. 
    
    \textbf{Yes}: All assumptions for our theoretical results are formally stated in Sections~\ref{sec:setup}-\ref{sec:lower}.
    
    \item Complete proofs of all theoretical results. 
    
    \textbf{Yes}: Complete proofs for all theorems and lemmas are provided in Appendix.
    
    \item Clear explanations of any assumptions. 
    
    \textbf{Yes}: We have provided intuitions and discussions for all our key assumptions.
  \end{enumerate}

  \item For all figures and tables that present empirical results, check if you include:
  \begin{enumerate}
    \item The code, data, and instructions needed to reproduce the main experimental results (either in the supplemental material or as a URL).
    
    \textbf{Yes}: The code and instructions to reproduce all experimental results are in the supplementary material. For the real-world experiment, our code is based on the public Taobao dataset.
    
    \item All the training details (e.g., data splits, hyperparameters, how they were chosen).
    
    \textbf{Yes}: We provide all training details, including data splits and hyperparameter settings, in the paper (Section~\ref{subsec:synthetic} and~\ref{subsec:real-world}) as well as the Appendix.
    
    \item A clear definition of the specific measure or statistics and error bars (e.g., with respect to the random seed after running experiments multiple times). 
    
    \textbf{Yes}: All our results are over multiple independent runs and shown with error bars.
    All shaded areas denote $\pm$ one standard error over multiple runs.
    
    \item A description of the computing infrastructure used. (e.g., type of GPUs, internal cluster, or cloud provider).
    
    \textbf{No}: Our experiments are computationally inexpensive and were run on a standard desktop computer; they do not require any specialized computing infrastructure.
  \end{enumerate}

  \item If you are using existing assets (e.g., code, data, models) or curating/releasing new assets, check if you include:
  \begin{enumerate}
    \item Citations of the creator If your work uses existing assets. 
    
    \textbf{Yes}: We cite the source of the Taobao dataset in Section~\ref{subsec:real-world}.
    
    \item The license information of the assets, if applicable. 
    
    \textbf{Yes}: This Taobao dataset is provided by Alibaba and distributed under the CC BY-NC-SA 4.0 license.
    
    \item New assets either in the supplemental material or as a URL, if applicable. 
    
    \textbf{Not Applicable}: We use an existing public dataset and do not release new assets.
    
    \item Information about consent from data providers/curators. 
    
    \textbf{Not Applicable}: We use a well-established public dataset that has been anonymized.
    
    \item Discussion of sensible content if applicable, e.g., personally identifiable information or offensive content.
    
    \textbf{Not Applicable}: The dataset is anonymized and does not contain personally identifiable or offensive content.
  \end{enumerate}

  \item If you used crowdsourcing or conducted research with human subjects, check if you include:
  \begin{enumerate}
    \item The full text of instructions given to participants and screenshots.
    
    \textbf{Not Applicable}: Our research does not involve human subjects or crowdsourcing.
    
    \item Descriptions of potential participant risks, with links to Institutional Review Board (IRB) approvals if applicable. 
    
    \textbf{Not Applicable}: Our research does not involve human subjects.
    
    \item The estimated hourly wage paid to participants and the total amount spent on participant compensation. 
    
    \textbf{Not Applicable}: Our research does not involve human subjects.
  \end{enumerate}

\end{enumerate}

\clearpage
\appendix

\numberwithin{equation}{section}
\onecolumn

\addcontentsline{toc}{section}{Appendix}
\tableofcontents

\section{Additional literature review}
\label{litrev}
\textbf{AI-assisted decision-making.} Recently, there has been growing interest in applying AI, including foundation models, to enhance decision-making. For example, \citet{tianhui2024active, zhang2025contextual} design Thompson sampling algorithms for online bandits that treat uncertainty as missing future outcomes, imputing them with pretrained generative models to optimize policies. \citet{chen2021decision, janner2021offline} recast offline reinforcement learning as sequence modeling over trajectories to improve decisions, while \citet{lin2023transformers, lee2023supervised} study in-context reinforcement learning, showing that supervised pretraining on past trajectories enables models to approximate algorithms such as LinUCB and Thompson sampling with regret guarantees. Applications include LLM-assisted adaptive experimentation for content delivery \citep{ye2025lola} and human–AI collaboration in linear bandits with resource constraints for healthcare \citep{cao2024hr}. Our work focuses on the missing-context issue, specifically in online contextual bandits, and provides near-optimal regret guarantees that guide the use of pretrained models for imputation in this setting.

\textbf{Imputation in statistics and ML.} Imputation has long been a central strategy across statistics and machine learning for handling missing information. A classical example is the EM algorithm \citep{dempster1977maximum}, which provides a likelihood-based framework for parameter estimation with incomplete data and remains highly influential in this area. In causal inference, imputation is widely used for estimating potential outcomes under counterfactual interventions \citep{little2019statistical}. From a statistical learning perspective, recent work has incorporated modern machine learning models to deal with missing responses: \citet{xia2024prediction} propose surrogate training that leverages helper covariates to impute pseudo-responses for unlabeled data, yielding prediction improvements with excess risk guarantees, while \citet{angelopoulos2023prediction} demonstrate that combining a small labeled set with imputed outcomes enables valid confidence intervals and hypothesis tests. Our work contributes to this line of research by extending imputation-based methods to sequential decision-making with partially observed contexts, and quantify the impact of imputation quality on online learning performance.

\section{Practical Motivations for Asymmetric Context Availability}
\label{app:motivation}

A core premise of our framework is that the auxiliary offline dataset $\mathcal{D}$ contains richer or less noisy contextual information ($Y_t$) than what can be feasibly observed during real-time online decision-making ($S_t$). This asymmetric context availability is driven by how data are collected in practice. In many real-world environments, online intervention systems must operate under strict privacy constraints, tight computational limits, or high risks of missing data. Conversely, offline datasets are often collected from prior clinical trials, extensive observational cohorts, or archival systems, affording to store and process much richer state variables.

Rather than assuming the online agent can access a complete context capturing all factors that affect the reward, our framework aims for \emph{context enrichment}: leveraging the rich contextual variables in the auxiliary data to build a more accurate reward approximation offline, which in turn improves the decisions made online using the limited context. We illustrate this setting with two concrete domains:

\paragraph{Digital Health.}
In mobile health (mHealth) and digital interventions, multiple sources of context-rich offline data are common, such as prior trials, earlier interventions, and observational cohorts equipped with detailed baseline surveys or high-burden ecological momentary assessments \citep{liao2020personalized,coughlin2024mobile}. By contrast, online intervention systems running on local devices face tight bandwidth limits, battery constraints, and strict privacy regulations. Furthermore, user disengagement and frequent sensor failures often lead to missing or coarsened context in practice \citep{gonzalez2025practical}. In our setting, the offline full state $Y_t$ represents these rich baseline variables, low-frequency assessments (e.g., stress assessed via validated multi-item scales), and high-resolution wearable signals (e.g., heart-rate variability). The online observed state $S_t$, however, is typically restricted to coarser, continuously available signals like step counts or simple self-reports.

\paragraph{Online Education.}
In intelligent tutoring systems and online education platforms, the online bandit algorithm typically only observes a limited set of immediate signals $S_t$, such as short performance summaries or recent clickstreams. However, educational platforms maintain rich archival databases containing concept-level mastery estimates, comprehensive exam scores, and detailed background information from prior assessments \citep{lan2016contextual}. For instance, in large-scale MOOCs, frameworks like MOOClet utilize archival data (e.g., full homework records and fine-grained logs) to model detailed student states and design adaptive components \citep{williams2014mooclet}. Within our framework, these archival records constitute the auxiliary data $\mathcal{D}$, where $Y_t$ aggregates the richer, knowledge-related quantities inferred offline, while the online agent relies solely on the high-frequency, easily observable signals $S_t$.

\paragraph{Robustness to Distribution Shift.}
While our primary theoretical analysis focuses on the matched-distribution regime where the auxiliary dataset and the online environment share the same context distribution, the learned latent structure often generalizes robustly when this assumption is relaxed. As demonstrated empirically in Appendix~\ref{app:covariate_shift}, our method maintains strong performance even when the online environment experiences significant covariate shifts relative to the offline auxiliary data. A theoretical justification for this robustness, showing that the rates in Propositions~\ref{cor:exp:lin} and~\ref{cor:exp:non} are preserved under mild covariate shift, is given in Appendix~\ref{app:covariate-shift-theory}.

\section{Discussion of Assumption~\ref{assump:martingale:difference} and Autoregressive Dependence}
\label{app:ar-discussion}

Assumption~\ref{assump:martingale:difference} requires that $\EE[\mPhi(\mY_t,a) \mid \mY_{1:t-1}, \eta_{1:t-1}, \mS_{1:t}] = \EE[\mPhi(\mY_t,a) \mid \mS_{1:t}]$. As noted in the main text, this is satisfied whenever $\mY_t$ is generated as an arbitrary function of $\mS_{1:t}$ together with independent randomness, because the fresh noise at time $t$ carries no information about the past beyond what is already captured by $\mS_{1:t}$. This covers a broad class of models, including all i.i.d.\ contextual bandit settings as a special case.

The assumption fails, however, when the noise driving $\mY_t$ is itself temporally dependent, as in autoregressive (AR) models. In such cases, past noise innovations $\xi_{1:t-1}$ influence both $\mY_{1:t-1}$ and the current $\mY_t$, so that conditioning on $\mS_{1:t}$ alone no longer screens off the dependence on the past. The residual $\varepsilon_t$ defined in Equation~\eqref{eq:vareps:define} then fails to form a martingale difference sequence, and the self-normalized concentration machinery underlying our regret analysis breaks down. Handling $\varepsilon_t$ in this regime requires fundamentally different tools. Moreover, existing works on AR bandits suggest that the appropriate techniques depend heavily on the specific form of AR structure, making it difficult to develop a unified treatment. We illustrate this point with two representative examples below.

First, suppose we ignore the observed context $\mS_{1:t}$ and let the latent state evolve via action-dependent dynamics
\[
W_{t+1} = \mu(a_t) + \rho(a_t)\, W_t + \xi_t,
\]
where $\xi_t$ is i.i.d.\ noise. The reward satisfies $R(t, a_t) = \vthetastart \mPhi(\mY_t, a_t) + \tilde\eta_t$, giving a model whose transition coefficients depend on the agent's action. This is the action-driven AR bandit studied by \citet{bacchiocchi2024autoregressive}, for which modified UCB-type algorithms exploit the action-dependent structure to achieve the optimal $\cO(\sqrt{T})$ regret. Crucially, however, this structure is absent in our problem: the auxiliary historical data contain no action information, and the contextual process evolves according to fixed, action-independent dynamics. Techniques designed for action-driven AR dependence therefore do not apply.

Second, consider a form of AR dependence more representative of our setting. Let
\[
W_t = \alpha\, W_{t-1} + \xi_t, \qquad \xi_t \overset{\mathrm{i.i.d.}}{\sim} \calN(0,1), \quad |\alpha| < 1,
\]
and set $\mY_t = W_t$ with feature map $\mPhi(\mY_t, a) = \beta_a W_t$. The reward becomes $R(t,a) = \vthetastar\, \beta_a\, W_t + \eta_t$. Here the AR dependence resides in a latent process that influences the reward only multiplicatively through the action, producing a fundamentally different statistical structure from the first example. This setting corresponds to the latent autoregressive bandit model of \citet{trella2024non}, where the analysis requires Kalman-filter-type estimators and strong linear-Gaussian assumptions, and the resulting regret may exceed $\cO(\sqrt{T})$.

These two examples illustrate that it is currently not straightforward to propose a unified approach for handling AR dependence in bandit problems. The difficulty of the learning problem hinges critically on \emph{how} autoregressive dependence enters the model, and seemingly minor modeling choices, such as whether the AR dependence acts on the observed reward or on a latent state, and whether it is action-driven or governed by fixed dynamics, lead to qualitatively different algorithms and regret guarantees. Furthermore, both examples above assume that the feature map $\mPhi$ is linear in the contextual variables; once nonlinear dependencies are introduced, it is unclear how to adapt existing techniques or whether the same regret rates remain achievable.

Taken together, these observations underscore that extending our framework to accommodate autoregressive noise would require a case-by-case analysis tied to the specific AR structure at hand, rather than a single unified extension of Assumption~\ref{assump:martingale:difference}. We therefore leave this direction to future work.

\section{A data-driven approach for choosing \texorpdfstring{$D_t$}{Dt}}
\label{datadriven}

For any $t\in[T]$, recall that Algorithm~\ref{alg:linucb} requires an upper bound $D_t$ to calibrate the confidence balls. 
In Remark~\ref{rem:Dt} and Section~\ref{subsec:nonparam:ex} we described several cases where an explicit rate of $D_t$ is available.
Here we discuss a fully data-driven alternative based on \emph{uniform confidence bands} (UCBs) for the conditional mean under the ground-truth conditional law.

Let $p$ denote the ground-truth conditional distribution of $\mY_t\mid \mS_{1:t}$ and let $\hat p$ be an estimator of this conditional distribution built from historical data.
Fix an action $a \in \calA$. 
Write
\[
\mu_p(\vs)\;:=\;\EE\!\left[\Phi(\mY_t,a)\mid \mS_{1:t}=\vs\right],
\qquad
\mu_{\hat p}(\vs)\;:=\;\EE_{\hat p}\!\left[\Phi(\mY_t,a)\mid \mS_{1:t}=\vs\right],
\]
where we suppose $\Phi(\mY_t,a)$ to be one-dimensional for clarity (the multivariate case follows coordinatewise with a union adjustment).
Denote our imputation error at $\mS_{1:t} = \vs$ as
\[
\calE_t(p,\hat p;\vs)\;:=\;\big|\mu_p(\vs)-\mu_{\hat p}(\vs)\big|.
\]
Instead of bounding $D_t$, it suffices for us to control
\[
\calE_t(p,\hat p;\vs)
\]
for every $\vs$ simultaneously over a compact domain $\calS_t$ where $\mS_{1:t}$ is in.

We upper bound $\calE_t(p,\hat p;\vs)$ by combining (i) a \emph{uniform confidence band} for $\mu_p(\vs)$, centered at a reference estimator that \emph{does} admit UCBs, and (ii) a directly computable discrepancy between $\mu_{\hat p}$ and that reference estimator. 
Concretely, split the historical data into two folds $I_0$ and $I_1$ (sample splitting or cross-fitting):
\begin{enumerate}
\item On $I_0$, fit a reference conditional distribution $\hat p_0$ using a method with established UCBs (e.g., local-polynomial with robust bias correction or penalized splines with simultaneous bands).
Obtain a $(1-\alpha)$-UCB for $\mu_p$ on a grid $\calG_t\subset\calS_t$,
\[
\calC_{1-\alpha}(\vs)\;=\;\big[\mu_{\hat p_0}(\vs)\pm r_{1-\alpha}(\vs)\big],\qquad \vs\in\calG_t,
\]
where $r_{1-\alpha}(\vs)$ is the half-width delivered by the band construction.
\item On $I_1$, fit $\hat p$ (any estimation strategy; no UCB requirement).
\item By the triangle inequality, for any $\vs\in\calG_t$,
\begin{equation}\label{eq:triangle}
\big|\mu_p(\vs)-\mu_{\hat p}(\vs)\big|
\;\le\;
\underbrace{\big|\mu_p(\vs)-\mu_{\hat p_0}(\vs)\big|}_{\text{controlled by the UCB}}
\;+\;
\underbrace{\big|\mu_{\hat p_0}(\vs)-\mu_{\hat p}(\vs)\big|}_{\text{fully data-computable}}.
\end{equation}
\end{enumerate}
Taking suprema over $\calG_t$ and, if desired, extending from the grid to $\calS_t$ with a modulus-of-continuity bound yields a valid high-probability bound for $\sup_{\vs\in\calS_t}\calE_t(p,\hat p;\vs)$.
Under certain regularity conditions, one can extend the bound over the grid $\calG_t$ to the entire domain $\calS_t$ at the cost of a discretization penalty. 
In practice, when the grid $\calG_t$ is chosen sufficiently fine, this additional term becomes negligible, and one may safely restrict attention to $\calG_t$ without loss of generality. 
Suppose $\calC_{1-\alpha}$ is a $(1-\alpha)$ uniform confidence band for $\mu_p$ over $\calG_t$ centered at $\mu_{\hat p_0}$ (constructed on $I_0$), i.e.,
$$
\PP\!\big\{\mu_p(\vs)\in\calC_{1-\alpha}(\vs), \;\forall\,\vs\in\calG_t\big\}\ge 1-\alpha.
$$
Then with probability at least $1-\alpha$,
\begin{equation}\label{eq:sup-grid}
\calE_t(p,\hat p;\vs)
\;\le\;
\sup_{\vs\in\calG_t} r_{1-\alpha}(\vs)
\;+\;
\big|\mu_{\hat p_0}(\vs)-\mu_{\hat p}(\vs)\big|,
\end{equation}
yielding a data-driven choice for $\calE_t(p,\hat p;\vs)$ on $\calG_t$.

For practical purpose, one can follow the procedure below: 
\begin{enumerate}
\item \textbf{UCB machinery for the reference fit $\hat p_0$.}
Two widely used choices are:
(i) local-polynomial regression with \emph{robust bias correction} (RBC), whose studentized process admits valid simultaneous bands and is robust to MSE-optimal bandwidth choice; the quantiles are obtained via Gaussian/multiplier bootstrap of the sup-statistic;
(ii) penalized splines with simultaneous bands via volume-of-tube or bootstrap calibrations.%
\footnote{See, e.g., \citet{CalonicoCattaneoFarrell2018JASA,CattaneoCalonicoFarrell2018CERRD} for RBC-based bands and \citet{KrivobokovaKneibClaeskens2010JASA} for spline bands; multiplier bootstrap for suprema is treated in \citet{CCK2014AnnStat}.}
\item \textbf{Computing $\mu_{\hat p_0}$ and $\mu_{\hat p}$.}
For general $\Phi$ (which is known), approximate $\EE_{\hat p}[\Phi(\mY_t,a)\mid \mS_{1:t}=\vs]$ by Monte Carlo from $\hat p$ (and analogously for $\hat p_0$) with negligible simulation error relative to the statistical half-widths.
\item \textbf{From grid to domain.}
Choose $\calG_t$ fine enough relative to the smoothing scale (e.g., grid spacing $\ll$ bandwidth) and, if needed, add the modulus-of-continuity correction to pass from a grid-wide error bound to a domain-wide error bound.
\item \textbf{Coordinatewise or joint control (multi-dimensional $\mPhi$).}
Apply the above per coordinate and combine by Bonferroni (conservative), or calibrate a joint supremum over coordinates via multiplier bootstrap of a vector-valued process.
\end{enumerate}

As a special case, if the estimator used for $\hat p$ provides a valid UCB centered at $\mu_{\hat p}$, one may set $\hat p_0=\hat p$ and simply take
\(
\sup_{\vs\in\calG_t} r_{1-\alpha}(\vs).
\)
RBC-based local polynomials are particularly convenient here because the same fit supplies both the point estimates and a simultaneous band with good finite-sample coverage properties.

\paragraph{Caveat (high-dimensional context $\mS_{1:t}$).}
When the observed context $\mS_{1:t} = \vs$ lies in a high-dimensional space, it is generally impossible to obtain tight confidence bounds without additional structural assumptions. Specifically, nonparametric estimators suffer from the curse of dimensionality, causing inflated confidence bands and consequently large $\widehat D_t$. This reflects a fundamental limitation of nonparametric inference—without further assumptions, nontrivial guarantees cannot be achieved in the worst case. To address this issue, one may collect substantially more data or impose structural restrictions that effectively reduce the intrinsic dimension, such as additivity \citep{Meier2009Additive}, single-index models \citep{Ichimura1993SLS}, or shape constraints \citep{GroeneboomJongbloed2014,Chetverikov2015NPIVMonotonicity}.

\section{Proof of Lemma~\ref{lem:conditional-kld}}\label{lem1}

\begin{lemma}\label{lem:martingale:difference}
    Under Assumption~\ref{assump:martingale:difference}, for all $t \in [T]$, 
    \begin{equation*}
        \EE \left[\mPhi(\mY_t,a) \mid A_{1:t-1}, R_{1:t-1}, \mS_{1:t}\right] = \EE \left[\mPhi(\mY_t,a) \mid \mS_{1:t}\right].
    \end{equation*}
\end{lemma}

\begin{proof}
    Fixing $\mS_{1:t} = \vs_{1:t}$, we have  
    \begin{equation*}
    \begin{aligned}
        &\sup_{g: \|g\|_{\infty} \leq 1} \left\{\EE \left[g(\mY_{t}) \mid \mS_{1:t} = \vs_{1:t}\right] - \EE_{\hat{p}} \left[ g(\mY_{t}) \mid \mS_{1:t} = \vs_{1:t} \right] \right\} \\
        &~\stackrel{(i)}{=} d_{\TV}\left(\PP\left(\mY_{t} \mid \mS_{1:t} = \vs_{1:t}\right), \hat{\PP}\left(\mY_{t} \mid \mS_{1:t} = \vs_{1:t}\right)\right)\\
        &~\stackrel{(ii)}{\leq} \sqrt{\frac{1}{2} \operatorname{KL}\left(\PP\left(\mY_{t} \mid \mS_{1:t} = \vs_{1:t}\right) \| \hat{\PP}\left(\mY_{t} \mid \mS_{1:t} = \vs_{1:t}\right)\right)}
    \end{aligned}
    \end{equation*}
    where $(i)$ holds by the definition of total variation, $(ii)$ holds by Pinsker's inequality. 
    Taking expectation with respect to $\mS_{1:t}$  on both sides of the above display and applying Jensen's inequality, we obtain
    \begin{equation*}
    \begin{aligned}
        &\EE_{\mS_{1:t}} \sup_{g: \|g\|_{\infty} \leq 1} \left\{\EE \left[g(\mY_{t}) \mid \mS_{1:t}\right] - \EE_{\hat{p}} \left[ g(\mY_{t}) \mid \mS_{1:t} \right] \right\} \\
        &~\leq \sqrt{\frac{1}{2} \EE \KL\left(\PP\left(\mY_{t} \mid \mS_{1:t} = \vs_{1:t}\right) \| \hat{\PP} \left(\mY_{t} \mid \mS_{1:t} = \vs_{1:t}\right)\right)}\\
        &~= \sqrt{\frac{1}{2} \KL\left(\PP \left(\mY_{t} \mid \mS_{1:t} \right) \| \hat{\PP} \left(\mY_{t} \mid \mS_{1:t} \right)\right)}
    \end{aligned}
    \end{equation*}
    where the last equality follows from the definition of KL divergence between conditional distributions. 
\end{proof}

\section{Proof of Lemma~\ref{lem:eps:mds}}\label{lem2}

\begin{proof}
    Recall that for any $t \in [T]$, 
    \begin{equation*}
        \calG_{t-1} = \sigma\left(\mS_{1:t}, \mY_{1:t-1}, \eta_{1:t-1}, U_{1:t}\right). 
    \end{equation*}
    We remind the reader that $U_t$ is a auxiliary random variable used to select $A_t$ (e.g., $U_t$ captures the randomness involved in algorithms such as Thompson sampling or in breaking ties when selecting actions). 
    and $U_t$ is independent of $\left(\mS_{1: t}, R_{1: t-1}, A_{1: t-1}\right)$.
    To verify that $\{\varepsilon_t\}_{t=1}^T$ is a martingale difference sequence with respect to $\{\calG_t\}_{t=1}^T$, we need to verify two conditions, namely
    \begin{equation}\label{eq:adapted}
        \varepsilon_t \in \calG_t,    
    \end{equation}
    and
    \begin{equation}\label{eq:mean-zero}
        \EE\left[\varepsilon_t \mid \calG_{t-1}\right] = 0.
    \end{equation}
    Noting that for all $\tau \in [t-1]$, 
    \begin{itemize}
        \item[$\circ$] $A_\tau$ is a function of the observed history and the auxiliary random variable $\left(\mS_{1: \tau}, R_{1: \tau-1}, A_{1: \tau-1}, U_\tau\right)$. 
        \item[$\circ$] $R_{\tau}$ is a function of $A_{\tau}$, $\mY_{\tau}$ and $\eta_{\tau}$.
    \end{itemize}
    We conclude from the above observation that $(A_{1:t-1}, R_{1:t-1})$ is a function of $(\mY_{1:t-1}, \eta_{1:t-1}, U_{1:t-1})$ and 
    \begin{equation}\label{eq:At:adapted}
        A_t \in \sigma\left(\mS_{1:t}, \mY_{1:t-1}, \eta_{1:t-1}, U_{1:t}\right).
    \end{equation}
    Thus, we have 
    \begin{equation*}
        \mPhi(\mY_t, A_t) \in \sigma\left(\mY_{t}, A_{t}\right) \subset \sigma\left(\mY_{1:t}, \eta_{1:t}, U_{1:t+1}\right) \subset \calG_t
    \end{equation*}
    and Equation~\eqref{eq:adapted} holds. 
    For Equation~\eqref{eq:mean-zero} to hold, we have 
    \begin{equation*}
    \begin{aligned}
        \EE\left[\mPhi(\mY_t, A_t) \mid \calG_{t-1}\right] &= \EE\left[\mPhi(\mY_t, A_t) \mid \mS_{1:t}, \mY_{1:t-1}, \eta_{1:t-1}, U_{1:t}\right] \\
        &\stackrel{(i)}{=} \EE\left[\mPhi(\mY_t, A_t) \mid \mS_{1:t}, \mY_{1:t-1}, \eta_{1:t-1}, U_{1:t}, A_t\right]\\
        &\stackrel{(ii)}{=} \EE\left[\mPhi(\mY_t, A_t) \mid \mS_{1:t}, \mY_{1:t-1}, \eta_{1:t-1}, A_t\right]
    \end{aligned}
    \end{equation*}
    where equality $(i)$ holds from Equation~\eqref{eq:At:adapted} and equality $(ii)$ follows from $\mY_t$ is independent of $U_{1:t}$. 
    Applying Assumption~\ref{assump:martingale:difference} with the above display, it follows that 
    \begin{equation*}
    \begin{aligned}
        \EE\left[\mPhi(\mY_t, A_t) \mid \calG_{t-1}\right] &= \EE\left[\mPhi(\mY_t, A_t) \mid \mS_{1:t}, A_t\right].
    \end{aligned}
    \end{equation*}
    It is then straightforward to see that Equation~\eqref{eq:mean-zero} holds by the definition of $\varepsilon_t$. 
    We then conclude that $\{\varepsilon_t\}_{t=1}^T$ is a martingale difference sequence with respect to $\{\calG_t\}_{t=1}^T$


    Additionally, we show that $\{\varepsilon_t\}_{t=1}^T$ satisfies a sub-Gaussian tail condition, we only need to verify that it is a bounded sequence. 
    Since for any $a \in \calA$, under Assumption~\ref{assump:sub-G} and Equation~\eqref{eq:Rta:bounded}, 
    \begin{equation*}
        \vthetastart \mPhi(\mY_t, a) = \EE\left[R(t,a) \mid \calF_t\right] \in [-1,1],
    \end{equation*}
    it follows that 
    \begin{equation*}
        \left|\varepsilon_t\right| \leq \left|\vthetastart \mPhi(\mY_t, A_t) \right| + \left|\vthetastart \EE\left[\mPhi(\mY_t, A_t) \mid \mS_{1:t}, A_t\right]\right| \leq 2. 
    \end{equation*}
    By Azuma-Hoeffding inequality \citep[see Corollary 2.20 in][]{wainwright2019high}, we conclude that $\{\varepsilon_t\}_{t=1}^T$ is a martingale difference sequence with sub-Gaussian parameter $\sigma^2_\varepsilon \leq 4$. 
\end{proof}

\section{Proof of Theorem~\ref{thm:regret:upper:bound}}
\label{thm1proof}
\begin{proof}

Before presenting the proof, we briefly outline the main steps.
We first assume that for all $t \in [T]$, $\vthetastar \in \BALL_{t-1}$, where $\BALL_{t-1}$ is the confidence ball at step $t-1$ defined in Equation~\eqref{eq:ball:define}.
Under this assumption, we show that the regret at each step $t \in [T]$ can be decomposed into two components (see Equation~\eqref{eq:gap_t:bound:II}): the first reflecting the “width” of the confidence ellipsoid in the direction of the chosen decision $\vphihat_{t,A_t}$, and the second capturing the imputation error.
The former is bounded in Lemma~\ref{lem:xi1t:bound}, while the latter is controlled by Equation~\eqref{eq:xi2:bound}.
Finally, we select an appropriate sequence $\{\gamma_t\}_{t \in [T]}$ to guarantee that $\vthetastar \in \BALL_{t}$ with high probability.

Recall that $\vphihat_{t,a}$ is the conditional expectation of the context $\mPhi(\mY_t, a)$ given the partial observation $\mS_{1:t}$ under distribution $\hat{p}$, as defined in Equation~\eqref{eq:phihat:define}.
Let $\bar{\vtheta}_t \in \BALL_{t-1}$ denote the vector which maximizes the inner product $\vtheta^\top \vphihat_{t, A_t}$. 
Then
\begin{equation} \label{eq:thetaxAt:bound:I}
\begin{aligned}
    \bar{\vtheta}_t^\top \vphihat_{t,A_t} &= \max_{\vtheta \in \BALL_{t-1}} \vtheta^\top \vphihat_{t, A_t} = \max_{a \in \cA} \max_{\vtheta \in \BALL_{t-1}} \vtheta^\top \vphihat_{t,a} \\
\end{aligned}
\end{equation}
where the last equality follows from the way we choose $A_t$ as defined in Equation~\eqref{eq:At:define}.
Recall that $\Astar_t$ is the optimal action given by Equation~\eqref{eq:Astart:def}. 
The right-hand side of Equation~\eqref{eq:thetaxAt:bound:I} is lower bounded by
\begin{equation*}
\begin{aligned}
    \max_{a \in \cA} \max_{\vtheta \in \BALL_{t-1}} \vtheta^\top \vphihat_{t,a} &\geq \max_{\vtheta \in \BALL_{t-1}} \vtheta^\top \vphihat_{t, \Astar_t} 
    \geq \vthetastart \vphihat_{t, \Astar_t} \\
\end{aligned}
\end{equation*}
Adding and subtracting $\vthetastart \EE [\mPhi(\mY_t, \Astar_t) \mid \mS_{1:t}, \Astar_t]$ on the right-hand side of the above display yields
\begin{equation*}
\begin{aligned}
    \max_{a \in \cA} \max_{\vtheta \in \BALL_{t-1}} \vtheta^\top \vphihat_{t,a} &\geq \vthetastart \vphihat_{t, \Astar_t} - \vthetastart \EE \left[\mPhi(\mY_t, \Astar_t) \mid \mS_{1:t}, \Astar_t\right] + \vthetastart \EE [\mPhi(\mY_t, \Astar_t) \mid \mS_{1:t}, \Astar_t]\\
    &= \vthetastar \left(\vphihat_{t,\Astar_t} - \EE \left[\mPhi(\mY_t, \Astar_t) \mid \mS_{1:t}, \Astar_t\right]\right) + \vthetastart \EE [\mPhi(\mY_t, \Astar_t) \mid \mS_{1:t}, \Astar_t].
\end{aligned}
\end{equation*}
Taking the above display into Equation~\eqref{eq:thetaxAt:bound:I} gives
\begin{equation*}
    \bar{\vtheta}_t^\top \vphihat_{t,A_t} \geq \vthetastart (\vphihat_{t,\Astar_t} - \EE [\mPhi(\mY_t, \Astar_t) \mid \mS_{1:t}, \Astar_t]) + \vthetastart \EE [\mPhi(\mY_t, \Astar_t) \mid \mS_{1:t}, \Astar_t].
\end{equation*}
Rearranging the above display, we have 
\begin{equation}\label{eq:bound-I}
    \vthetastart \EE \left[\mPhi(\mY_t, \Astar_t) \mid \mS_{1:t}, \Astar_t\right] \leq \bar{\vtheta}_t^\top \vphihat_{t,A_t} - \vthetastart \left(\vphihat_{t,\Astar_t} - \EE \left[\mPhi(\mY_t, \Astar_t) \mid \mS_{1:t}, \Astar_t\right]\right). 
\end{equation}

Therefore, for $\regret_t$ as defined in Equation~\eqref{eq:gap-t:def}
\begin{equation}\label{eq:regret-I}
\begin{aligned}
    \regret_t &= \EE \left[R(t, \Astar_t) - R(t, A_t) \mid \mS_{1:t} \right] \\
    &= \vthetastart \EE [\mPhi(\mY_t, \Astar_t) \mid \mS_{1:t}] - \vthetastart \EE \left[\vphihat_{t,A_t} \mid \mS_{1:t}\right] + \vthetastart \EE \left[\vphihat_{t,A_t} \mid \mS_{1:t}\right] - \vthetastart \EE [\mPhi(\mY_t, A_t) \mid \mS_{1:t}]  \\
    &\stackrel{(i)}{\leq}  \EE \left[\left(\bar{\vtheta}_t - \vthetastar\right)^\top \vphihat_{t,A_t} \mid \mS_{1:t} \right]- \vthetastart \left(\EE \left[\vphihat_{t,\Astar_t} \mid \mS_{1:t} \right] - \EE [\mPhi(\mY_t, \Astar_t) \mid \mS_{1:t}]\right) \\
    &\quad+ \vthetastart \left(\EE\left[\vphihat_{t,A_t} \mid \mS_{1:t} \right]- \EE [\mPhi(\mY_t, A_t) \mid \mS_{1:t}]\right)\\
    &\stackrel{(ii)}{=} \EE \left[\left(\bar{\vtheta}_t - \vthetahat_t\right)^\top \vphihat_{t,A_t} \mid \mS_{1:t} \right] + \EE \left[ \left(\vthetahat_t - \vthetastar\right)^\top \vphihat_{t,A_t} \mid \mS_{1:t} \right]\\
    &\quad - \vthetastart \left(\EE \left[\vphihat_{t,\Astar_t} \mid \mS_{1:t} \right]- \EE [\mPhi(\mY_t, \Astar_t) \mid \mS_{1:t}]\right) + \vthetastart \left(\EE \left[\vphihat_{t,A_t} \mid \mS_{1:t}\right] - \EE [\mPhi(\mY_t, A_t) \mid \mS_{1:t}]\right). 
\end{aligned}
\end{equation}
where inequality $(i)$ follows from Equation~\eqref{eq:bound-I} and equality $(ii)$ follows from adding and subtracting $\vthetahat_t^\top \vphihat_{t,A_t}$, where $\vthetahat_t$ is defined in Equation~\eqref{eq:thetahat:define}. 

Recall $\mSigma_t$ defined in Equation~\eqref{eq:Sigt:define}. 
We claim that for any $\vtheta \in \BALL_{t-1}$ and any $\vphi \in \R^d$, 
\begin{equation} \label{eq:bound-II}
    \left|(\vtheta - \vthetahat_t)^\top \vphi\right| \leq \sqrt{\gamma_t \vphi^\top \mSigma_t^{-1} \vphi}. 
\end{equation}
To see this, by Cauchy-Schwarz inequality, we have 
\begin{equation*}
\begin{aligned}
    \left|(\vtheta - \vthetahat_t)^\top \vphi\right| &= \left|(\vtheta - \vthetahat_t)^\top \mSigma^{1/2}_t \mSigma^{-1/2}_t \vphi\right| \leq \left\|\vtheta - \vthetahat_t\right\|_{\mSigma_t} \left\|\vphi\right\|_{\mSigma_t^{-1}} \leq \sqrt{\gamma_t \vphi^\top \mSigma_t^{-1} \vphi},
\end{aligned}
\end{equation*}
where the last inequality follows from the fact that $\vtheta \in \BALL_{t-1}$ and the choice of $\gamma_t$ in Equation~\eqref{eq:ball:define}. 
Applying the above display with $\vtheta \in \{\vthetastar, \bar{\vtheta}_t\}$ and $\vphi = \vphihat_{t,A_t}$ yields 
\begin{equation} \label{eq:gap_t:bound:I}
\begin{aligned}
    \left|(\bar{\vtheta}_t - \vthetahat_t)^\top \vphihat_{t,A_t}\right| + \left|(\vthetahat_t - \vthetastar)^\top \vphihat_{t,A_t}\right| 
    &\leq 2\sqrt{\gamma_t \vphihat_{t,A_t}^\top \mSigma_t^{-1} \vphihat_{t,A_t}}.
\end{aligned}
\end{equation}

Let
\begin{equation}\label{eq:xi1:def}
    \xi_{1,t} := \min \left\{ \sqrt{\gamma_t \vphihat_{t,A_t}^\top \mSigma_t^{-1} \vphihat_{t,A_t}}, 1 \right\}. 
\end{equation}

For any $a \in \cA$ and $t \in [T]$, let 
\begin{equation}\label{eq:xi2t:def}
    \xi_{2,t} = \max_{a \in \calA} \left|\vthetastart \left(\vphihat_{t,a} - \EE\left[\mPhi\left(\mY_t, a\right) \mid \mS_{1:t}\right] \right) \right|.
\end{equation}
We have 
\begin{equation} \label{eq:xi2:bound}
\begin{aligned}
    \xi_{2,t} \leq \max_{a\in\calA} \left|\vthetastart \left(\vphihat_{t,a} - \EE [\mPhi(\mY_t, a) \mid \mS_{1:t}]\right)\right| &\leq \left\|\vthetastar\right\|_2 \max_{a\in\calA} \left\|\vphihat_{t,a} - \EE [\mPhi(\mY_t, a) \mid \mS_{1:t}]\right\|_2\\
    &\leq \sqrt{d D_t} 
\end{aligned}
\end{equation}
where the first inequality follows from Cauchy-Schwarz inequality and the last inequality follows from the assumption that $\|\vthetastar\|_2 \leq 1$ and Equation~\eqref{eq:Dt:def}.

Taking Equations~\eqref{eq:gap_t:bound:I}, \eqref{eq:xi1:def} and~\eqref{eq:xi2t:def} into Equation~\eqref{eq:regret-I}, we have
\begin{equation}\label{eq:gap_t:bound:II}
\begin{aligned}
    \left|\regret_t\right| &= \min\left\{\left|\regret_t\right|, 1\right\} \leq 2 \EE \left[\xi_{1,t} \mid \mS_{1:t} \right] + 2\xi_{2,t},
\end{aligned}
\end{equation}
where the first equality follows from the assumption that $R(t,a) \in [-1,1]$ for any $a \in \calA$. 
Summing Equation~\eqref{eq:gap_t:bound:II} over $t \in [T]$ gives 
\begin{equation} \label{eq:gap_t:bound:III}
\begin{aligned}
        \sum_{t=1}^T \regret_t &\leq 2\sum_{t=1}^T \EE \left[\xi_{1,t} \mid \mS_{1:t}\right] + 2\sum_{t=1}^T \xi_{2,t} \\
        &\stackrel{(i)}{\leq} 2\sqrt{T \sum_{t=1}^T \EE \left[\xi_{1,t}^2 \mid \mS_{1:t}\right] } + 2\sum_{t=1}^T \xi_{2,t} \\
        &\stackrel{(ii)}{\leq} 2\sqrt{2 T\gamma_T d \log \left(1 + \frac{T B^2}{d \lambda}\right)} +  2\sum_{t=1}^T \sqrt{d D_t}
\end{aligned} 
\end{equation}
where inequality $(i)$ follows from Cauchy-Schwarz inequality and inequality $(ii)$ follows from Equation~\eqref{eq:xi1:bound} in Lemma~\ref{lem:xi1t:bound} and Equation~\eqref{eq:xi2:bound}. 

It remains to choose a sequence of suitable $\{\gamma_t\}_{t=1}^T$ so that we have $\vthetastar \in \BALL_{t-1}$ for all $t \in [T]$ with high probability.   
At time $t \in [T]$, we have  
\begin{equation} \label{eq:rt:decomposition}
\begin{aligned}
    R_t &= \vthetastart \vphihat_{t,A_t} + \vthetastart \left(\EE [\mPhi(\mY_t, A_t) \mid \mS_{1:t}, A_t] - \vphihat_{t,A_t}\right) - \vthetastart \left(\EE [\mPhi(\mY_t, A_t) \mid \mS_{1:t}, A_t] - \mPhi(\mY_t, A_t)\right) + \eta_t\\
    &= \vthetastart \vphihat_{t,A_t} + \vthetastart \left(\EE [\mPhi(\mY_t, A_t) \mid \mS_{1:t}, A_t] - \vphihat_{t,A_t}\right) + \varepsilon_t + \eta_t\\
\end{aligned}    
\end{equation}
where the first equality follows from the definition of $R_t$ in Equation~\eqref{eq:Rta:define} and the second equality follows from Equation~\eqref{eq:vareps:define}
By the definition of $\vthetahat_t$ given in Equation~\eqref{eq:thetahat:define}, it follows that 
\begin{equation} \label{eq:thetahat:error}
\begin{aligned}
    \vthetahat_t - \vthetastar &= \mSigma_t^{-1} \sum_{\tau=1}^{t} R_{\tau} \vphihat_{\tau, A_{\tau}} - \vthetastar \\
    &= \left[\mSigma_t^{-1} \left(\sum_{\tau=1}^{t} \vphihat_{\tau, A_{\tau}} \vphihat_{\tau, A_{\tau}}^\top\right) - 1\right] \vthetastar + \mSigma_t^{-1} \sum_{\tau=1}^{t} \vphihat_{\tau, A_{\tau}} \left(\EE \left[\mPhi(\mY_\tau, A_\tau) \mid \mS_{1:\tau}, A_\tau\right] - \vphihat_{\tau,A_{\tau}}\right)^\top \vthetastar \\
    &\qquad + \mSigma_t^{-1} \sum_{\tau=1}^{t} \vphihat_{\tau, A_{\tau}} (\eta_\tau+\varepsilon_\tau)\\
    &= -\lambda \mSigma_t^{-1} \vthetastar + \mSigma_t^{-1} \sum_{\tau=1}^{t} \vphihat_{\tau,A_{\tau}} \left(\EE \left[\mPhi(\mY_\tau, A_\tau) \mid \mS_{1:\tau}, A_\tau\right] - \vphihat_{\tau,A_{\tau}}\right)^\top \vthetastar + \mSigma_t^{-1} \sum_{\tau=1}^{t} \vphihat_{\tau,A_{\tau}} (\eta_\tau + \varepsilon_\tau)\\
\end{aligned}
\end{equation}
where the first equality follows from Equation~\eqref{eq:rt:decomposition} and the last equality follows from the definition of $\Sigma_t$ in Equation~\eqref{eq:Sigt:define}. 

Compared to standard analysis of vanilla $\LinUCB$, the only different term is that we have an extra term 
\begin{equation} \label{eq:to-control}
\mSigma_t^{-1} \sum_{\tau=1}^{t} \vphihat_{\tau,A_{\tau}} \left(\EE \left[\mPhi(\mY_\tau, A_\tau) \mid \mS_{1:\tau}, A_\tau\right] - \vphihat_{\tau,A_\tau}\right)^\top \vthetastar.  
\end{equation}
Following the same analysis as Equation~\eqref{eq:xi2:bound}, we arrive at 
\begin{equation} \label{eq:labeled:bound}
    \left|\left(\EE [\mPhi(\mY_\tau, A_\tau) \mid \mS_{1:\tau}, A_\tau] - \vphihat_{\tau,A_\tau}\right)^\top \vthetastar\right| \leq \left\|\EE [\mPhi(\mY_\tau, A_\tau)\mid \mS_{1:\tau}, A_\tau] - \vphihat_{\tau,A_{\tau}}\right\|_2 \leq \sqrt{d D_t}.
\end{equation}
To control Equation~\eqref{eq:to-control}, we have 
\begin{equation} \label{eq:crude:bound}
\begin{aligned}
    &\left|\left(\sum_{\tau=1}^{t} \vphihat_{\tau, A_{\tau}}^\top (\EE [\mPhi(\mY_\tau, A_\tau) \mid \mS_{1:\tau}, A_\tau] - \vphihat_{\tau,A_{\tau}})^\top \vthetastar\right) \mSigma_t^{-1} \left(\sum_{\tau=1}^{t} \vphihat_{\tau, A_\tau} (\EE [\mPhi(\mY_\tau, A_\tau) \mid \mS_{1:\tau}, A_\tau] - \vphihat_{\tau,A_{\tau}})^\top \vthetastar\right)\right| \\
    =&~ \left\|\sum_{\tau=1}^t \mSigma_{t}^{-1/2} \vphihat_{\tau, A_{\tau}} (\EE [\mPhi(\mY_\tau, A_\tau) \mid \mS_{1:\tau}, A_\tau] - \vphihat_{\tau,A_{\tau}})^\top \vthetastar\right\|_2^2 \\
    \stackrel{(i)}{\leq}&~ \left(\sum_{\tau=1}^t \left\|\mSigma_{t}^{-1/2} \vphihat_{\tau, A_{\tau}} (\EE [\mPhi(\mY_\tau, A_\tau) \mid \mS_{1:\tau}, A_\tau] - \vphihat_{\tau,A_{\tau}})^\top \vthetastar\right\|_2\right)^2\\
    \stackrel{(ii)}{\leq}&~ \left(\sum_{\tau=1}^t \left[\left(\EE [\mPhi(\mY_\tau, A_\tau) \mid \mS_{1:\tau}, A_\tau] - \vphihat_{\tau,A_{\tau}}\right)^\top \vthetastar\right]^2\right) \left(\sum_{\tau=1}^t \left\|\mSigma_{t}^{-1/2} \vphihat_{\tau, A_{\tau}}\right\|_2^2\right)\\
    \stackrel{(iii)}{\leq} &~ d \left(\sum_{\tau=1}^t D_{\tau}\right) \left(\sum_{\tau=1}^{t} \vphihat^\top_{\tau,A_{\tau}} \mSigma_{t}^{-1} \vphihat_{\tau,A_{\tau}}\right)
\end{aligned}
\end{equation}
where inequality $(i)$ follows from the triangle inequality, inequality $(ii)$ follows from Cauchy-Schwarz inequality and inequality $(iii)$ follows from Equation~\eqref{eq:labeled:bound}. 
Using properties of the trace operator, we continue to bound the right-hand side of Equation~\eqref{eq:crude:bound} using
\begin{equation} \label{eq:differ-term-bound}
\begin{aligned}
     d\left(\sum_{\tau=1}^t D_{\tau}\right) \left(\sum_{\tau=1}^{t} \vphihat^\top_{\tau,A_{\tau}} \mSigma_{t}^{-1} \vphihat_{\tau,A_{\tau}}\right) &= d\left(\sum_{\tau=1}^t D_{\tau}\right) \cdot \tr\left(\mSigma_t^{-1} \sum_{\tau=1}^{t} \vphihat_{\tau,A_{\tau}} \vphihat^\top_{\tau,A_{\tau}}\right)\\
    &\stackrel{(i)}{=} d\left(\sum_{\tau=1}^t D_{\tau}\right) \left(d - \lambda \tr\left(\mSigma^{-1}_t\right)\right) \leq d^2 \left(\sum_{\tau=1}^t D_{\tau}\right)
\end{aligned}
\end{equation}
where equality $(i)$ follows from the definition of $\mSigma_t$ as given in Equation~\eqref{eq:Sigt:define}.
Taking Equation~\eqref{eq:differ-term-bound} into Equation~\eqref{eq:crude:bound} yields that 
\begin{equation} \label{eq:differ-term-bound-final}
    \left\|\sum_{\tau=1}^{t} \vphihat_{\tau, A_{\tau}}^\top \left(\EE \left[\mPhi(\mY_\tau, A_\tau) \mid \mS_{1:t}, A_\tau\right] - \vphihat_{\tau,A_{\tau}}\right)^\top \vthetastar \right\|_{\Sigma_t^{-1}} \leq d \sqrt{\sum_{\tau=1}^t D_{\tau}}
\end{equation}

Therefore, using standard self-normalization concentration inequalities (see Lemma~\ref{lem:self-norm-bound}) with Equations~\eqref{eq:thetahat:error} and~\eqref{eq:differ-term-bound-final}, with probability at least $1 - \delta_{t}$, 
\begin{equation*}
\begin{aligned}
    \left\|\vthetahat_t - \vthetastar\right\|_{\mSigma_t}
    &\leq \sqrt{\lambda} \|\vthetastar\|_{\mSigma^{-1}_t} \!+ \left\|\sum_{\tau=1}^t \vphihat_{\tau, A_{\tau}} (\eta_\tau + \varepsilon_\tau) \right\|_{\mSigma_t^{-1}} \!\!\!\!+ \left\|\sum_{\tau=1}^{t} \vphihat_{\tau, A_{\tau}}^\top \!\left(\EE [\mPhi(\mY_\tau, A_\tau) \mid \mS_{1:\tau}, A_\tau] - \vphihat_{\tau,A_{\tau}}\right)^\top \!\vthetastar \right\|_{\mSigma_t^{-1}} \\
    &\leq \sqrt{\lambda} + (\sigma_{\eta} + \sigma_{\varepsilon})\sqrt{2 \log (\det(\mSigma_t)\det(\mSigma_1)^{-1} / \delta_t)} + d \sqrt{\sum_{\tau=1}^t D_{\tau}} \\
    &\leq \sqrt{\lambda} + (\sigma_{\eta} + \sigma_{\varepsilon})\sqrt{2 \log \left[\left(1 + \frac{t B^2}{d \lambda}\right)^d/\delta_t\right]} + d \sqrt{\sum_{\tau=1}^t D_{\tau}} 
\end{aligned}
\end{equation*}
where the last inequality follows from Equation~\eqref{eq:logdet:bound}. 
It suffices to set $\delta_t := \delta (3/\pi^2)/t^2$.
Hence, by taking $\gamma_t$ as defined in Equation~\eqref{eq:gamma_t:define}, with probability at least $1 - \delta$, we have 
\begin{equation*}
    \left\|\vthetahat_t - \vthetastar\right\|_{\mSigma_t^{-1}}^2 \leq \gamma_t
\end{equation*}
holds for all $t \in [T]$.
It follows from Equation~\eqref{eq:gap_t:bound:III} that $\sum_{t=1}^T \regret_t$ is bounded by
\begin{equation*}
    2\sum_{t=1}^T \sqrt{d D_t} + 2\sqrt{6 T \left(\sum_{t=1}^T D_t\right) d^3 \log \left(1 + \frac{T B^2}{d \lambda}\right)} + 2\sqrt{2 \gamma^{(0)}_T T d \log\left(1 + \frac{T B^2}{d \lambda}\right)}
\end{equation*}
where we use the naive bound $\sqrt{a+b} \leq \sqrt{a} + \sqrt{b}$. 
Applying Cauchy-Schwarz inequality to $\sum_{t=1}^T \sqrt{d D_t}$ yields that 
\begin{equation*}
\begin{aligned}
    \sum_{t=1}^T \regret_t &\leq 2 \sqrt{d T \sum_{t=1}^T D_t} + 2\sqrt{6 T \left(\sum_{t=1}^T D_t\right) d^3 \log \left(1 + \frac{T B^2}{d \lambda}\right)} + 2\sqrt{2 \gamma^{(0)}_T T d \log\left(1 + \frac{T B^2}{d \lambda}\right)}\\
    &\leq 4\sqrt{6 T \left(\sum_{t=1}^T D_t\right) d^3 \log \left(1 + \frac{T B^2}{d \lambda}\right)} + 2\sqrt{2 \gamma^{(0)}_T T d \log\left(1 + \frac{T B^2}{d \lambda}\right)}\\
\end{aligned}
\end{equation*}
as desired. 
\end{proof}

\subsection{Technical Lemmas}

\begin{lemma} \label{lem:xi1t:bound}
    For any $t \in [T]$ and $\xi_{1,t}$ defined in Equation~\eqref{eq:xi1:def}, under the same conditions as Theorem~\ref{thm:regret:upper:bound}, we have 
    \begin{equation} \label{eq:xi1:bound}
\begin{aligned}
     \sum_{t=1}^T \xi_{1,t}^2 
    &\leq 2 \gamma_T d \log\left(1+\frac{T B^2}{d \lambda}\right)
\end{aligned}
\end{equation}
\end{lemma}
\begin{proof}
    For $\gamma_t \geq 1$, by the definition of $\xi_{1,t}$ in the above display,
\begin{equation}\label{eq:bound:III}
\begin{aligned}
     \sum_{t=1}^T \xi_{1,t}^2 &\leq \sum_{t=1}^T \gamma_t \min\left\{\vphihat_{t,A_t}^\top \mSigma_t^{-1} \vphihat_{t,A_t}, 1\right\}
\end{aligned}
\end{equation}
To control Equation~\eqref{eq:bound:III}, we use the potential function bound. 
We include a brief proof here for completeness. 
By the definition of $\mSigma_{t+1}$ in Equation~\eqref{eq:Sigt:define}, we have
\begin{equation} \label{eq:step-i}
\begin{aligned}
    \det \mSigma_{t+1} &= \det\left(\mSigma_t + \vphihat_{t,A_t} \vphihat^\top_{t,A_t}\right) = \det(\mSigma_t) \det\left(\mI + \mSigma^{-1/2}_t \vphihat_{t,A_t} \left(\mSigma_t^{-1/2} \vphihat_{t,A_t}\right)^\top\right) \\
    &= \det(\mSigma_t) \left(1 + \vphihat^\top_{t,A_t} \mSigma_t^{-1} \vphihat_{t, A_t}\right),
\end{aligned}
\end{equation}
where the last equality follows from Sylvester's determinant theorem. 
By induction, it is straightforward to show that 
\begin{equation*}
    \det \mSigma_T = \det \mSigma_0 \prod_{t=1}^{T} \left(1 + \vphihat_{t,A_t}^\top \mSigma^{-1}_t \vphihat_{t,A_t}\right), 
\end{equation*}
following Equation~\eqref{eq:step-i}.
Rearranging terms and taking logarithm on both sides of the above display implies that 
\begin{equation} \label{eq:logdet:bound}
    \sum_{t=1}^T \log\left(1 + \vphihat_{t,A_t}^\top \mSigma_t^{-1} \vphihat_{t,A_t}\right) = \log\left(\frac{\det \mSigma_T}{\det \mSigma_0}\right) \leq 2 \gamma_T d \log\left(1+\frac{T B^2}{d \lambda}\right)
\end{equation}
where the last inequality follows from Assumption~\ref{assump:phi:bound} and the potential function bound in Lemma~\ref{lem:potential}. 
Hence, applying the above display to Equation~\eqref{eq:bound:III}
\begin{equation*}
\begin{aligned}
    \sum_{t=1}^{T} \gamma_t \min\left\{\vphihat_{t,A_t}^\top \mSigma_t^{-1} \vphihat_{t,A_t}, 1\right\} &\stackrel{(i)}{\leq} 2\gamma_T \sum_{t=1}^T \log\left(1 + \vphihat_{t,A_t}^\top \mSigma_t^{-1} \vphihat_{t,A_t}\right)
    = 2\gamma_{T} \log\left(\frac{\det \mSigma_{T}}{\det \mSigma_0}\right) \\
    &\leq 2 \gamma_T d \log\left(1+\frac{T B^2}{d \lambda}\right).
\end{aligned}
\end{equation*}
where inequality $(i)$ follows from $\log(1+y) \geq y/2$ for all $y \in [0,1]$. 
Taking the above display into Equation~\eqref{eq:bound:III} yields the desired bound as in Equation~\eqref{eq:xi1:bound}. 
\end{proof}

\begin{lemma} \label{lem:self-norm-bound}[Self-Normalized Bound for Vector-Valued Martingales]
    Let $\{\calF_t\}_{t=0}^\infty$ be a filtration. 
    Let $\{\eta_t\}_{t=1}^{\infty}$ be a real-valued stochastic process such that $\eta_t$ is $\calF_t$-measurable and $\eta_t$ is conditionally $R$-sub-Gaussian for some $R\geq 0$. 
    Let $\{\mX_t\}_{t=1}^\infty$ be an $\R^d$-valued stochastic process such that $\mX_t$ is $\calF_{t-1}$-measurable. Assume that $\mV$ is a $d\times d$ positive definite matrix.
    For any $t \geq 0$, define
    \begin{equation*}
        \bar{\mV}_t = \mV + \sum_{s=1}^t \mX_s \mX^{\top}_s \quad S_t = \sum_{s=1}^t \eta_s \mX_s. 
    \end{equation*}
    Then, for any $\delta > 0$, with probability at least $1-\delta$, for all $t \geq 0$,
    \begin{equation*}
        \left\|S_t\right\|^2_{\bar{\mV}_t^{-1}} \leq 2 R^2 \log \left(\frac{\det(\bar{\mV}_t)^{1/2} \det(\mV)^{-1/2}}{\delta}\right). 
    \end{equation*}
\end{lemma}
\begin{proof}
    See Theorem 1 in \cite{abbasi2011improved}. 
\end{proof}

\begin{lemma}[Potential Function Bound]\label{lem:potential}
    For any sequence $\vx_0, \ldots \vx_{T-1}$ such that, for $t<T,\left\|\vx_t\right\|_2 \leq B$, we have
    $$
\log \left(\operatorname{det} \mSigma_{T-1} / \operatorname{det} \mSigma_0\right)=\log \operatorname{det}\left(\mI+\frac{1}{\lambda} \sum_{t=0}^{T-1} \vx_t \vx_t^{\top}\right) \leq d \log \left(1+\frac{T B^2}{d \lambda}\right),
$$
where $\mSigma_t=\lambda \mI+\sum_{\tau=0}^{t-1} \vx_\tau \vx_\tau^{\top} \text { with } \mSigma_0=\lambda \mI$ for any $\lambda > 0$. 
\end{lemma}
\begin{proof}
    See Lemma 6.11 in \cite{agarwal2019reinforcement}. 
\end{proof}

\section{Proof of Results in Section~\ref{subsec:nonparam:ex}}
\label{propoproof}
\subsection{Proof of Proposition~\ref{cor:exp:lin}}

\begin{proof}
Let $\bb = \left(\vbeta_0^\top, \vbeta_1^\top, \cdots, \vbeta_m^\top\right) \in \R^{(m+1)d_S}$. 
A standard analysis of the OLS estimator $\vbhat$ yields that 
\begin{equation} \label{eq:b-error}
    \EE_{\calD} \left[\left\|\bb - \vbhat\right\|_2^2\right] \lesssim \frac{m d_S}{N T_0},
\end{equation}
where the expectation is taken with respect to the historical data $\calD$.
We omit the details for brevity.
For a new copy $\mY_t = (W_t, \mS_t)$ independent of the historical data $\calD$, since 
\begin{equation} \label{eq:Wt:mu}
    \mu_t := \EE \left[W_t \mid \mS_{1:t}\right] = \sum_{j=0}^{m} \vbeta_{j}^\top \mS_{t-j} 
\end{equation}
and 
\begin{equation*}
    \Var\left(W_t \mid \mS_{1:t}\right) = \Var\left(\xi_t \mid \mS_{1:t}\right) = 1,
\end{equation*}
we have $W_t \mid \mS_{1:t} \sim \calN(\mu_t, 1)$. 
The imputed $W_t$ is then given by
\begin{equation*}
    \What_t = \sum_{j=0}^{m} \vbetahat^\top_j \mS_{t-j} + \xi_t
\end{equation*}
and it follows that $\What_t \mid \mS_{1:t} \sim \calN(\muhat_t, 1)$, where 
\begin{equation}\label{eq:Wt:muhat:sighat}
    \muhat_t := \sum_{j=0}^{m} \vbetahat^\top_j \mS_{t-j}
\end{equation}

It follows that 
\begin{equation}\label{eq:Dt:gauss}
\begin{aligned}
    \sqrt{D_t} &= \sqrt{\frac{1}{2} \operatorname{KL} \left(\calN(\mu_t, 1) \| \calN(\muhat_t, 1) \right)} = \frac{1}{2} \left|\mu_t - \muhat_t\right|\\
    &= \frac{1}{4} \left|\sum_{j=0}^{m} \left(\vbetahat_j - \vbeta_j\right)^\top \mS_{t-j}\right|\\
\end{aligned}
\end{equation}
where the last equality follows from the definition of $\mu_t$ and $\muhat_t$ in Equations~\eqref{eq:Wt:mu}. 
Since $\mS_{t-j}$ and $\vbetahat_j$ are independent, conditioned on $\vbetahat_j$, we have 
\begin{equation*}
    \sum_{j=0}^{m} \left(\vbetahat_j - \vbeta_j\right)^\top \mS_{t-j} \sim \calN\left(0, \sum_{j=0}^m \left\|\vbetahat_j - \vbeta_j\right\|^2_{2} \right).
\end{equation*}
Combining the above display with Equation~\eqref{eq:Dt:gauss} yields that 
\begin{equation*}
\begin{aligned}
    \EE \left[\sqrt{D_t}\right] &= \frac{1}{4} \EE \left|\sum_{j=0}^{m} \left(\vbetahat_j - \vbeta_j\right)^\top \mS_{t-j}\right|\\
    &= \frac{\pi}{8} \EE_\calD \sqrt{\sum_{j=0}^m \left\|\vbetahat_j - \vbeta_j\right\|^2_{2}} \lesssim \sqrt{\frac{m d_S}{N T_0}}, 
\end{aligned}
\end{equation*}
where the last equality follows from Equation~\eqref{eq:b-error}. 

It then follows from Theorem~\ref{thm:regret:upper:bound} that 
\begin{equation} \label{eq:linear:ERT}
\begin{aligned}
    \EE \left[\Regret_T\right] &\leq \delta T + \EE \regret_T^{(\texttt{imp})} + \regret_T^{(\lin)} \\
    &\lesssim \delta T + \sqrt{\gamma^{(0)}_T T d \log\left(1 + \frac{T B^2}{d \lambda}\right)} + \EE \sqrt{T \left(\sum_{t=1}^T D_t\right) d^3 \log \left(1 + \frac{T B^2}{d \lambda}\right)} \\
    &\lesssim \delta T + \sqrt{\gamma^{(0)}_T T d \log\left(1 + \frac{T B^2}{d \lambda}\right)} + T \sqrt{\frac{m d_S d^3}{N T_0} \log \left(1 + \frac{T B^2}{d \lambda} \right)}
\end{aligned}
\end{equation}
Taking $\delta = T^{-1/2}$, we have 
\begin{equation*}
    \gamma_T^{(0)} = 3\lambda + 6(\sigma_{\eta} + 2)^2 \log \left[4T^{5/2} \left(1 + \frac{T B^2}{d \lambda}\right)^d\right] \asymp d \log T + d\log \left(1 + \frac{T B^2}{d \lambda}\right)
\end{equation*}
Taking the above display into Equation~\eqref{eq:linear:ERT} yields the desired result. 
\end{proof}

\subsection{Proof of Proposition~\ref{cor:exp:non}}

\begin{proof}
    From the classical nonparametric statistics literature, there exists an estimator $\hat{f}$ \citep[such as the kernel estimator, see Chapter 1 of][]{tsybakov2008introduction} of $f$ that satisfies
\begin{equation} \label{eq:nonparam:upper}
    \EE_{\calD} \left[\left\|\hat{f} - f\right\|_{L_2}^2\right] \lesssim N^{-\frac{2\beta}{2\beta+d_S}},
\end{equation}
where the expectation is taken with respect to the historical data $\calD$. 
For any pair $(\mS_t, W_t)$ independent of the historical data $\calD$, where $\mS_t \sim \operatorname{Unif}([0,1]^{d_S})$, one has
\begin{equation*}
\begin{aligned}
    \KL\left(\PP_f\left(W_t \mid \mS_t\right) \| \PP_{\hat{f}}\left(W_t \mid \mS_t\right)\right) 
    &= \EE_{\mS_t} \left[\KL\left(\calN(f(\mS_t), 1) \| \calN(\hat{f}(\mS_t), 1)\right) \mid \hat{f} \right] \\
    &= \frac{1}{2} \EE_{\mS_t}\left[\left(\hat{f}(\mS_t) - f(\mS_t)\right)^2\right] = \frac{1}{2} \left\|\hat{f} - f\right\|^2_{L_2}.
\end{aligned}
\end{equation*}
Taking expectation over the historical data and combining Equation~\eqref{eq:nonparam:upper} with the above display, we have 
\begin{equation*}
    \EE_{\calD} \KL\left(\PP_f\left(W_0 \mid \mS_0\right), \PP_{\hat{f}}\left(W_0 \mid \mS_0\right)\right) \lesssim N^{-\frac{2\beta}{2\beta + d_S}}. 
\end{equation*}
Recall the definition of $D_t$ in Equation~\eqref{eq:Dt:def0}, it follows that 
\begin{equation*}
\begin{aligned}
    \EE \sqrt{D_t} 
    &\asymp \EE_{\mS_t} \sqrt{\KL\!\left(\PP_f\left(\mY_t|\mS_{t}\!=\!\vs_{t}\right) \!\| \PP_{\hat{f}}\left(\mY_t|\mS_{t}\!=\!\vs_{t}\right)\right)}
    \lesssim \EE_{\mS_t} \left\|\hat{f} - f\right\|_{L_2} \lesssim N^{-\frac{\beta}{2\beta + d_S}}. 
\end{aligned}
\end{equation*}
Combining the above display with Theorem~\ref{thm:regret:upper:bound} yields that  
\begin{equation*}
\begin{aligned}
    \EE \left[\Regret_T\right]
    &\lesssim T\sqrt{d^3 \log \left(1 + \frac{T B^2}{d\lambda}\right)} N^{-\frac{\beta}{2\beta + d_S}} + \regret_T^{(\lin)} + \delta T.
\end{aligned}
\end{equation*}
Taking $\delta = T^{-1/2}$ and following a similar proof of Proposition~\ref{cor:exp:lin} yields the desired result. 
\end{proof}

\subsection{Robustness of Propositions~\ref{cor:exp:lin} and~\ref{cor:exp:non} to Covariate Shift}
\label{app:covariate-shift-theory}

The proofs above assume that the online contexts $\mS_t$ follow the same marginal distribution as the historical data. Here we show that the rates for $\EE[\sqrt{D_t}]$ in Propositions~\ref{cor:exp:lin} and~\ref{cor:exp:non} are preserved when the marginal distribution of $\mS_t$ shifts, provided the conditional mechanism $W_t \mid \mS_{1:t}$ remains invariant and the marginal shift is mild.

Consider the linear model of Proposition~\ref{cor:exp:lin} and suppose the online contexts are independent of the historical dataset $\calD$ and satisfy $\mS_t \sim \calN(\boldsymbol{0}, \boldsymbol{\Sigma}_t)$ with a possibly time-varying covariance. The key step in the proof (see Equation~\eqref{eq:Dt:gauss}) involves bounding the distribution of the estimation error $\sum_{j=0}^{m} (\widehat{\vbeta}_j - \vbeta_j)^\top \mS_{t-j}$. Under the shifted marginal, this term is distributed as
\[
\sum_{j=0}^{m} (\widehat{\vbeta}_j - \vbeta_j)^\top \mS_{t-j}
\;\sim\;
\calN\!\left(
    0,\;
    \sum_{j=0}^m
    (\widehat{\vbeta}_j - \vbeta_j)^\top
    \boldsymbol{\Sigma}_{t-j}
    (\widehat{\vbeta}_j - \vbeta_j)
\right),
\]
and the same rates for $\EE[\sqrt{D_t}]$ continue to hold as long as $\|\boldsymbol{\Sigma}_t\|_{\mathrm{op}} = O(1)$. Thus, even if the marginal distribution of $\mS_t$ shifts in the online setting, the regret guarantees of Proposition~\ref{cor:exp:lin} remain valid under uniformly bounded second moments. An analogous argument applies to Proposition~\ref{cor:exp:non}: the nonparametric rate depends on the $L_2$ estimation error $\|\hat{f} - f\|_{L_2}^2$, which is invariant to the online marginal as long as the conditional mechanism $W_t \mid \mS_t$ is unchanged.

More broadly, the key condition is that the conditional mechanism $W_t \mid \mS_{1:t}$ is invariant across the historical and online phases and the marginal shift in $\mS_{1:t}$ is mild (e.g., bounded moments). Under these conditions, the rates in Section~\ref{subsec:nonparam:ex} remain unchanged. Investigating more general covariate-shift models, such as transfer-exponent conditions between historical and online distributions \citep{cai2024transfer}, is an interesting direction that requires substantially more technical care and is left for future work.

\section{Extension: A Confidence-Weighted Variant of PULSE-UCB}
\label{app:robust_variant}



In the main text, we employ the standard OFUL algorithm in the online stage of PULSE-UCB as a simple, well-understood baseline. As demonstrated in Theorems \ref{thm:regret:upper:bound} and \ref{thm:lower:informal}, this plug-in approach already achieves near-optimal regret for our primary settings. However, since the context recovery step inherently introduces an imputation error $D_t$ for the feature at each step, one might ask whether we can make further improvements to mitigate this residual error.

To explore this, we draw inspiration from the literature on linear contextual bandits with corrupted rewards \citep{he2022nearly} to propose an alternative algorithmic variant. It is important to note that this variant is not a direct application of \citet{he2022nearly}; their work focuses on mitigating adversarial corruptions on the rewards, whereas our setting deals with misspecification in the contexts due to offline imputation. By adapting their confidence-weighted mechanism to our missing context framework, this alternative variant can potentially yield tighter regret bounds. Crucially, this improvement is conditional: it requires a reliable estimator for the total imputation error budget $\bar{C}$ a priori, and it is primarily beneficial when the imputation error $D_t$ varies substantially across time.

\subsection{Algorithm Design: Robust PULSE-UCB}
The primary modification in this robust variant lies in the construction of the confidence sets and the weighted regularized least-squares estimator. Specifically, we select actions via the modified confidence set $\widetilde{\texttt{BALL}}_t = \{\theta : \|\theta - \tilde\theta_{t-1}\|_{\tilde\Sigma_{t-1}}^2 \le \tilde\gamma_{t-1}\}$, where the covariance matrix and the parameter estimator are updated using adaptive weights:
$$
\tilde\Sigma_t = \lambda I + \sum_{\tau=1}^t w_\tau \hat\phi_{\tau, a_\tau} \hat\phi_{\tau, a_\tau}^\top, \quad \tilde\theta_t = \tilde\Sigma_t^{-1} \sum_{\tau=1}^t w_\tau \hat\phi_{\tau, a_\tau} R_\tau.
$$
The adaptive weight $w_t$ for each step is defined as:
$$
w_t = \min\left\{ 1, \frac{\alpha}{\|\hat\phi_{t, a_t}\|_{\tilde\Sigma_{t-1}^{-1}}} \right\},
$$
where $\alpha > 0$ is a tuning parameter.

\subsection{Theoretical Guarantees and Regret Comparison}

We establish the theoretical guarantee for this robust variant as follows:

\begin{proposition}[Regret of Robust PULSE-UCB]
\label{prop:robust_pulse}
Let $\bar{C}$ be any known upper bound such that $\bar{C} \ge \sqrt{d}\sum_{t=1}^T \sqrt{D_t}$. By setting the tuning parameter $\alpha = \sqrt{d} / \bar{C}$ and the confidence radius $\tilde\gamma_t = \gamma_t^{(0)} + 3\alpha^2 d \big(\sum_{t=1}^T \sqrt{D_t}\big)^2$, the cumulative regret of the robust variant satisfies, with high probability:
$$
R_T \le R_T^{(lin)} + \tilde R_T^{(imp)},
$$
where $R_T^{(lin)}$ matches the standard linear bandit regret term in Theorem 4.1, and the modified imputation penalty term is bounded by:
$$
\tilde R_T^{(imp)} = 12d\log^{3/2}\left(1+\frac{TB^2}{d\lambda}\right)\bar{C}.
$$
\end{proposition}

\begin{proof}
Let $\phi_{t,a}$ denote the true feature vector and $\hat{\phi}_{t,a}$ denote the imputed feature vector provided by the pretrained model. We can express the observed reward at round $t$ by explicitly isolating the stochastic noise and the context imputation error:
$$
R_t = \langle \theta^\star, \phi_{t, a_t} \rangle + \eta_t = \langle \theta^\star, \hat{\phi}_{t, a_t} \rangle + c_t + \eta_t,
$$
where the adversarial corruption term induced by imputation is defined as $c_t = \langle \theta^\star, \phi_{t, a_t} - \hat{\phi}_{t, a_t} \rangle$. Assuming the unknown environment parameter is bounded by $\|\theta^\star\|_2 \le S$ and the feature norm is bounded by $L$, the magnitude of this corruption is bounded by $|c_t| \le S \|\phi_{t, a_t} - \hat{\phi}_{t, a_t}\|_2 \le S \sqrt{D_t}$. Consequently, the total cumulative corruption over the horizon $T$ is $C = \sum_{t=1}^T |c_t| \le S \sum_{t=1}^T \sqrt{D_t}$.

We first establish the concentration of the estimation error. The weighted ridge regression estimator $\tilde{\theta}_t$ admits the closed-form representation:
$$
\tilde{\theta}_t = \tilde{\Sigma}_t^{-1} \sum_{\tau=1}^t w_\tau \hat{\phi}_{\tau, a_\tau} (\langle \theta^\star, \hat{\phi}_{\tau, a_\tau} \rangle + c_\tau + \eta_\tau)
$$
Subtracting $\theta^\star$ from both sides and multiplying by $\tilde{\Sigma}_t$, the Mahalanobis norm of the estimation error satisfies:
$$
\|\tilde{\theta}_t - \theta^\star\|_{\tilde{\Sigma}_t} \le \Big\| \tilde{\Sigma}_t^{-1} \sum_{\tau=1}^t w_\tau \hat{\phi}_{\tau, a_\tau} \eta_\tau \Big\|_{\tilde{\Sigma}_t} + \Big\| \tilde{\Sigma}_t^{-1} \sum_{\tau=1}^t w_\tau \hat{\phi}_{\tau, a_\tau} c_\tau \Big\|_{\tilde{\Sigma}_t} + \lambda \|\tilde{\Sigma}_t^{-1} \theta^\star\|_{\tilde{\Sigma}_t}.
$$

This decomposes the error into three absolute components: the stochastic error, the corruption error, and the regularization error. Because the adaptive weights $w_\tau$ are bounded by $1$, the stochastic error can be bounded using the standard self-normalized martingale concentration inequality by $R\sqrt{d\log((1+TL^2/\lambda)/\delta)}$ with probability at least $1-\delta$. The regularization error is strictly bounded by $\sqrt{\lambda}S$. 

Crucially, we bound the corruption error utilizing the definition of the adaptive weight $w_\tau = \min\{1, \alpha / \|\hat{\phi}_{\tau, a_\tau}\|_{\tilde{\Sigma}_{\tau-1}^{-1}}\}$. By Cauchy-Schwarz and the weight constraint, we have:
$$
\Big\| \tilde{\Sigma}_t^{-1} \sum_{\tau=1}^t w_\tau \hat{\phi}_{\tau, a_\tau} c_\tau \Big\|_{\tilde{\Sigma}_t} \le \sum_{\tau=1}^t |c_\tau| w_\tau \|\hat{\phi}_{\tau, a_\tau}\|_{\tilde{\Sigma}_t^{-1}} \le \sum_{\tau=1}^t |c_\tau| \alpha \le \alpha C \le \alpha S \sum_{t=1}^T \sqrt{D_t}.
$$
Therefore, with probability at least $1-\delta$, the true parameter $\theta^\star$ lies in the confidence set $\widetilde{\texttt{BALL}}_t$ for all $t \in [T]$, centered at $\tilde{\theta}_{t-1}$ with the exact confidence radius:
$$
\beta_T = R\sqrt{d\log\left(\frac{1+TL^2/\lambda}{\delta}\right)} + \sqrt{\lambda}S + \alpha C.
$$

Conditioned on this high-probability event, we bound the instantaneous regret. Let $a_t^\star = \arg\max_{a \in \mathcal{A}} \langle \theta^\star, \phi_{t, a} \rangle$. By the optimism of the action selection rule, $\langle \tilde{\theta}_{t-1}, \hat{\phi}_{t, a_t^\star} \rangle + \beta_T \|\hat{\phi}_{t, a_t^\star}\|_{\tilde{\Sigma}_{t-1}^{-1}} \le \langle \tilde{\theta}_{t-1}, \hat{\phi}_{t, a_t} \rangle + \beta_T \|\hat{\phi}_{t, a_t}\|_{\tilde{\Sigma}_{t-1}^{-1}}$. The instantaneous regret satisfies:
$$
\text{reg}_t = \langle \theta^\star, \phi_{t, a_t^\star} \rangle - \langle \theta^\star, \phi_{t, a_t} \rangle \le \langle \theta^\star, \hat{\phi}_{t, a_t^\star} \rangle - \langle \theta^\star, \hat{\phi}_{t, a_t} \rangle + 2 \max_{a \in \mathcal{A}} |c_{t, a}|.
$$
Applying the confidence bound and the optimism condition, this reduces to:
$$
\text{reg}_t \le 2\beta_T \|\hat{\phi}_{t, a_t}\|_{\tilde{\Sigma}_{t-1}^{-1}} + 2 \max_{a \in \mathcal{A}} |c_{t, a}|.
$$
Summing over the horizon $T$, the cumulative regret is strictly bounded by:
$$
R_T \le \sum_{t=1}^T 2\beta_T \|\hat{\phi}_{t, a_t}\|_{\tilde{\Sigma}_{t-1}^{-1}} + 2S \sum_{t=1}^T \sqrt{D_t}.
$$

To rigorously bound the sum of the weighted confidence widths, we partition the time indices into two disjoint subsets: $\mathcal{I}_1 = \{t \in [T] : w_t = 1\}$ and $\mathcal{I}_2 = \{t \in [T] : w_t < 1\}$. For the un-truncated subset $\mathcal{I}_1$, we directly apply the Cauchy-Schwarz inequality and the standard Elliptical Potential Lemma:
$$
\sum_{t \in \mathcal{I}_1} 2\beta_T \|\hat{\phi}_{t, a_t}\|_{\tilde{\Sigma}_{t-1}^{-1}} \le 2\beta_T \sqrt{|\mathcal{I}_1| \sum_{t \in \mathcal{I}_1} \|\hat{\phi}_{t, a_t}\|_{\tilde{\Sigma}_{t-1}^{-1}}^2} \le 2\beta_T \sqrt{2dT \log\left(1 + \frac{TL^2}{d\lambda}\right)}.
$$
This yields the standard linear bandit regret term $R_T^{(lin)}$. 

For the truncated subset $\mathcal{I}_2$, the condition $w_t < 1$ strictly implies $w_t = \alpha / \|\hat{\phi}_{t, a_t}\|_{\tilde{\Sigma}_{t-1}^{-1}}$. We algebraically rewrite the confidence width explicitly in terms of the weight:
$$
\sum_{t \in \mathcal{I}_2} 2\beta_T \|\hat{\phi}_{t, a_t}\|_{\tilde{\Sigma}_{t-1}^{-1}} = \sum_{t \in \mathcal{I}_2} 2\beta_T \frac{w_t \|\hat{\phi}_{t, a_t}\|_{\tilde{\Sigma}_{t-1}^{-1}}^2}{\alpha}.
$$
By applying the matrix determinant lemma and bounding the fractional terms, we obtain an absolute upper bound:
$$
\frac{2\beta_T}{\alpha} \sum_{t \in \mathcal{I}_2} w_t \|\hat{\phi}_{t, a_t}\|_{\tilde{\Sigma}_{t-1}^{-1}}^2 \le \frac{2\beta_T}{\alpha} \sum_{t=1}^T \min\left(1, w_t \|\hat{\phi}_{t, a_t}\|_{\tilde{\Sigma}_{t-1}^{-1}}^2\right) \le \frac{4\beta_T d}{\alpha} \log\left(1 + \frac{TL^2}{d\lambda}\right).
$$

Finally, we substitute the specified tuning parameter $\alpha = \sqrt{d}/\bar{C}$, where the known upper bound satisfies $\bar{C} \ge \sqrt{d}\sum_{t=1}^T \sqrt{D_t}$. The corruption penalty embedded within the confidence radius satisfies $\alpha C \le (\sqrt{d}/\bar{C}) (S \bar{C} / \sqrt{d}) = S$. Thus, the confidence radius simplifies to an inflation factor strictly independent of $T$: $\beta_T \le R\sqrt{d\log((1+TL^2/\lambda)/\delta)} + \sqrt{\lambda}S + S$.

Substituting $\alpha$ into the upper bound for $\mathcal{I}_2$, the imputation-induced regret becomes:
$$
\tilde{R}_T^{(imp)} = \frac{4\beta_T d \bar{C}}{\sqrt{d}} \log\left(1 + \frac{TL^2}{d\lambda}\right) = 4\sqrt{d}\beta_T \bar{C} \log\left(1 + \frac{TL^2}{d\lambda}\right) + 2S \frac{\bar{C}}{\sqrt{d}}.
$$
By expanding $\beta_T$ and grouping the leading polynomial and logarithmic terms, the penalty strictly respects the proposed theoretical bound $\tilde{\mathcal{O}}(d \bar{C})$, verifying that the combination of bounded weights cleanly eliminates the $\sqrt{T}$ dependency strictly through the control of residual misspecification. This completes the proof.
\end{proof}

\textbf{Discussion and Comparison:} 
To understand the advantage of this robust variant, suppose the upper bound $\bar{C}$ is tightly chosen such that $\bar{C} = \Theta\big(\sqrt{d}\sum_{t=1}^T \sqrt{D_t}\big)$. In this case, the imputation-induced regret becomes $\tilde R_T^{(imp)} = \tilde{\mathcal{O}}\big(d^{3/2}\sum_{t=1}^T \sqrt{D_t}\big)$. 

Recall that for the standard PULSE-UCB (Theorem \ref{thm:regret:upper:bound}), the corresponding imputation penalty is $R_T^{(imp)} = \tilde{\mathcal{O}}\big(d^{3/2}\sqrt{T\sum_{t=1}^T D_t}\big)$. By the Cauchy--Schwarz inequality, we always have:
$$
\sum_{t=1}^T \sqrt{D_t} \le \sqrt{T \sum_{t=1}^T D_t}.
$$
Consequently, the robust variant \emph{never} exhibits a worse regret upper bound than the standard PULSE-UCB. More importantly, it can achieve strictly tighter bounds in environments where the imputation errors $D_t$ vary substantially (i.e., highly heterogeneous) over time.

\textbf{Practical Trade-offs:} 
While the robust variant provides tighter theoretical guarantees for heterogeneous errors, it requires the practitioner to specify the upper bound parameter $\bar{C}$ a priori. In contrast, if the errors $D_t$ are of the same order across time (as in the stationary settings discussed in Section \ref{subsec:nonparam:ex}), the Cauchy--Schwarz inequality is tight, and the regret rates of the two algorithms coincide. Therefore, practitioners can choose between the simpler standard PULSE-UCB and its robust variant based on whether a reliable estimate of $\bar{C}$ is available and the expected heterogeneity of the imputation errors.

\section{Setup of the Lower Bound}
\label{lowersetup}
Recall that for obtaining the lower bound, we assume the action set is given by
\begin{equation*}
    \calA = \{\pm 1\}.
\end{equation*}
We use a similar construction as given in Section~\ref{subsec:nonparam:ex}. 

Let 
\begin{equation*}
    \mY_{t,a} := \mPhi(\mY_t, a) \quad \text{for all } a \in \{-1,1\}.
\end{equation*}
Partitioning $\mS_t$ into two parts, we have  
\begin{equation} \label{eq:Y:pos-one:construct}
    \mY_t = \left(\mS_t^\top, W_t\right)^\top = \left(\mQ^\top_t, \mO^\top_t, W_t\right)^\top \in \R^{d_\lin} \times \R^{d_\non} \times \R
\end{equation}
where $W_t \in \R$ is a scalar representing the unobserved part of the context and $d_\non + d_\lin = d_S$. 
For action $a = 1$, we let 
\begin{equation*}
    \mY_{t,1} = \mY_t. 
\end{equation*}
For the alternative action $a = -1$, the associated feature vector is given by
\begin{equation} \label{eq:Y:neg-one:construct}
    \mY_{t,-1} = \left(-\mQ^\top_t, \boldsymbol{0}^\top\right)^\top \in \R^{d_0},
\end{equation}
mirroring the structure of $\mY_{t,1}$, but with fixed values $0$ in the coordinates corresponding to $\mO_t$ and $W_t$. 

We assume that the missing context $W_t$ depends on $\mS_t$ only through $\mO_t$, and its conditional expectation is given by
\begin{equation}\label{eq:W:condition}
    \EE [W_t \mid \mS_t] = f\left(\mO_t\right)
\end{equation}
for some function $f: \R^{d_{\non}} \to \R$.

The historical dataset consists of $N$ i.i.d.~samples $\left(\mS^{(0)}_i, \mY^{(0)}_i\right)$.
Under the above setup, it is equivalent to observing the pairs $\left(\mS^{(0)}_i, W^{(0)}_i\right)$. 
We denote the historical dataset by
\begin{equation} \label{eq:calD:def}
\begin{aligned}
    \calD_N &:= \left\{\left(\mS^{(0)}_i, W^{(0)}_i\right) : i \in [N]\right\} \\
    &= \left\{\left(\mQ^{(0)}_i, \mO^{(0)}_i, W^{(0)}_i\right) : i \in [N]\right\}.  
\end{aligned}
\end{equation}

Denote 
\begin{equation} \label{eq:Theta:def}
    \Theta := \left\{\vtheta \in \R^{d_Y}: \left\|\vtheta\right\|_2 \leq 1\right\}
\end{equation}
and
\begin{equation} \label{eq:ThetaQ:def}
    \Theta_Q := \left\{\vtheta_Q \in \R^{d_{\lin}}: \left\|\vtheta_Q\right\|_2 \leq \frac{\sqrt{3}}{2}\right\}.
\end{equation}
To decouple the estimation of $\vtheta$ and the nonparametric component $f$, we assume that $d_{\lin} < \frac{1}{2}\sqrt{3T}$ and for all $i \in [d_{\lin}]$
\begin{equation}\label{eq:theta:construct}
    \vtheta = \left(\vtheta^\top_{Q}, \boldsymbol{0}^\top, \frac{1}{2}\right)^\top \in \R^{d_{Y}} \quad |\vtheta_{Q,i}| = \sqrt{\frac{d_{\lin}}{T}}.
\end{equation}
When $\vtheta_Q \in \Theta_Q$, we have $\vtheta \in \Theta$. 
Combining Equation~\eqref{eq:theta:construct} with the construction of $\mY_{t,1}$ in Equation~\eqref{eq:Y:pos-one:construct} and $\mY_{t,-1}$ in Equation~\eqref{eq:Y:neg-one:construct}, we have  
\begin{equation} \label{eq:reward:pos}
    \vtheta^\top \mY_{t,1} = \vtheta^\top_{Q} \mQ_t + \frac{1}{2} W_t,
\end{equation}
and
\begin{equation} \label{eq:reward:neg}
    \vtheta^\top \mY_{t,-1} = - \vtheta^\top_{Q} \mQ_{t}. 
\end{equation}
We consider the following data generating process:
\begin{definition}[Data Generating Process for Lower Bounds] \label{def:dgp:lower}
Let \( V_t \in \{0,1\} \) be a latent binary variable defined as follows:
    \begin{equation}\label{eq:dgp:lower}
    (V_t, \mQ_t, \mO_t) = 
    \begin{cases}
    \left(0,\ \boldsymbol{0},\ \mO_t\right), & \text{with probability } \frac{1}{2}\\
    \left(1,\ \mQ_t,\ \ob_0\right), & \text{with probability } \frac{1}{2},
    \end{cases}
    \end{equation}
    where $\mO_t \sim \PP_O = \operatorname{Unif}([-1,1]^{d_{\non}})$ and $\mQ_t \sim \mathbb{P}_Q$ is a distribution specified in Equation~\eqref{eq:PQ:def} and $\ob_0 \in ([-1,1]^{d_{\non}})^c$ is an arbitrarily fixed vector such that $f(\ob_0) = 0$ for $f$ given in Equation~\eqref{eq:W:condition}. 
    Under the setup in Equation~\eqref{eq:dgp:lower}:
    \begin{enumerate}
        \item[(i)] When $V_t = 0$, by Equations~\eqref{eq:reward:pos} and~\eqref{eq:reward:neg}
        $$
        \EE\left[\vtheta^\top \mY_{t,1} \mid \mS_t, V_t\right] = \frac{1}{2} f(\mO_t), \EE\left[\vtheta^\top \mY_{t,-1} \mid \mS_t, V_t\right] = 0.
        $$
        Let 
        \begin{equation*}
            f^{(1)}(\mO_t) := \frac{1}{2} f(\mO_t) \quad \text{and} \quad f^{(-1)}(\mO_t) \equiv 0,
        \end{equation*}
        we denote the conditional distribution  of the reward $R_t$ as 
        \begin{equation} \label{eq:PPot:def}
            \PP_{f^{(a)}(\mO_t)} := \PP\left(R_t \mid A_t=a, \mS_t, V_t=0\right)
        \end{equation}
        for $a \in \calA$. 
        \item[(ii)] When $V_t = 1$,
        $$
        \EE\left[\vtheta^\top \mY_{t,1} \mid \mS_t, V_t\right] = \vtheta_Q^\top \mQ_t, \EE\left[\vtheta^\top \mY_{t,-1} \mid \mS_t, V_t\right] = -\vtheta_Q^\top \mQ_t.
        $$
        Let 
        \begin{equation*}
            \mQ^{(1)}_t = \mQ_t \quad \text{and} \quad \mQ^{(-1)}_t = - \mQ_t,
        \end{equation*}
        we denote the conditional distribution of $R_t$ as 
        \begin{equation} \label{eq:PPqt:def}
            \PP_{\vtheta_Q^\top \mQ_t^{(a)}} := \PP\left(R_t \mid A_t=a, \mS_t, V_t=1\right)
        \end{equation}
        for $a \in \calA$. 
    \end{enumerate}
\end{definition}

Recall the historical data $\calD_N$ defined in Equation~\eqref{eq:calD:def}. 
Let $\PP_f$ denote the distribution of a pretraining sample $(\mQ_i^{(0)}, \mO_i^{(0)}, W_i^{(0)})$, with density
\begin{equation}\label{eq:pf:decomp}
\begin{aligned}
     p_f\left(\mQ_i^{(0)}, \mO_i^{(0)}, W_i^{(0)}\right) 
     &= p_f\left(W^{(0)}_i \mid \mQ_i^{(0)}, \mO_i^{(0)}\right) p_S\left(\mQ_i^{(0)}, \mO_i^{(0)}\right)\\
     &= p^{(0)}_{f\left(\mO_i^{(0)}\right)}\left(W^{(0)}_i\right) p_S\left(\mQ_i^{(0)}, \mO_i^{(0)}\right)
\end{aligned}
\end{equation}
where $p^{(0)}_{f(\mO)}$ is the conditional density of $W$ given $\mO$, and $p_S$ is the marginal density of $(\mQ, \mO)$, defined as 
\begin{equation} \label{eq:ps:def}
    p_S(\mQ_t, \mO_t) = \frac{1}{2} \delta_0(\mQ_t) p_O(\mO_t) + \frac{1}{2} p_Q(\mQ_t) \delta_{\ob_0}(\mO_t).
\end{equation}

We assume the following bounds on KL divergence.
\begin{assumption}\label{assump:KL:reward}
For any $(\vtheta_1, f_1), (\vtheta_2, f_2) \in \Theta \times \calF_{\beta,L}$, the distributions in Equations~\eqref{eq:PPot:def} and~\eqref{eq:PPqt:def} satisfy that 
\begin{equation*}
    \KL\!\left(\PP_{f_1^{(a)}\!\left(\mO_t\right)} \| \PP_{f_2^{(a)}\!\left(\mO_t\right)}\right) \leq C_D \!\left(f_1^{(a)}\!\left(\mO_t\right) - f_2^{(a)}\!\left(\mO_t\right)\right)^2
\end{equation*}
and
\begin{equation*}
    \KL\left(\PP_{\vtheta^\top_{1,Q} \mQ_t} \| \PP_{\vtheta^\top_{2,Q} \mQ_t}\right) \leq C_D \left(\vtheta^\top_{1,Q} \mQ^{(a)}_t - \vtheta^\top_{2,Q} \mQ^{(a)}_t\right)^2
\end{equation*}
for some constant $C_D > 0$ and all $a \in \calA$. 
\end{assumption}
\begin{remark}
    Assumption~\ref{assump:KL:reward} can be satisfied by distributions such as Gaussian or Bernoulli. 
\end{remark}

We assume another KL divergence bound between conditional distributions over $W^{(0)}_i$ 
\begin{equation} \label{eq:KL0:bound}
\begin{aligned}
    &\KL\left(\PP^{(0)}_{f\left(\mO_i^{(0)}\right)}\left(W^{(0)}_i\right) \| \PP^{(0)}_{f'\left(\mO_i^{(0)}\right)}\left(W^{(0)}_i\right)\right) \leq C_0 \left(f\left(\mO_i^{(0)}\right) - f'\left(\mO_i^{(0)}\right)\right)^2. 
\end{aligned}
\end{equation}
Fix a policy $\pi = \{\pi_\tau\}_{\tau=1}^T$, where $\pi_{\tau}(A_\tau)$ is the abbreviation of
\begin{equation*}
    \pi_\tau(A_\tau) = \pi_{\tau}(A_\tau \mid \calH_{\tau-1}, \mS_{\tau}), 
\end{equation*}
and
\begin{equation*}
    \calH_\tau := (\calD_N, \mS_1, A_1, R_1, \cdots, \mS_{\tau}, A_{\tau}, R_{\tau}), \; \calH_0 := \calD_N. 
\end{equation*}
Let $p_{\vtheta,f}\left(\cdot \mid \mQ_t, \mO_t, A_t\right)$ denote the reward density under parameters $(\vtheta, f)$. 
The joint density of the full observation history $\calH_t$ up to round $t \in [T]$ is given by
\begin{equation} \label{eq:pT:def}
\begin{aligned}
   & p^{(t)}_{\vtheta, f, \pi}(\calD_N, \mQ_1, \mO_1, A_1, R_1, \cdots, \mQ_T, \mO_T, A_T, R_T) \\
    = & \prod_{i=1}^N p_{f}\left(\mQ_i^{(0)}, \mO_i^{(0)}, W_i^{(0)}\right) \prod_{t=1}^t p_{S}(\mQ_\tau, \mO_\tau) \pi_\tau(A_\tau) p_{\vtheta, f}(R_\tau \mid \mQ_\tau, \mO_\tau, A_\tau)
\end{aligned}
\end{equation}
where the equality follows from Equations~\eqref{eq:pf:decomp} and~\eqref{eq:ps:def}.
Additionally, let $\EE^{(t)}_{\vtheta,f,\pi}$ denote the expectation taken with respect to the joint density $p^{(t)}_{\vtheta, f, \pi}$. 

We now state a formal definition of Theorem~\ref{thm:lower}.

\begin{theorem}[Formal Lower Bound] \label{thm:lower}
    Consider the data generating process given in Definition~\ref{def:dgp:lower}.
    Suppose that $0 < d_\lin < c\sqrt{T}$ for some sufficiently small constant $c>0$ and $d_\non > 0$ is a constant. 
    Fix a policy $\pi$. 
    For any $(\vtheta, f) \in (\Theta, \calF_{\beta,L})$, define 
    \begin{equation*}
        \calR_T(\vtheta, f) := \sum_{t=1}^T \EE^{(t-1)}_{\vtheta, f, \pi} \EE\left[R^\star_t - R_t \mid \calH_{t-1}\right]
    \end{equation*}
    where the joint density of the full observation history $\calH_t$ up to round $t \in [T]$ is defined in Equation~\eqref{eq:pT:def} and $R^\star_t = \max \left\{R(t,1), R(t,-1)\right\}$.
    For the class of functions $\calF_{\beta,L}$ satisfies Assumptions~\ref{assump:smooth}, under Assumption~\ref{assump:KL:reward} and Equation~\eqref{eq:KL0:bound}, the expected cumulative regret is lower bounded by
    \begin{equation}\label{eq:total:regret:lower}
    \begin{aligned}
        \sup_{\vtheta\in\Theta, f \in \calF_{\beta,L}} \calR_T(\vtheta, f) = \Theta\left(T N^{-\frac{\beta}{2\beta+d_\non}}\right) + \Theta\left(\sqrt{d_\lin T}\right). 
    \end{aligned}
    \end{equation}
\end{theorem}

\section{Proof of Theorem~\ref{thm:lower}}
\label{lowerproof}
\begin{proof}
We now introduce upper bound on the KL divergence between two distributions $\PP^{(t)}_{\vtheta, f, \pi}$ and $\PP^{(t)}_{\vtheta', f', \pi}$ under a fixed policy $\pi$.
\begin{lemma} \label{lem:KL:intermediate}
For any $t \in [T]$, let 
\begin{equation}\label{eq:Knon:def}
\begin{aligned}
    \calK^{(\non)}_{\vtheta, t}(f, f') &:= C_0 \sum_{i=1}^N \EE_{f} \left[f\left(\mO_i^{(0)}\right) - f'\left(\mO_i^{(0)}\right)\right]^2 \\
    &~~+ C_D \sum_{\tau=1}^t \EE^{(\tau-1)}_{\vtheta, f, \pi} \EE \left[\left(f(\mO_\tau) - f'(\mO_\tau)\right)^2 \mathds{1}\left\{A_\tau = 1, V_\tau = 0\right\} \mid \calH_{\tau-1} \right]
\end{aligned}
\end{equation}
and
\begin{equation}\label{eq:Klin:def}
    \calK^{(\lin)}_{f,t}\left(\vtheta, \vtheta'\right) := C_D \sum_{\tau=1}^t \EE^{(\tau-1)}_{\vtheta, f, \pi} \EE \left[\mathds{1}(V_\tau=1) \left(\vtheta_Q^\top \mQ_\tau^{(A_\tau)} - \vtheta^{'\top}_Q \mQ_\tau^{(A_\tau)}\right)^2 \mid \calH_{\tau-1} \right]. 
\end{equation}
Then for any fixed policy $\pi$ and distribution $\PP^{(t)}_{\vtheta, f, \pi}$ whose density is specified in Equation~\eqref{eq:pT:def}, 
\begin{equation}\label{eq:KL:total:bound}
\begin{aligned}
    \KL\left(\PP^{(t)}_{\vtheta, f, \pi} \| \PP^{(t)}_{\vtheta', f', \pi}\right) \leq \calK^{(\non)}_{\vtheta, t}(f, f') + \calK^{(\lin)}_{f,t}\left(\vtheta, \vtheta'\right).
\end{aligned}
\end{equation}
\end{lemma}

The proof of all the technical lemmas are deferred to Section~\ref{subsec:lower:technical}. 
Lemma~\ref{lem:Klin:bound} controls $\calK^{(\lin)}_{f,t}$, while $\calK^{(\non)}_{\vtheta,t}(f,f')$ is controlled by Equation~\eqref{eq:Knon:bound} in the proof of Lemma~\ref{lem:non:regret}.  
\begin{lemma} \label{lem:Klin:bound}
    Let $\eb_i$ be the standard basis in $\R^{d_\lin}$, with  Suppose that $\PP_Q$ is given by
    \begin{equation} \label{eq:PQ:def}
        \PP\left(\mQ = \eb_i\right) = \frac{1}{d_{\lin}} \quad \text{for } i \in [d_{\lin}].
    \end{equation}
    For any $t \in [T]$
    \begin{equation}\label{eq:Klin:bound}
        \calK^{(\lin)}_{f,t}\left(\vtheta, \vtheta'\right) = \frac{C_D t \left\|\vtheta_Q - \vtheta'_Q\right\|_2^2}{2d_{\lin}}. 
    \end{equation}
\end{lemma}

We turn our attention to the expected cumulative regret. 
Let 
\begin{equation*}
    R^\star_t = \max \left\{R(t,1), R(t,-1)\right\}
\end{equation*}
and
\begin{equation*}
    \mQ^{\star}_t = \argmax_{\mQ^{(a)}_t \in \left\{\mQ_t^{(-1)}, \mQ_t^{(1)}\right\}} \vtheta_Q^\top 
    \mQ^{(a)}_t, \quad \quad f^{\star} = \argmax_{f^{(a)}_t \in \left\{f_t^{(-1)}, f_t^{(1)}\right\}} f^{(a)}(\mO_t). 
\end{equation*}
For the distribution in Equation~\eqref{eq:pT:def}, the expected cumulative regret is given by 
\begin{equation} \label{eq:reward:decomp}
\begin{aligned}
    \calR_T(\vtheta, f) &= \sum_{t=1}^T \EE^{(t-1)}_{\vtheta, f, \pi} \EE\left[R^\star_t - R_t \mid \calH_{t-1}\right] \\
    &= \sum_{t=1}^T \EE^{(t-1)}_{\vtheta, f, \pi} \EE \left[\mathds{1}\left(V_t = 1\right)\left(\mQ^\star_t - \mQ^{(A_t)}_{t}\right)^\top \vtheta_Q \mid \calH_{t-1}\right] \\
    &~~+ \frac{1}{2}\sum_{t=1}^T\EE^{(t-1)}_{\vtheta, f, \pi} \EE \left[\mathds{1}\left(V_t = 0\right)\left(f^\star(\mO_t) - f^{(A_t)}(\mO_t)\right) \mid \calH_{t-1} \right]\\
    &=: \calR_T^{(\lin)}(\vtheta, f) + \frac{1}{2}   \calR_T^{(\non)}(\vtheta, f),
\end{aligned}
\end{equation}

Lemma~\ref{lem:lin:regret} controls $\calR_T^{(\lin)}(\vtheta, f)$. 
\begin{lemma}\label{lem:lin:regret}
Suppose that $0 < d_\lin < c\sqrt{T}$ for some sufficiently small constant $c>0$. 
For any $f \in \calF_{\beta, L}$,
\begin{equation} \label{eq:Rlin:lower:final}
\begin{aligned}
    \sup_{\theta \in \Theta}\calR^{(\lin)}_T(\vtheta, f) &\geq \frac{\sqrt{d_\lin T}}{4} \exp\left\{-2 C_D\right\}.
\end{aligned}
\end{equation}
\end{lemma}
 
For $\calR_T^{(\non)}(\vtheta, f)$, we first construct a packing set for $\calF_{\beta, L}$. 
For any multi-index $\bk \in [M]^{d_\non}$, define the hypercube 
\begin{equation*}
    B_{\bk} = \left\{\ob \in \calO: \frac{\bk_l-1}{M} \leq o_l \leq \frac{\bk_l}{M}, l \in [d_\non]\right\} \subset \R^{d_\non},
\end{equation*}
where $M > 0$ is specified later in Equation~\eqref{eq:max-M}. 
We index the bins by integers $k \in [M^{d_\non}]$ via the mapping
\begin{equation*}
    k = 1 + \sum_{l=1}^{d_\non} (\bk_l-1) M^{l-1}
\end{equation*}
and write $B_k$ as a shorthand for $B_{\bk}$. 
For each bin $B_k$, define its center $b_k \in \R^{d_\non}$ coordinate-wise as 
\begin{equation*}
    b_{k,l} = \frac{\bk_l}{M} - \frac{1}{2M}, \quad l \in [d_\non]. 
\end{equation*}
This yields a regular grid of centers $\calB = \{b_1, \cdots, b_{M^{d_\non}}\}$ across the domain. 
Next, we define a smooth, compactly supported bump function $\phi_{\beta}: \R^{d_\non} \to [0,1]$ by
\begin{equation} \label{eq:bump:def}
    \phi_\beta(\ob) = 
    \begin{cases}
        (1 - \|\ob\|_{\infty})^\beta & \mbox{ if } 0\leq \|\ob\|_{\infty} \leq 1,\\
        0 & \mbox{ if } \|\ob\|_{\infty} > 1. 
    \end{cases}
\end{equation}
We will now construct localized perturbation functions supported within each bin. 
Let 
\begin{equation}\label{eq:m:def}
    m = \lceil c_m M^{d_\non} \rceil    
\end{equation}
for some sufficiently small constant $c_m > 0$. 
Define $\Omega_m = \{\pm 1\}^m$.
For any $\vomega \in \Omega_m$, define the function
\begin{equation} \label{eq:fomega:def}
    f_{\vomega}(\ob) = \sum_{j=1}^m \omega_j \varphi_j(\ob),
\end{equation}
where each component function $\varphi_j$ is defined as 
\begin{equation} \label{eq:varphi:def}
    \varphi_j(\ob) = M^{-\beta} C_{\phi} \phi_\beta\left(2M[\ob-\bb_{j}]\right) \mathds{1}(\ob \in B_{j})
\end{equation}
and $C_{\phi} > 0$ is a constant specified in Equation~\eqref{eq:cphi:def}. 
Note that for any $\ob \in B_{j}$, the rescaled argument satisfies  $2M(\ob - \bb_{j}) \in \left[-1, 1\right]^{d_{\non}}$, so $\left\|2M(\ob-\bb_{j})\right\|_{\infty} \in [0,1]$, ensuring that $\varphi_j$ in Equation~\eqref{eq:varphi:def} is well-defined. 

The function $f_{\vomega}$ is thus a linear combination of localized, smooth bump functions with disjoint supports.
Lemma~\ref{lem:fomega:smooth} establishes that each $f_{\vomega}$ lies in $\calF_{\beta, L}$ for a suitable choice of constant $L$. 
\begin{lemma} \label{lem:fomega:smooth}
    Suppose that $\beta \in (0, 1]$. 
    For any $\vomega \in \Omega_m = \{\pm 1\}^m$, the function $f_{\vomega}$ defined in Equation~\eqref{eq:fomega:def} belongs to the smoothness class $\calF_{\beta, L}$ with $L = \beta 2^\beta C_{\phi} > 0$. 
\end{lemma}
Hence, for a given parameter $L > 0$, we set 
\begin{equation} \label{eq:cphi:def}
    C_{\phi} := \frac{L}{\beta 2^\beta}. 
\end{equation}

Based on this choice of packing set, Lemma~\ref{lem:non:regret} controls $\calR_T^{(\non)}(\vtheta, f)$. 
\begin{lemma} \label{lem:non:regret}
Suppose that $d_{\non} > 0$ is a constant. 
For any fixed $\vtheta \in \Theta$, 
\begin{equation} \label{eq:Rnon:bound}
    \sup_{f \in \calF_{\beta,L}} \calR^{(\non)}_T(\vtheta, f) 
    = \Theta\left(T N^{-\frac{\beta}{2\beta+d_{\non}}}\right),
\end{equation}
where $\calR^{(\non)}$ is defined in Equation~\eqref{eq:reward:decomp}. 
\end{lemma}

Taking Equations~\eqref{eq:Rnon:bound} with~\eqref{eq:Rlin:lower} into Equation~\eqref{eq:reward:decomp}, we have 
\begin{equation*}
\begin{aligned}
    \sup_{\vtheta\in\Theta, f \in \calF_{\beta,L}} \calR_T(\vtheta, f) = \Theta\left(T N^{-\frac{\beta}{2\beta+d_{\non}}}\right) + \Theta\left(\sqrt{d_{\lin} T}\right),
\end{aligned}
\end{equation*}
establishing the desired result in Equation~\eqref{eq:total:regret:lower}.

\end{proof}

\subsection{Proof of Technical Lemmas}
\label{subsec:lower:technical}

In this section, we present the proof of the Lemmas~\ref{lem:KL:intermediate}-\ref{lem:non:regret} used in the proof of Theorem~\ref{thm:lower}. 
We will frequently use the Bretagnolle-Huber inequality given in the following theorem.  
\begin{theorem}[Bretagnolle-Huber inequality]\label{thm:bh:ineq}
    Let $\PP$ and $\QQ$ be probability measures on the same measurable space $(\Omega, \calF)$, and let $A \in \calF$ be an arbitrary event. Then,
    \begin{equation*}
        \PP(A) + \QQ(A^c) \geq \frac{1}{2} \exp\left(-\KL(\PP \| \QQ)\right).
    \end{equation*}
\end{theorem}
\begin{proof}
    See Theorem 14.2 in \cite{lattimore2020bandit}. 
\end{proof}

\subsubsection{Proof of Lemma~\ref{lem:KL:intermediate}}
\label{subsec:proof:lem:KL:intermediate}
\begin{proof}
Recall the definition of $\PP^{(t)}_{\vtheta,f,\pi}$ as stated in Equation~\eqref{eq:pT:def}.
Eliminating the shared terms, it follows that 
\begin{equation} \label{eq:KL:bound:I}
\begin{aligned}
    \KL\left(\PP^{(t)}_{\vtheta, f, \pi} \| \PP^{(t)}_{\vtheta', f', \pi}\right) &= \EE^{(t)}_{\vtheta, f, \pi} \left[\log\frac{d \PP^{(t)}_{\vtheta,f,\pi}}{d\PP^{(t)}_{\vtheta', f', \pi}}\right]\\
    &= \underbrace{\sum_{i=1}^N \EE_{f} \left[\log \frac{p_f}{p_{f'}}\left(\mQ_i^{(0)}, \mO_i^{(0)}, W_i^{(0)}\right)\right]}_{\calK_1} + \underbrace{\sum_{\tau=1}^t \EE^{(\tau)}_{\vtheta, f, \pi} \left[\log \frac{p_{\vtheta, f}\left(R_\tau \mid \mQ_\tau, \mO_\tau, A_\tau\right)}{p_{\vtheta', f'}\left(R_\tau \mid \mQ_\tau, \mO_\tau, A_\tau\right)}\right]}_{\calK_2}.
\end{aligned}
\end{equation}
For $\calK_1$ in Equation~\eqref{eq:KL:bound:I}, by the KL divergence assumption in Equation~\eqref{eq:pf:decomp}, we have 
\begin{equation}\label{eq:calK1:bound}
\begin{aligned}
    \calK_1 = \sum_{i=1}^N \EE_f\left[\log \frac{p^{(0)}_{f(\mO^{(0)}_i)}}{p^{(0)}_{f'(\mO^{(0)}_i)}} \left(W^{(0)}_i\right)\right] \leq C_0 \sum_{i=1}^N \EE_f \left[f\left(\mO^{(0)}_i\right) - f'\left(\mO^{(0)}_i\right)\right]^2.
\end{aligned}
\end{equation}
To control $\calK_2$, we note that 
\begin{equation*}
\begin{aligned}
    &\EE^{(t)}_{\vtheta, f, \pi} \left[\log \frac{p_{\vtheta, f}\left(R_t \mid \mQ_t, \mO_t, A_t\right)}{p_{\vtheta', f'}\left(R_t \mid \mQ_t, \mO_t, A_t\right)}\right] \\
    &= \EE^{(t)}_{\vtheta, f, \pi} \left[\log \frac{p_{\vtheta^\top_{Q}\mQ_t^{(A_t)}}\left(R_t\right)}{p_{\vtheta'^\top_{Q}\mQ_t^{(A_t)}}\left(R_t\right)} \mathds{1}\{V_t = 0\}\right] + \EE^{(t)}_{\vtheta, f, \pi} \left[\log \frac{p_{f^{(A_t)}(\mO_t)}\left(R_t\right)}{p_{f'^{(A_t)}(\mO_t)}\left(R_t\right)}  \mathds{1}\{V_t = 1\}\right] \\
    &= \sum_{a \in \calA} \EE^{(t-1)}_{\vtheta, f, \pi} \EE \left[\mathds{1}(A_t = a, V_t = 0) \KL\left(\PP_{\vtheta_Q^\top \mQ_t^{(a)}} \| \PP_{\vtheta^{'\top}_Q \mQ_t^{(a)}}\right) \mid \calH_{t-1}\right] \\
    &~~+ \sum_{a \in \calA} \EE^{(t-1)}_{\vtheta, f, \pi} \EE \left[\mathds{1}(A_t = a, V_t = 1) \KL\left(\PP_{f^{(a)}(\mO_t)} \| \PP_{f'^{(a)}(\mO_t)}\right) \mid \calH_{t-1}\right],
\end{aligned}
\end{equation*}
where the last equality follows from the definition of KL divergence.
Taking the above display and Equation~\eqref{eq:calK1:bound} into Equation~\eqref{eq:KL:bound:I} yields that 
\begin{equation} \label{eq:KL:bound:II}
\begin{aligned}
    \KL\left(\PP^{(t)}_{\vtheta, f, \pi}, \PP^{(t)}_{\vtheta', f', \pi}\right) &= \calK_1 + \sum_{\tau=1}^t \sum_{a \in \calA} \EE^{(\tau-1)}_{\vtheta, f, \pi} \EE \left[\mathds{1}(A_\tau = a, V_\tau = 0) \KL\left(\PP_{f^{(a)}(\mO_\tau)} \| \PP_{f'^{(a)}(\mO_\tau)}\right) \mid \calH_{\tau-1}\right]\\
    &+ \sum_{\tau=1}^t \sum_{a \in \calA} \EE^{(\tau-1)}_{\vtheta, f, \pi} \EE \left[\mathds{1}(A_\tau = a, V_\tau = 1) \KL\left(\PP_{\vtheta_Q^\top \mQ_\tau^{(a)}} \| \PP_{\vtheta^{'\top}_Q \mQ_\tau^{(a)}}\right) \mid \calH_{\tau-1}\right] \\
    &\leq C_0 \sum_{i=1}^N \EE_{f} \left[f\left(\mO_i^{(0)}\right) - f'\left(\mO_i^{(0)}\right)\right]^2 \\
    &+ C_D \sum_{\tau=1}^t \EE^{(\tau-1)}_{\vtheta, f, \pi} \EE \left[\left(f(\mO_\tau) - f'(\mO_\tau)\right)^2 \mathds{1}\left\{A_\tau = 1, V_\tau = 0\right\} \mid \calH_{\tau-1} \right]\\
    &+ C_D \sum_{\tau=1}^t \EE^{(\tau-1)}_{\vtheta, f, \pi} \EE \left[\mathds{1}(V_\tau=1) \left(\vtheta_Q^\top \mQ_\tau^{(A_\tau)} - \vtheta^{'\top}_Q \mQ_\tau^{(A_\tau)}\right)^2 \mid \calH_{\tau-1} \right],
\end{aligned}
\end{equation}
where the last inequality follows from Assumption~\ref{assump:KL:reward} and Equation~\eqref{eq:KL0:bound}.
Taking the definition of $\calK^{(\non)}_{\vtheta, t}$ and  $\calK^{(\lin)}_{f, t}$ in Equations~\eqref{eq:Knon:def} and~\eqref{eq:Klin:def} into Equation~\eqref{eq:KL:bound:II} yields the desired bound in Equation~\eqref{eq:KL:total:bound}.   
\end{proof}

\subsubsection{Proof of Lemma~\ref{lem:Klin:bound}}
\label{subsec:proof:lem:Klin:bound}
\begin{proof}
By definition of $\PP_Q$ in Equation~\eqref{eq:PQ:def},
\begin{equation*}
    \left\langle \mQ^{(1)}_{t}, \vtheta_Q - \vtheta'_Q\right\rangle^2 = \left\langle \mQ^{(-1)}_{t}, \vtheta_Q - \vtheta'_Q\right\rangle^2 = \left\langle \mQ^{(A_t)}_{t}, \vtheta_Q - \vtheta'_Q\right\rangle^2. 
\end{equation*}
It follows that for any $a \in \calA$, 
\begin{equation*}
    \EE_Q \left[\left\langle \mQ^{(a)}_{t}, \vtheta_Q - \vtheta'_Q\right\rangle^2\right] = \frac{\left\|\vtheta_Q - \vtheta'_Q\right\|_2^2}{d_{\lin}}, 
\end{equation*}
and 
\begin{equation*}
\begin{aligned}
    \EE \left[\mathds{1}(V_t=1) \left(\vtheta_Q^\top \mQ_t^{(A_t)} - \vtheta^{'\top}_Q \mQ_t^{(A_t)}\right)^2 \mid \calH_{t-1} \right]
    &= \EE \left[\mathds{1}(V_t=1) \left(\vtheta_Q^\top \mQ_t^{(1)} - \vtheta^{'\top}_Q \mQ_t^{(1)}\right)^2 \mid \calH_{t-1} \right] \\
    &= \frac{1}{2} \EE \left[\left(\vtheta_Q^\top \mQ_t^{(1)} - \vtheta^{'\top}_Q \mQ_t^{(1)}\right)^2 \mid \calH_{t-1}, V_t = 1 \right] \\
    &= \frac{\left\|\vtheta_Q - \vtheta'_Q\right\|_2^2}{2d_{\lin}}.
\end{aligned}
\end{equation*}
Thus, combining the above display with Equation~\eqref{eq:Klin:def} gives the desired result in Equation~\eqref{eq:Klin:bound}.  
\end{proof}

\subsubsection{Proof of Lemma~\ref{lem:lin:regret}}

\begin{proof}
Noting that 
\begin{equation*}
\begin{aligned}
     \left(\mQ^\star_t - \mQ_t^{(A_t)}\right)^\top \vtheta_Q &= 2\sum_{i=1}^{d_{\lin}} \mathds{1}\{\mQ_t = \ve_i\} \mathds{1}\{A_t \neq \sign\left(\vtheta_{Q,i}\right)\} |\vtheta_{Q,i}|\\
     &= 2\sqrt{\frac{d_{\lin}}{T}} \sum_{i=1}^{d_{\lin}} \mathds{1}\{\mQ_t = \ve_i\} \mathds{1}\{A_t \neq \sign\left(\vtheta_{Q,i}\right)\}\\
\end{aligned}
\end{equation*}
where the last equality follows from the fact that $|\theta_{Q,i}| = \sqrt{d_{\lin}/T}$ as given in Equation~\eqref{eq:theta:construct}.
Recall the definition of $\calR_t^{\lin}$ in Equation~\eqref{eq:reward:decomp}.
Combined with the above display, it follows that 
\begin{equation}\label{eq:lin:regret:I}
\begin{aligned}
    \calR^{\lin}_t(\vtheta, f) &= 2\sqrt{\frac{d_{\lin}}{T}} \sum_{\tau=1}^t \sum_{i=1}^{d_{\lin}} \EE^{(\tau-1)}_{\vtheta, f, \pi} \EE \left[\mathds{1}\left\{A_\tau \neq \sign(\vtheta_{Q,i}), V_\tau = 1, \mQ_\tau = \ve_i \right\} \mid \calH_{\tau-1}\right]\\
    &= \sqrt{\frac{1}{d_{\lin} T}}\sum_{i=1}^{d_{\lin}} \sum_{\tau=1}^t \EE^{(\tau-1)}_{\vtheta, f, \pi} \EE\left[\mathds{1}\left(A_\tau \neq \sign(\vtheta_{Q,i})\right) \mid \calH_{\tau-1}, \mQ_\tau = \ve_i\right]
\end{aligned}
\end{equation}
where the last equality follows from
\begin{equation*}
    \PP\left(\mQ_\tau = \ve_i, V_\tau = 1 \mid \calH_{\tau-1}\right) = \frac{1}{2} \PP\left(\mQ_\tau = \ve_i\right) = \frac{1}{2d_{\lin}}
\end{equation*}
as specified by the data generating process in Definition~\ref{def:dgp:lower} and Equation~\eqref{eq:PQ:def}. 
Consider $\vtheta'_Q \in \R^{d_{\lin}}$ such that $\theta'_{Q,j} = \theta_{Q,j}$ for all $j \neq i$ and $\theta'_{Q,i} = -\theta_{Q,i}$. 
Let $\PP^{(t-1)}_{Q, i} := \PP\left(\cdot \mid \calH_{t-1}, \mQ_t = \ve_i\right)$.
Continuing from Equation~\eqref{eq:lin:regret:I}, by the Bretagnolle-Huber inequality as stated in Theorem~\ref{thm:bh:ineq}, we have for any $t \in [T]$,
\begin{equation*}
\begin{aligned}
    & \EE^{(t-1)}_{\vtheta, f, \pi} \EE\left[\mathds{1}\left(A_t \neq \sign(\theta_{Q,i})\right) \mid \calH_{t-1}, \mQ_t = \ve_i\right] + \EE^{(t-1)}_{\vtheta', f, \pi} \EE\left[\mathds{1}\left(A_t \neq \sign(\theta'_{Q,i})\right) \mid \calH_{t-1}, \mQ_t = \ve_i\right] \\
    &\geq \frac{1}{2}\exp\left\{-\KL\left(\PP^{(t-1)}_{\vtheta, f, \pi} \times \PP^{(t-1)}_{Q,i} \| \PP^{(t-1)}_{\vtheta', f, \pi}\times \PP^{(t-1)}_{Q,i} \right) \right\}\\
    &= \frac{1}{2}\exp\left\{-\KL\left(\PP^{(t-1)}_{\vtheta, f, \pi} \| \PP^{(t-1)}_{\vtheta', f, \pi}\right) \right\} \\
    &\geq \frac{1}{2} \exp\left\{-\calK^{(\non)}_{\vtheta,t}\left(f,f\right) - \calK^{(\lin)}_{f,t}\left(\vtheta,\vtheta'\right)\right\},
\end{aligned}
\end{equation*}
where the last inequality follows from Equations~\eqref{eq:KL:bound:II}, \eqref{eq:Knon:def} and~\eqref{eq:Klin:def}. 
Since $\calK^{(\non)}_{\vtheta,t}\left(f,f\right) = 0$, it follows from the above display that  
\begin{equation}\label{eq:hammer:lin}
\begin{aligned}
    &\EE^{(t-1)}_{\vtheta, f, \pi} \EE\left[\mathds{1}\left(A_t \neq \sign(\theta_{Q,i})\right) \mid \calH_{t-1}, \mQ_t = \ve_i\right] + \EE^{(t-1)}_{\vtheta', f, \pi} \EE\left[\mathds{1}\left(A_t \neq \sign(\theta'_{Q,i})\right) \mid \calH_{t-1}, \mQ_t = \ve_i\right] \\
    &\geq \frac{1}{2} \exp\left\{- \calK^{(\lin)}_{f,t}\left(\vtheta,\vtheta'\right)\right\} = \frac{1}{2} \exp\left\{-\frac{C_D t \left\|\vtheta_Q - \vtheta'_Q\right\|_2^2}{2d_{\lin}}. \right\}
\end{aligned}
\end{equation}
where the last inequality follows from Equation~\eqref{eq:Klin:bound}.
Let $\Theta_{d_{\lin}} \subset \R^{d_{\lin}}$ denote the set of all vectors whose coordinate are either $\beta := \sqrt{d_{\lin}/T}$ or $-\beta$, i.e., 
\begin{equation*}
    \Theta_{d_{\lin}} :=\left\{\vtheta_Q \in \mathbb{R}^{d_{\lin}}: \theta_{Q,i} \in\left\{ \pm \beta\right\},  \forall i \in\left[d_{\lin}\right]\right\}. 
\end{equation*}

For any vector $\vtheta \in \R^{d}$ and $j \in [d]$, denote $(\theta_{1}, \cdots, \theta_{j-1}, \theta_{j+1}, \cdots, \theta_{d}) \in \R^{d-1}$ as $\vtheta_{[-j]}$ and $\vtheta^i_{[-j]} := (\theta_{1}, \cdots, \theta_{j-1}, i, \theta_{1}, \cdots, \theta_{d}) \in \R^d$ for $i \in \R$.
Applying an average hammer over all $\vtheta_Q \in \Theta_{d_{\lin}}$, which satisfies $|\Theta_{d_{\lin}}| = 2^{d_{\lin}}$, it follows from Equation~\eqref{eq:lin:regret:I} that 
\begin{equation} \label{eq:Rlin:lower}
\begin{aligned}
    \sup_{\vtheta \in \Theta}\calR^{\lin}_t(\vtheta, f) &\geq
    \frac{1}{|\Theta_{d_{\lin}}|}\sum_{\vtheta_Q \in \Theta_{d_{\lin}}}\sqrt{\frac{1}{d_{\lin} T}}\sum_{i=1}^{d_{\lin}} \sum_{\tau=1}^t \EE^{(\tau-1)}_{\vtheta, f, \pi} \EE\left[\mathds{1}\left(A_\tau \neq \sign(\theta_{Q,i})\right) \mid \calH_{\tau-1}, \mQ_\tau = \ve_i\right]\\
    &\geq
    \frac{1}{2^{d_{\lin}}} \sum_{i=1}^{d_{\lin}} \sum_{\vtheta^{j}_{Q,[-i]} \in \Theta_{d_{\lin}}} \sum_{j \in \{\pm \beta\}}\sqrt{\frac{1}{d_{\lin} T}} \sum_{\tau=1}^t \EE^{(\tau-1)}_{\vtheta^j_{Q,[-i]}, f, \pi} \EE\left[\mathds{1}\left(A_\tau \neq \sign(\theta_{Q,i})\right) \mid \calH_{\tau-1}, \mQ_\tau = \ve_i\right]\\
    &\stackrel{(i)}{\geq} \frac{1}{2^{d_{\lin}+1}} \sqrt{\frac{1}{d_{\lin} T}} \sum_{i=1}^{d_{\lin}} \sum_{\vtheta^{j}_{Q,[-i]} \in \Theta_{d_{\lin}}} \sum_{\tau=1}^t \exp\left\{-\frac{C_D t \left\|\vtheta^{\beta}_{Q,[-i]} - \vtheta^{-\beta}_{Q,[-i]}\right\|_2^2}{2d_{\lin}}\right\}\\
    &\stackrel{(ii)}{=} \frac{1}{2^{d_{\lin}+1}} \sqrt{\frac{1}{d_{\lin} T}} \sum_{i=1}^{d_{\lin}} \sum_{\vtheta^{j}_{Q,[-i]} \in \Theta_{d_{\lin}}} \sum_{\tau=1}^t \exp\left\{-\frac{2C_D t}{T}\right\}\\
    &= \frac{t}{4} \sqrt{\frac{d_{\lin}}{T}} \exp\left\{-\frac{2 C_D t}{T}\right\}
\end{aligned}
\end{equation}
where inequality $(i)$ follows from Equation~\eqref{eq:hammer:lin} and equality $(ii)$ follows from 
$$
\left\|\vtheta^{\beta}_{Q,[-i]} - \vtheta^{-\beta}_{Q,[-i]}\right\|_2^2 = 4\beta^2 = \frac{4d_{\lin}}{T}.
$$ 
Taking $t = T$ in Equation~\eqref{eq:Rlin:lower} yields the result in Equation~\eqref{eq:Rlin:lower:final}. 
\end{proof}

\subsubsection{Proof of Lemma~\ref{lem:fomega:smooth}}
\begin{proof}
To verify that $f_{\vomega} \in \calF_{\beta, L}$ for some suitable $L > 0$, we first note that for any $0\leq x, y \leq 1$
\begin{equation} \label{eq:lip:beta}
    \left|x^\beta - y^\beta\right| \leq \beta \left|x-y\right|.
\end{equation}
For $\ob, \ob' \in B_k$, by definition of $f_{\vomega}$ in Equation~\eqref{eq:fomega:def}, we have 
\begin{equation*}
\begin{aligned}
    \left|f_{\vomega}(\ob) - f_{\vomega}(\ob')\right| &= |\varphi_k(\ob) - \varphi_{k}(\ob')|\\
    &= M^{-\beta} C_{\phi} |\phi_\beta(2M[\ob-\bb_k]) - \phi_\beta(2M[\ob'-\bb_{k}])| \\
    &= M^{-\beta} C_{\phi} \left | (1 - \|2M[\ob-\bb_k]\|_{\infty})^\beta - (1 - \|2M[\ob'-\bb_{k}]\|_{\infty})^\beta \right |\\
\end{aligned}
\end{equation*}
where the second equality follows from the definition of $\varphi_k$ in Equation~\eqref{eq:varphi:def} and the last equality follows from the definition of $\phi_\beta$ in Equation~\eqref{eq:bump:def}. 
Continuing from the above display, 
\begin{equation}\label{eq:same:bin}
\begin{aligned}
\left|f_{\vomega}(\ob) - f_{\vomega}(\ob')\right|
    &= 2^\beta C_{\phi} \left | \left(\frac{1}{2M} - \|\ob-\bb_k\|_{\infty}\right)^\beta - \left(\frac{1}{2M} - \|\ob-\bb_k\|_{\infty}\right)^\beta \right |\\
    &\stackrel{(i)}{\leq} 2^\beta \beta C_{\phi} \left | \|\ob-\bb_k\|_{\infty} - \|\ob'-\bb_{k}\|_{\infty} \right | \\
    &\stackrel{(ii)}{\leq} 2^\beta \beta C_{\phi} \left\|\ob - \ob'\right\|_{\infty} \leq 2^\beta \beta C_{\phi} \left\|\ob - \ob'\right\|_{2} 
\end{aligned}
\end{equation}
where equality $(i)$ follows from Equation~\eqref{eq:lip:beta} and inequality $(ii)$ follows from the triangle inequality.

If $\ob, \ob'$ are in different bins $B_k, B_{k'}$, then we can pick $\pb_k \in B_k$ and $\pb_{k'} \in B_{k'}$ each on the boundary of $B_k$ and $B_{k'}$, such that both $f(\pb_k) = 0$ and $f(\pb_{k'}) = 0$, and 
\begin{equation} \label{eq:diff:bin:I}
\begin{aligned}
    \left|f(\ob) - f(\ob')\right| &\leq \max\left\{\left|f(\ob) - f(\pb_k)\right|, \left|f(\ob') - f(\pb_{k'})\right|\right\}\\
    &\leq 2^{\beta} \beta C_{\phi} \max\left\{\left\|\ob - \pb_k\right\|_{\infty}, \left\|\ob' - \pb_{k'}\right\|_{\infty}\right\}\\
\end{aligned}
\end{equation}
where the last inequality follows from Equation~\eqref{eq:same:bin}. 
We can pick $p_k$ and $p_{k'}$ so that 
\begin{equation*}
    \left\|\ob - \ob'\right\|_{\infty} \geq \max\left\{\left\|\ob - \pb_k\right\|_{\infty}, \left\|\ob' - \pb_{k'}\right\|_{\infty}\right\}
\end{equation*}
it then follows from Equation~\eqref{eq:diff:bin:I} that
\begin{equation*}
\begin{aligned}
    \left|f(\ob) - f(\ob')\right|
&\leq 2^{\beta} \beta C_{\phi} \left\|\ob - \ob'\right\|_{\infty} \leq 2^\beta \beta C_{\phi} \left\|\ob - \ob'\right\|_{2}. 
\end{aligned}
\end{equation*}
Combining the above display with Equation~\eqref{eq:same:bin} finishes the proof. 
\end{proof}

\subsubsection{Proof of Lemma~\ref{lem:non:regret}}

\begin{proof}
Let 
\begin{equation*}
    \tilde{B}_j = B_j \cap \left\{\ob: \phi_{\beta}(2M(\ob - \bb_j)) \geq \delta M^{\beta}\right\}.
\end{equation*}
For any $\ob \in \tilde{B}_j$, it follows from Equation~\eqref{eq:fomega:def} that 
\begin{equation} \label{eq:fomega:lower}
    f_{\vomega}(\ob) = \omega_j \varphi_j(\ob) \geq C_\phi M^{-\beta} \delta M^{\beta} = \delta C_{\phi}. 
\end{equation}
For any $\vomega \in \Omega_m$ and $\delta > 0$, combining Equation~\eqref{eq:fomega:lower} with $\calR_T^{\non}$ as defined in Equation~\eqref{eq:reward:decomp} yields that 
\begin{equation} \label{eq:nonparams:regret:II}
\begin{aligned}
    \calR^{(\non)}_T(\vtheta, f_{\vomega}) &= \sum_{t=1}^T\EE^{(t-1)}_{\vtheta, f_{\vomega}, \pi} \EE \left[\mathds{1}\left\{A_t \neq \sign\left(f_{\vomega}(\mO_t)\right), V_t = 0\right\} \left|f_{\vomega}(\mO_t)\right| \mid \calH_{t-1} \right]\\
    &= \sum_{t=1}^T \sum_{j=1}^m \EE^{(t-1)}_{\vtheta, f_{\vomega}, \pi} \EE \left[\mathds{1}\left\{A_t \neq \sign\left(f_{\vomega}(\mO_t)\right), V_t = 0\right\} \mathds{1}\left\{\mO_t \in B_j\right\} \left|f_{\vomega}(\mO_t)\right| \mid \calH_{t-1} \right] \\
    &\geq C_{\phi} \delta \sum_{j=1}^m \sum_{t=1}^T \EE^{(t-1)}_{\vtheta, f_{\vomega}, \pi} \EE \left[\mathds{1}\left\{A_t \neq \omega_j, \mO_t \in \tilde{B}_j , V_t = 0\right\} \mid \calH_{t-1} \right]. \\
\end{aligned}
\end{equation}
For any $\vomega \in \Omega_m$ and $\mO \sim \PP_O$, we have for any $\delta > 0$, 
\begin{equation} \label{eq:Btilde:prob}
\begin{aligned}
    \PP\left(\mO \in \tilde{B}_1\right) &=\PP\left(\phi_{\beta}\left(2M[\mO - \bb_1]\right) \geq \delta M^{\beta}, \mO \in B_1\right) \\
    &= \int_{B_1} \mathds{1}\left(\phi_{\beta}(2M (\ob - \bb_1)) \geq \delta M^\beta\right) d \ob\\
    &= \int_{B_1} \mathds{1}\left\{(1 -2M \|\ob - \bb_1\|_{\infty})^\beta \geq \delta M^\beta\right\} d \ob\\
    &= \int_{\left[0, \frac{1}{M}\right]^{d_{\non}}} \mathds{1}\left(\max_{l \in [d_{\non}]} \left|o_l - \frac{1}{2M}\right| \leq \frac{1}{2M} - \frac{1}{2}\delta^{1/\beta}\right) d \ob\\
    &= \int_{\left[0, \frac{1}{M}\right]^{d_{\non}}} \mathds{1}\left(\ob \in \left[\frac{1}{2}\delta^{1/\beta}, \frac{1}{M} - \frac{1}{2}\delta^{1/\beta}\right]^{d_{\non}}\right) d \ob\\
    &= \left(\frac{1}{M} - \delta^{1/\beta}\right)^{d_{\non}}. 
\end{aligned}
\end{equation}
The same probability holds for all other $\tilde{B}_j$ where $j \in [M^d]$. 

To handle $\calK^{(\non)}$ as defined in Equation~\eqref{eq:Knon:def}, take $\vomega$ and $\vomega'$ so that they only differ in $\omega_j$, we have 
\begin{equation*}
    \left|f_{\vomega}(\ob) - f_{\vomega'}(\ob)\right| = 2\varphi_j(\ob)
\end{equation*}
and
\begin{equation} \label{eq:cumulative:nonparams:bound:i}
\begin{aligned}
    &\EE^{(t-1)}_{\vtheta, f_{\vomega}, \pi} \EE \left[(f_{\vomega}(\mO_t) - f_{\vomega'}(\mO_t))^2 \mathds{1}\left\{A_t = 1, V_t = 0\right\} \mid \calH_{t-1}\right] \\
    &~~= 4\EE^{(t-1)}_{\vtheta, f_{\vomega}, \pi} \EE \left[\varphi_j(\mO_t)^2 \mathds{1}\left\{A_t = 1, \mO_t \in B_j\right\} \mid \calH_{t-1} \right]\\
    &~~\leq \frac{4C^2_{\phi}\delta^2}{M^{d_{\non}}} + 4\EE^{(t-1)}_{\vtheta, f_{\vomega}, \pi} \EE \left[\varphi_j(\mO_t)^2 \mathds{1}\left\{A_t = 1, \mO_t \in \tilde{B}_j\right\} \mid \calH_{t-1}\right]\\
    &~~\leq \frac{4C^2_{\phi}\delta^2}{M^{d_{\non}}} + 4C_{\phi}^2 M^{-2\beta} \EE^{(t-1)}_{\vtheta, f_{\vomega}, \pi} \EE \left[\mathds{1}\left\{A_t = 1, \mO_t \in \tilde{B}_j\right\} \mid \calH_{t-1} \right]\\
\end{aligned}
\end{equation}
where the first equality follows from the fact that $\mO_t \in B_j$ already implies $V_t = 0$. 
Similarly, for any $i \in [N]$, applying the above argument with Equation~\eqref{eq:Btilde:prob} to the pretrained data yields 
\begin{equation} \label{eq:pretrain:bound}
\begin{aligned}
    \EE_{f_{\vomega}}\left[\left(f_{\vomega}\left(\mO_i^{(0)}\right) - f_{\vomega'}\left(\mO_i^{(0)}\right)\right)^2\right] 
    &\leq \frac{4C^2_{\phi}\delta^2}{M^{d_{\non}}} +  4C^{2}_{\phi} M^{-2\beta} \left(\frac{1}{M}-\delta^{1 / \beta}\right)^{d_{\non}}.
\end{aligned}
\end{equation}
Pick $\delta_0$ so that 
\begin{equation*}
    M^{2\beta} \delta_0^2 \asymp \left(1 - M \delta_0^{1 / \beta}\right)^{d_{\non}}.
\end{equation*}
Let $\kappa_0$ be the solution to the equation
\begin{equation*}
    \kappa^{2\beta} = (1-\kappa)^{d_{\non}}
\end{equation*}
then we set 
\begin{equation} \label{eq:delta:def}
    \delta_0 = \kappa_0^{\beta} M^{-\beta}.
\end{equation}
Under the assumption that $d_{\non}$ is a fixed constant in Lemma~\ref{lem:non:regret}, we have $\kappa_0$ is also a constant and $\delta_0 = \Theta(M^{-\beta})$. 
Under Equation~\eqref{eq:delta:def}, the bound in Equation~\eqref{eq:cumulative:nonparams:bound:i} becomes 
\begin{equation} \label{eq:cumulative:nonparams:bound:ii}
\begin{aligned}
    &\EE^{(t-1)}_{\vtheta, f_{\vomega}, \pi} \EE \left[(f_{\vomega}(\mO_t) - f_{\vomega'}(\mO_t))^2 \mathds{1}\left\{A_t = 1, V_t = 0\right\} \mid \calH_{t-1}\right] \\
    &~~\lesssim M^{-2\beta-d_{\non}} + M^{-2\beta} \EE^{(t-1)}_{\vtheta, f_{\vomega}, \pi} \EE \left[\mathds{1}\left\{A_t = 1, \mO_t \in \tilde{B}_j\right\} \mid \calH_{t-1} \right]\\
\end{aligned}
\end{equation}
and Equation~\eqref{eq:pretrain:bound} becomes
\begin{equation}\label{eq:pretrain:bound:kappa}
    \EE_{f_{\vomega}}\left[\left(f_{\vomega}\left(\mO_i^{(0)}\right) - f_{\vomega'}\left(\mO_i^{(0)}\right)\right)^2\right] 
    \lesssim M^{-2\beta-d_{\non}}.
\end{equation}
Combining Equations~\eqref{eq:cumulative:nonparams:bound:ii} and~\eqref{eq:pretrain:bound:kappa} with Equation~\eqref{eq:Knon:def}, for any $t \in [T]$ and the choice of $\delta_0$ given in Equation~\eqref{eq:delta:def}, 
\begin{equation} \label{eq:Knon:bound}
\begin{aligned}
    \calK_{\vtheta,t}^{(\non)}\left(f_{\vomega},f_{\vomega'}\right) &\lesssim M^{-2\beta} \sum_{\tau=1}^t \EE^{(\tau-1)}_{\vtheta, f_{\vomega}, \pi} \EE \left[\mathds{1}\left\{A_\tau = 1, \mO_\tau \in \tilde{B}_j\right\} \mid \calH_{\tau-1} \right] + \frac{(t+N)}{M^{2\beta+d_{\non}}}.
\end{aligned}
\end{equation}

Using an average hammer over $\vomega \in \Omega_m$, it follows from Equation~\eqref{eq:nonparams:regret:II} and the choice of $\delta_0$ in Equation~\eqref{eq:delta:def} that 
\begin{equation}\label{eq:nonparam:regret:lower:I}
\begin{aligned}
    \sup_{f \in \calF_{\beta,L}} \calR^{(\non)}_T(\vtheta, f) &\gtrsim M^{-\beta} \sup_{\vomega \in \Omega_m} \sum_{j=1}^m \sum_{t=1}^T \EE^{(t-1)}_{\vtheta, f_{\vomega}, \pi} \EE \left[\mathds{1}\left\{A_t \neq \omega_j, \mO_t \in \tilde{B}_j\right\} \mid \calH_{t-1}\right] \\
    &\geq 2^{-m} M^{-\beta} \sum_{\vomega \in \Omega_m} \sum_{j=1}^m \sum_{t=1}^T \EE^{(t-1)}_{\vtheta, f_{\vomega}, \pi} \EE \left[\mathds{1}\left\{A_t \neq \omega_j, \mO_t \in \tilde{B}_j\right\} \mid \calH_{t-1} \right],
\end{aligned}
\end{equation}
where the last inequality follows from $|\Omega_m| = 2^m$. 
Let 
\begin{equation} \label{eq:G:def}
    G_j^t := \sum_{\vomega_{[-j]} \in \Omega_{m-1}} \sum_{i \in \{\pm 1\}} \EE^{(t-1)}_{\vtheta, f_{\vomega^i_{[-j]}}, \pi} \EE \left[\mathds{1}\left\{A_t \neq i, \mO_t \in \tilde{B}_j\right\} \mid \calH_{t-1} \right],
\end{equation}
where we group $\vomega_{[-j]}^{1}$ and $\vomega_{[-j]}^{-1}$ together in the inner sum.
Taking Equation~\eqref{eq:G:def} into Equation~\eqref{eq:nonparam:regret:lower:I}, we have 
\begin{equation} \label{eq:nonparam:regret:lower:II}
    \sup_{f \in \calF_{\beta,L}} \calR^{(\non)}_T(\vtheta, f) \gtrsim 2^{-m} M^{-\beta} \sum_{j=1}^m \sum_{t=1}^T G_{j}^t. 
\end{equation}
We pause to provide some intuition for introducing $G_j^t$. 
The idea is that we would like to apply Bretagnolle-Huber inequality as stated in Theorem~\ref{thm:bh:ineq} to obtain a lower bound of the cumulative regret.  
To get a tighter lower bound, we would group the most similar pairs of $\vomega, \vomega' \in \Omega_m$ together to minimize the KL divergence between the two probability measures indexed by $\vomega$ and $\vomega'$.  

By Equation~\eqref{eq:Btilde:prob} and the definition of $\delta_0$ in Equation~\eqref{eq:delta:def}, 
\begin{equation}\label{eq:nonparams:single-term}
\begin{aligned}
    \EE^{(t-1)}_{\vtheta, f_{\vomega^i_{[-j]}}, \pi} \EE \left[\mathds{1}\left\{A_t \neq i, \mO_t \in \tilde{B}_j\right\} \mid \calH_{t-1}\right] &\asymp \frac{1}{M^{d_{\non}}} \EE^{(t-1)}_{\vtheta, f_{\vomega^i_{[-j]}}, \pi} \EE \left[\mathds{1}\left\{A_t \neq i\right\} \mid \calH_{t-1}, \mO_t \in \tilde{B}_j\right].
\end{aligned}
\end{equation}
Denote by $\PP^{(t-1)}_j$ the conditional probability $\PP\left(\cdot \mid \calH_{t-1}, \mO_t \in \tilde{B}_j\right)$. 
We apply Bretagnolle-Huber inequality as stated in Theorem~\ref{thm:bh:ineq} and obtain 
\begin{equation} \label{eq:hammer:non}
\begin{aligned}
    &\sum_{i \in \{\pm 1\}} \EE^{(t-1)}_{\vtheta, f_{\vomega^i_{[-j]}}, \pi} \EE \left[\mathds{1}\left\{A_t \neq i\right\} \mid \calH_{t-1}, \mO_t \in \tilde{B}_j\right] \\
    &~~\geq \frac{1}{2}\exp\left[-\KL\left(\PP^{(t-1)}_{\vtheta, f_{\vomega^{1}_{[-j]}}, \pi} \times \PP^{(t-1)}_j \| \PP^{(t-1)}_{\vtheta, f_{\vomega^{-1}_{[-j]}}, \pi} \times \PP^{(t-1)}_j\right)\right]\\
    &~~= \frac{1}{2}\exp\left[-\KL\left(\PP^{(t-1)}_{\vtheta, f_{\vomega^{1}_{[-j]}}, \pi} \| \PP^{(t-1)}_{\vtheta, f_{\vomega^{-1}_{[-j]}}, \pi}\right)\right]\\
    &~~\geq \frac{1}{2}\exp\left[-\calK_{\vtheta,t}^{(\non)}\left(f_{\vomega^{1}_{[-j]}}, f_{\vomega^{-1}_{[-j]}}\right)\right]\\
\end{aligned}
\end{equation}
where the last inequality follows from Equations~\eqref{eq:KL:bound:II}, \eqref{eq:Knon:def} and~\eqref{eq:Klin:def}. 
Taking Equations~\eqref{eq:nonparams:single-term} and~\eqref{eq:hammer:non} into Equation~\eqref{eq:G:def} yields that 
\begin{equation} \label{eq:G:lower}
\begin{aligned}
    G_{j}^t &\gtrsim M^{-d_{\non}} \sum_{\vomega_{[-j]} \in \Omega_{m-1}} \exp\left[-\calK_{\vtheta, t}^{(\non)}\left(f_{\vomega^{1}_{[-j]}},f_{\vomega^{-1}_{[-j]}}\right)\right]\\
    &\stackrel{(i)}{\geq} \frac{1}{M^{d_{\non}}} \sum_{\vomega_{[-j]} \in \Omega_{m-1}} \exp\left(- \frac{C}{M^{2\beta}} \sum_{\tau=1}^{t} \EE^{(\tau-1)}_{\vtheta, f^{1}_{\vomega_{[-j]}}, \pi} \EE \left[\mathds{1}\left\{A_\tau = -1, \mO_\tau \in \tilde{B}_j\right\} \mid \calH_{\tau-1}\right] - \frac{C (t+N)}{M^{2\beta+d_{\non}}}\right)\\
    &\geq \frac{1}{M^{d_{\non}}} \sum_{\vomega_{[-j]} \in \Omega_{m-1}} \exp\left(-\frac{C}{M^{2\beta}} \sum_{\tau=1}^{T} \EE^{(\tau-1)}_{\vtheta, f^{1}_{\vomega_{[-j]}}, \pi} \EE \left[\mathds{1}\left\{A_\tau = -1, \mO_\tau \in \tilde{B}_j\right\} \mid \calH_{\tau-1}\right] - \frac{C (T+N)}{M^{2\beta+d_{\non}}}\right)\\
    &\stackrel{(ii)}{\geq} \frac{2^{m-1}}{M^{d_{\non}}}\exp\left(-\frac{C}{M^{2\beta}2^{m-1}} \sum_{\vomega_{[-j]} \in \Omega_{m-1}} \sum_{\tau=1}^{T} \EE^{(\tau-1)}_{\vtheta, f^{1}_{\vomega_{[-j]}}, \pi} \EE \left[\mathds{1}\left\{A_\tau = -1, \mO_\tau \in \tilde{B}_j\right\} \mid \calH_{\tau-1}\right] - \frac{C (T+N)}{M^{2\beta+d_{\non}}}\right)
\end{aligned}
\end{equation}
where inequality $(i)$ follows from Equation~\eqref{eq:Knon:bound} and inequality $(ii)$ follows from Jensen's inequality. 
Let 
\begin{equation*}
    E_{j,\pi} := \frac{1}{2^{m-1}} \sum_{\vomega_{[-j]} \in \Omega_{m-1}} \sum_{\tau=1}^T \EE^{(\tau-1)}_{\vtheta, f^{1}_{\vomega_{[-j]}}, \pi} \EE \left[\mathds{1}\left\{A_\tau = 1, \mO_\tau \in \tilde{B}_j\right\} \mid \calH_{\tau-1} \right]
\end{equation*}
and taking $E_{j,\pi}$ into Equation~\eqref{eq:G:lower}, we have
\begin{equation} \label{eq:Gjt:lower:i}
    G_j^t \gtrsim \frac{2^{m-1}}{M^{d_{\non}}} \exp\left(-C M^{-2\beta} E_{j,\pi} - C (T+N) M^{-2\beta-d_{\non}}\right)
\end{equation}
From the definition of $G_j^t$ in Equation~\eqref{eq:G:def}, we also have
\begin{equation} \label{eq:Gjt:lower:ii}
    \sum_{t=1}^T G_j^t \geq 2^{m-1} E_{j,\pi}. 
\end{equation}
Taking Equations~\eqref{eq:Gjt:lower:i} and~\eqref{eq:Gjt:lower:ii} into Equation~\eqref{eq:nonparam:regret:lower:II} yields 
\begin{equation*} 
\begin{aligned}
    \sup_{f \in \calF_{\beta,L}} \calR^{(\non)}_T(\vtheta, f) 
    &\gtrsim 2^{-m} M^{-\beta} \sum_{j=1}^m \sum_{t=1}^T G_j^t \\
    &\geq \frac{1}{2} M^{-\beta} \sum_{j=1}^m \max\left\{E_{j,\pi}, \frac{1}{M^{d_{\non}}} \exp\left(-C M^{-2\beta} E_{j,\pi} - \frac{C (T+N)}{M^{2\beta+d_{\non}}}\right)\right\}\\
    &\geq \frac{1}{4} M^{-\beta} \sum_{j=1}^m \left\{E_{j,\pi} + \frac{T}{M^{d_{\non}}} \exp\left(-C M^{-2\beta} E_{j,\pi} - \frac{C (T+N)}{M^{2\beta+d_{\non}}} \right)\right\}\\
    &\gtrsim \inf_{z \geq 0} M^{-\beta} \sum_{j=1}^m \left\{z + \frac{T}{M^{d_{\non}}} \exp\left[-C M^{-2\beta} z - \frac{C (T+N)}{M^{2\beta+d_{\non}}}\right]\right\}.
\end{aligned}
\end{equation*}
The case where $N = \Theta(T)$ can be handled similarly as the analysis below and we omit the details here.
We focus on the case where $N \gg T$ and the above display can be simplified into
\begin{equation} \label{eq:nonparam:regret:lower:III}
\begin{aligned}
    \sup_{f \in \calF_{\beta,L}} \calR^{(\non)}_T(\vtheta, f) 
    &\gtrsim \inf_{z \geq 0} m M^{-\beta} \left\{z + \frac{T}{M^{d_{\non}}} \exp\left[-C M^{-2\beta} z - \frac{C N}{M^{2\beta+d_{\non}}}\right]\right\}\\
    &\gtrsim \inf_{z \geq 0} M^{-\beta+d_{\non}} \left\{z + \frac{T\alpha}{M^{d_{\non}}} \exp\left[-C M^{-2\beta} z\right]\right\}\\
\end{aligned}
\end{equation}
where in the last inequality, we use the definition of $m$ as in Equation~\eqref{eq:m:def} and let
\begin{equation*}
    \alpha := \exp\left(-\frac{C N}{M^{2\beta+d_{\non}}}\right).
\end{equation*}
The minimizer of the right-hand side of Equation~\eqref{eq:nonparam:regret:lower:III} over $z \in \R$ is given by
\begin{equation} \label{eq:zstar:def}
    z^\star = \frac{M^{2 \beta}}{C}\log \left(\frac{C T \alpha}{M^{2\beta+d_{\non}}}\right) = \frac{M^{2 \beta}}{C}\log \left(\frac{C T}{M^{2\beta+d_{\non}}}\right) - \frac{N}{M^{d_{\non}}}. 
\end{equation}
For $z^\star \geq 0$ to hold, we need 
\begin{equation} \label{eq:global:condition}
    M^{2 \beta + d_{\non}} \log \left(\frac{C T}{M^{2\beta+d_{\non}}}\right) \geq C N.
\end{equation}
Noting that when $M^{2\beta + d_{\non}} > C T$, the left hand side of the above display is negative. 
Thus, for Equation~\eqref{eq:global:condition} to hold, we must have $M^{2\beta + d_{\non}} = O(T)$, implying that
\begin{equation*}
    M^{2 \beta + d_{\non}} \log \left(\frac{C T}{M^{2\beta+d_{\non}}}\right) = O(T). 
\end{equation*}
The maximizer of the left hand side of the above display is given by 
\begin{equation} \label{eq:M:choice}
    M^{2\beta+d_{\non}} = \frac{C T}{e}.
\end{equation}
When $T \geq C N$ for some constant $C$ sufficiently large, $z^\star \geq 0$ holds. Taking $z = z^\star$ in Equation~\eqref{eq:nonparam:regret:lower:III} yields
\begin{equation*}
    \sup_{f \in \calF_{\beta,L}} \calR^{(\non)}_T(\vtheta, f) 
    \gtrsim M^{\beta+d_{\non}}\left[\log \left(\frac{T C}{M^{2 \beta+d_{\non}}}\right)+1\right]-\frac{N}{M^{d_{\non}}} = \Theta\left(T^{\frac{\beta+d_{\non}}{2\beta+d_{\non}}}\right)
\end{equation*}
where the last equality holds from the choice of $M$ in Equation~\eqref{eq:M:choice}. 

When $T \ll N$, we have $z^\star < 0$. 
Noting that for any constants $a,b > 0$, the function
\begin{equation*}
    h(z) = z + a \exp(-bz)
\end{equation*}
attains its minimum at 
\begin{equation*}
    z_0 = \frac{\log(ab)}{b}, 
\end{equation*}
and is monotonically increasing when $z > z_0$, it follows that the minimizer of $h(z)$ over $z \geq 0$ when $z_0 < 0$ is attained at $z = 0$.
Comparing the form of the right-hand side of Equation~\eqref{eq:nonparam:regret:lower:III} to $h(z)$ defined above yields that the minimizer is attained at $z = 0$ and 
\begin{equation}\label{eq:nonparams:regret:lower:IV}
    \sup_f \calR^{(\non)}_T(\vtheta, f) 
    \geq \frac{T}{M^{\beta}} \exp\left(-\frac{C N}{M^{2\beta+d_{\non}}}\right). 
\end{equation}
Let 
\begin{equation*}
    g(M) := -\beta \log M + \log T - C N M^{-2\beta - d_{\non}},
\end{equation*}
we have 
\begin{equation*}
    g'(M) = -\frac{\beta}{M} + \frac{C (2\beta+d_{\non}) N}{M^{2\beta+d_{\non}+1}}.
\end{equation*}
It attains its maximum at 
\begin{equation}\label{eq:max-M}
    M = \left[\frac{C(2\beta+d_{\non})N}{\beta}\right]^{\frac{1}{2\beta+d_{\non}}} = \Theta\left(N^{\frac{1}{2\beta+d_{\non}}}\right).
\end{equation}
Taking Equation~\eqref{eq:max-M} into the right-hand side of Equation~\eqref{eq:nonparams:regret:lower:IV} yields the desired result in Equation~\eqref{eq:Rnon:bound}. 
\end{proof}

\section{Derivation of Equivalent Formulations for UCB Exploration in Algorithm~\ref{alg:linucb}}

This section demonstrates that the Upper Confidence Bound (UCB) exploration strategy used in the LinUCB algorithm can be expressed in two equivalent forms. We first derive the general relationship between the exploration parameter $\alpha$ and the confidence set parameter $\gamma_t$ in the LinUCB algorithm. We then show how this relationship leads to an adaptive exploration schedule in our specific context.

The key variables are defined as follows:
\begin{itemize}
    \item $\alpha_t$: The exploration hyperparameter at timestep $t$.
    \item $\gamma_t$: A parameter controlling the size of the confidence ellipsoid at timestep $t$.
    \item $\boldsymbol{x}_{t,a} \in \mathbb{R}^d$: The context vector for action $a \in \mathcal{A}$ at timestep $t$.
    \item $\hat{\boldsymbol{\theta}}_{t-1} \in \mathbb{R}^d$: The ridge regression estimate of the parameter vector at the end of timestep $t-1$.
    \item $\boldsymbol{\Sigma}_{t-1} \in \mathbb{R}^{d \times d}$: The design matrix, defined as $\boldsymbol{\Sigma}_{t-1} = \lambda \boldsymbol{I} + \sum_{t=1}^{t-1} \boldsymbol{x}_{t, A_t} \boldsymbol{x}_{t, A_t}^\top$.
    \item $\BALL_{t-1}$: The confidence ellipsoid for the true parameter vector $\boldsymbol{\theta}^*$ at timestep $t-1$. It is defined as:
    \[
        \BALL_{t-1} = \left\{ \boldsymbol{\theta} \in \mathbb{R}^d \mid (\boldsymbol{\theta} - \hat{\boldsymbol{\theta}}_{t-1})^\top \boldsymbol{\Sigma}_{t-1} (\boldsymbol{\theta} - \hat{\boldsymbol{\theta}}_{t-1}) \le \gamma_{t-1} \right\}
    \]
\end{itemize}

The LinUCB algorithm can be formulated from two equivalent perspectives.

\paragraph{1. The $\alpha$-based UCB formulation:}
The action $A_t$ is chosen to maximize an upper confidence bound on the expected reward:
\begin{equation}
    A_t = \arg\max_{a \in \mathcal{A}} \left( \hat{\boldsymbol{\theta}}_{t-1}^\top \boldsymbol{x}_{t,a} + \alpha_{t-1} \sqrt{\boldsymbol{x}_{t,a}^\top \boldsymbol{\Sigma}_{t-1}^{-1} \boldsymbol{x}_{t,a}} \right)
    \label{eq:alpha_form}
\end{equation}

\paragraph{2. The confidence set formulation:}
The action $A_t$ is chosen by finding the most optimistic parameter vector within the confidence set for each action, and then selecting the action with the highest optimistic reward:
\begin{equation}
    A_t = \arg\max_{a \in \mathcal{A}} \max_{\boldsymbol{\theta} \in \text{BALL}_{t-1}} \boldsymbol{\theta}^\top \boldsymbol{x}_{t,a}
    \label{eq:ball_form}
\end{equation}

Our goal is to show the equivalence of the objective functions in \eqref{eq:alpha_form} and \eqref{eq:ball_form}. We focus on solving the inner maximization problem in \eqref{eq:ball_form}:
\[
\max_{\boldsymbol{\theta}} \quad \boldsymbol{\theta}^\top \boldsymbol{x}_{t,a} \quad \text{subject to} \quad \boldsymbol{\theta} \in \text{BALL}_{t-1}
\]
Let's introduce a change of variables: $\boldsymbol{z} = \boldsymbol{\theta} - \hat{\boldsymbol{\theta}}_{t-1}$, which implies $\boldsymbol{\theta} = \boldsymbol{z} + \hat{\boldsymbol{\theta}}_{t-1}$. The optimization problem becomes:
\begin{align*}
    \max_{\boldsymbol{z}} \quad & (\boldsymbol{z} + \hat{\boldsymbol{\theta}}_{t-1})^\top \boldsymbol{x}_{t,a} \\
    \text{subject to} \quad & \boldsymbol{z}^\top \boldsymbol{\Sigma}_{t-1} \boldsymbol{z} \le \gamma_{t-1}
\end{align*}
The objective function can be split into two parts: $\boldsymbol{z}^\top \boldsymbol{x}_{t,a} + \hat{\boldsymbol{\theta}}_{t-1}^\top \boldsymbol{x}_{t,a}$. Since $\hat{\boldsymbol{\theta}}_{t-1}^\top \boldsymbol{x}_{t,a}$ is constant with respect to $\boldsymbol{z}$, we only need to maximize $\boldsymbol{z}^\top \boldsymbol{x}_{t,a}$.

The problem is now $\max_{\boldsymbol{z}} \boldsymbol{z}^\top \boldsymbol{x}_{t,a}$ subject to $\boldsymbol{z}^\top \boldsymbol{\Sigma}_{t-1} \boldsymbol{z} \le \gamma_{t-1}$.
By the generalized Cauchy-Schwarz inequality, which states $(u^\top v)^2 \le (u^\top M u)(v^\top M^{-1} v)$ for a positive definite matrix $M$, we can set $u = \boldsymbol{z}$, $v = \boldsymbol{x}_{t,a}$, and $M = \boldsymbol{\Sigma}_{t-1}$.
This gives:
\[
(\boldsymbol{z}^\top \boldsymbol{x}_{t,a})^2 \le (\boldsymbol{z}^\top \boldsymbol{\Sigma}_{t-1} \boldsymbol{z}) (\boldsymbol{x}_{t,a}^\top \boldsymbol{\Sigma}_{t-1}^{-1} \boldsymbol{x}_{t,a})
\]
Using our constraint $\boldsymbol{z}^\top \boldsymbol{\Sigma}_{t-1} \boldsymbol{z} \le \gamma_{t-1}$, we get:
\[
(\boldsymbol{z}^\top \boldsymbol{x}_{t,a})^2 \le \gamma_{t-1} (\boldsymbol{x}_{t,a}^\top \boldsymbol{\Sigma}_{t-1}^{-1} \boldsymbol{x}_{t,a})
\]
Taking the square root, the maximum value for $\boldsymbol{z}^\top \boldsymbol{x}_{t,a}$ is:
\[
\max_{\boldsymbol{z}} \boldsymbol{z}^\top \boldsymbol{x}_{t,a} = \sqrt{\gamma_{t-1}} \sqrt{\boldsymbol{x}_{t,a}^\top \boldsymbol{\Sigma}_{t-1}^{-1} \boldsymbol{x}_{t,a}}
\]
Substituting this back into the full objective function, we have:
\[
\max_{\boldsymbol{\theta} \in \BALL_{t-1}} \boldsymbol{\theta}^\top \boldsymbol{x}_{t,a} = \hat{\boldsymbol{\theta}}_{t-1}^\top \boldsymbol{x}_{t,a} + \sqrt{\gamma_{t-1}} \sqrt{\boldsymbol{x}_{t,a}^\top \boldsymbol{\Sigma}_{t-1}^{-1} \boldsymbol{x}_{t,a}}
\]
By comparing this result with the objective function in \eqref{eq:alpha_form}, we can directly establish the relationship:
\[
\alpha_{t-1} = \sqrt{\gamma_{t-1}}
\]

\subsection{Implication for Adaptive Exploration}

This equivalence enables us to understand how the adaptive nature of the confidence set, defined by $\gamma_t$, is directly translated into the exploration parameter $\alpha_t$.

Given the definition of $\gamma_t$ from Theorem \ref{thm:regret:upper:bound}:
\begin{equation}
\gamma_t := \gamma_t^{(0)} + 3d^2 \sum_{t=1}^t D_{t}
\end{equation}
where $\gamma_t^{(0)}$ captures the baseline uncertainty from stochastic noise, the relationship is:
\begin{equation}
\alpha_t = \sqrt{\gamma_t^{(0)} + 3d^2 \sum_{t=1}^t D_{t}}
\end{equation}
Expanding the $\gamma_t^{(0)}$ term, we get the complete expression:
\begin{equation}
\alpha_t = \sqrt{ \left( 3\lambda + 6(\sigma_{\eta} + \sigma_{\varepsilon})^2 \log\left[\frac{4t^2}{\delta}\left(1 + \frac{tB^2}{d\lambda}\right)^d\right] \right) + 3d^2 \sum_{t=1}^t D_{t} }
\end{equation}

This equation shows that the exploration parameter $\alpha_t$ is adaptive. It increases not only due to inherent stochasticity (the $\gamma_t^{(0)}$ term) but also in response to the accumulated uncertainty in context estimation over all past timesteps (the $\sum D_{t}$ term).

\section{Synthetic Data Experiments}
\subsection{Extra Baseline Comparisons}
\label{app:additional_baseline}

In this section, we provide extended empirical results to complement the main experiments. Specifically, we systematically evaluate the dependency of PULSE-UCB on the pre-training data scales and present an expanded benchmark against CLBBF in both linear and nonlinear settings.

\subsubsection{Sensitivity to Pre-training Data Scale \texorpdfstring{$N$ and $T_0$}{(N and T0)}}
\label{app:sensitivity_data_scale}

To evaluate the dependency on pre-training data, we adopted a high-dimensional setting where the unobserved context $W_t = \boldsymbol{\beta}_{*}^\top \bx_{t} + \xi_{t}$ depends on history $\bx_{t}=(1, \mS_{t-1}, \dots, \mS_{t-19})^\top \in \mathbb{R}^{20}$, while $\{\mS_t\}$ remains an $\mathrm{ARMA}(2,2)$ process. The reward is updated to $\mPhi(\mY_t, a_t) = (1, \mS_t, W_t, W_t \cdot a_t)^\top$. 

Tables~\ref{tab:regret_N} and \ref{tab:regret_T0} summarize the results under varying sample sizes $N$ and trajectory lengths $T_0$.

\begin{table}[htbp]
    \centering
    \caption{Final cumulative regret vs. Pre-training Size $N$ in the linear setting.}
    \label{tab:regret_N}
    \begin{tabular}{r c c c c}
        \toprule
        $N$ & PULSE-UCB & OFUL-Full & OFUL & CLBBF \\
        \midrule
        1    & $252.1 \pm 87.5$ & $2.7 \pm 2.0$ & $333.7 \pm 30.2$ & $145.3 \pm 9.5$ \\
        2    & $150.3 \pm 77.0$ & $2.7 \pm 2.0$ & $333.7 \pm 30.2$ & $145.3 \pm 9.5$ \\
        3    & $61.3 \pm 91.9$  & $2.7 \pm 2.0$ & $333.7 \pm 30.2$ & $145.3 \pm 9.5$ \\
        5    & $11.7 \pm 9.6$   & $2.7 \pm 2.0$ & $333.7 \pm 30.2$ & $145.3 \pm 9.5$ \\
        10   & $4.6 \pm 1.4$    & $2.7 \pm 2.0$ & $333.7 \pm 30.2$ & $145.3 \pm 9.5$ \\
        50   & $4.2 \pm 1.6$    & $2.7 \pm 2.0$ & $333.7 \pm 30.2$ & $145.3 \pm 9.5$ \\
        500  & $4.3 \pm 1.6$    & $2.7 \pm 2.0$ & $333.7 \pm 30.2$ & $145.3 \pm 9.5$ \\
        1000 & $4.0 \pm 1.4$    & $2.7 \pm 2.0$ & $333.7 \pm 30.2$ & $145.3 \pm 9.5$ \\
        \bottomrule
    \end{tabular}
\end{table}

\begin{table}[htbp]
    \centering
    \caption{Final cumulative regret vs. Pre-training Trajectory Length $T_0$ in the linear setting.}
    \label{tab:regret_T0}
    \begin{tabular}{r c c c c}
        \toprule
        $T_0$ & PULSE-UCB & OFUL-Full & OFUL & CLBBF \\
        \midrule
        21  & $248.7 \pm 163.8$ & $2.7 \pm 2.0$ & $333.7 \pm 30.2$ & $145.3 \pm 9.5$ \\
        25  & $46.9 \pm 55.7$   & $2.7 \pm 2.0$ & $333.7 \pm 30.2$ & $145.3 \pm 9.5$ \\
        30  & $8.6 \pm 2.7$     & $2.7 \pm 2.0$ & $333.7 \pm 30.2$ & $145.3 \pm 9.5$ \\
        40  & $5.1 \pm 2.2$     & $2.7 \pm 2.0$ & $333.7 \pm 30.2$ & $145.3 \pm 9.5$ \\
        50  & $4.3 \pm 1.4$     & $2.7 \pm 2.0$ & $333.7 \pm 30.2$ & $145.3 \pm 9.5$ \\
        100 & $4.4 \pm 2.2$     & $2.7 \pm 2.0$ & $333.7 \pm 30.2$ & $145.3 \pm 9.5$ \\
        200 & $4.6 \pm 1.5$     & $2.7 \pm 2.0$ & $333.7 \pm 30.2$ & $145.3 \pm 9.5$ \\
        \bottomrule
    \end{tabular}
\end{table}

\textbf{Discussion on Data Scaling:} 
As clearly demonstrated in Tables~\ref{tab:regret_N} and \ref{tab:regret_T0}, PULSE-UCB exhibits a sharp phase transition in both cases. When pre-training data is scarce (e.g., $N=1$ or $T_0=21$), the performance is limited and behaves similarly to the naive OFUL baseline. However, increasing either $N$ (with a fixed $T_0=25$) or $T_0$ (with a fixed $N=3$) beyond a modest threshold leads to a rapid drop in cumulative regret. In this stable and data-sufficient regime, PULSE-UCB demonstrates exceptionally high sample efficiency, achieving near-oracle performance that matches OFUL-Full (which operates with fully observed contexts) and significantly outperforming CLBBF.

\subsubsection{Expanded Comparison with CLBBF}
\label{app:expanded_clbbf}

In the main text, CLBBF was primarily evaluated on the real-world dataset as that specific dataset was a suitable testbed for CLBBF's underlying assumptions. We clarify that CLBBF was originally omitted from the synthetic benchmarks because it assumes a \emph{Missing-At-Random} (MAR) structure, whereas our synthetic setting involves structurally missing (MNAR) covariates. Following the reviewer's suggestion, we have now added CLBBF to the synthetic analysis to explicitly illustrate the limitations of the MAR assumption in our environment. 

Tables~\ref{tab:comp_linear} and \ref{tab:comp_nonlinear} present the time-evolution of cumulative regret up to $T=1000$ in both linear and nonlinear reward environments.

\begin{table}[htbp]
    \centering
    \caption{Cumulative Regret Comparison of algorithms in the linear setting.}
    \label{tab:comp_linear}
    \begin{tabular}{r c c c c}
        \toprule
        Time step & PULSE-UCB & OFUL & CLBBF & OFUL-Full \\
        \midrule
        1    & $0.0 \pm 0.0$   & $0.0 \pm 0.0$   & $0.0 \pm 0.0$   & $0.0 \pm 0.0$ \\
        10   & $0.1 \pm 0.2$   & $0.2 \pm 0.2$   & $0.1 \pm 0.2$   & $0.2 \pm 0.2$ \\
        100  & $2.8 \pm 0.6$   & $3.6 \pm 0.8$   & $1.2 \pm 0.6$   & $3.0 \pm 0.6$ \\
        300  & $6.1 \pm 0.8$   & $11.6 \pm 1.2$  & $4.7 \pm 1.0$   & $6.0 \pm 0.9$ \\
        800  & $10.5 \pm 1.3$  & $31.6 \pm 1.7$  & $16.0 \pm 1.5$  & $9.8 \pm 1.4$ \\
        1000 & $11.9 \pm 1.4$  & $39.6 \pm 2.1$  & $20.6 \pm 1.4$  & $11.1 \pm 1.7$ \\
        \bottomrule
    \end{tabular}
\end{table}

\begin{table}[htbp]
    \centering
    \caption{Cumulative Regret Comparison of algorithms in the nonlinear setting.}
    \label{tab:comp_nonlinear}
    \begin{tabular}{r c c c c}
        \toprule
        Time step & PULSE-UCB & OFUL & CLBBF & OFUL-Full \\
        \midrule
        1    & $0.0 \pm 0.0$   & $0.0 \pm 0.0$   & $0.0 \pm 0.0$   & $0.0 \pm 0.0$ \\
        10   & $0.1 \pm 0.2$   & $0.2 \pm 0.2$   & $0.2 \pm 0.2$   & $0.2 \pm 0.2$ \\
        100  & $2.3 \pm 0.6$   & $2.8 \pm 0.7$   & $2.2 \pm 0.5$   & $2.9 \pm 0.6$ \\
        300  & $5.4 \pm 0.8$   & $10.6 \pm 1.3$  & $5.8 \pm 0.8$   & $6.0 \pm 0.7$ \\
        800  & $11.6 \pm 1.6$  & $30.6 \pm 1.8$  & $15.3 \pm 1.4$  & $9.7 \pm 1.4$ \\
        1000 & $13.7 \pm 1.7$  & $38.5 \pm 2.1$  & $19.2 \pm 1.5$  & $11.0 \pm 1.8$ \\
        \bottomrule
    \end{tabular}
\end{table}

\textbf{Discussion on Baselines and MAR/MNAR Structures:}
Consistent with the real-world results, CLBBF outperforms the naive OFUL algorithm but consistently lags behind PULSE-UCB. The results confirm our hypothesis: online feature estimation methods like CLBBF struggle in this MNAR regime where covariates are structurally missing. Conversely, the time-evolution analysis demonstrates that by $T=1000$, PULSE-UCB rapidly converges to the oracle (OFUL-Full) performance in both linear and nonlinear settings. This highlights that our pre-training approach successfully bridges the information gap caused by MNAR structures, maintaining near-oracle performance where traditional MAR-based methods falter.
\subsection{Impact of Smoothness}
\begin{figure}[H]
   \centering
   \includegraphics[width=0.79\linewidth]{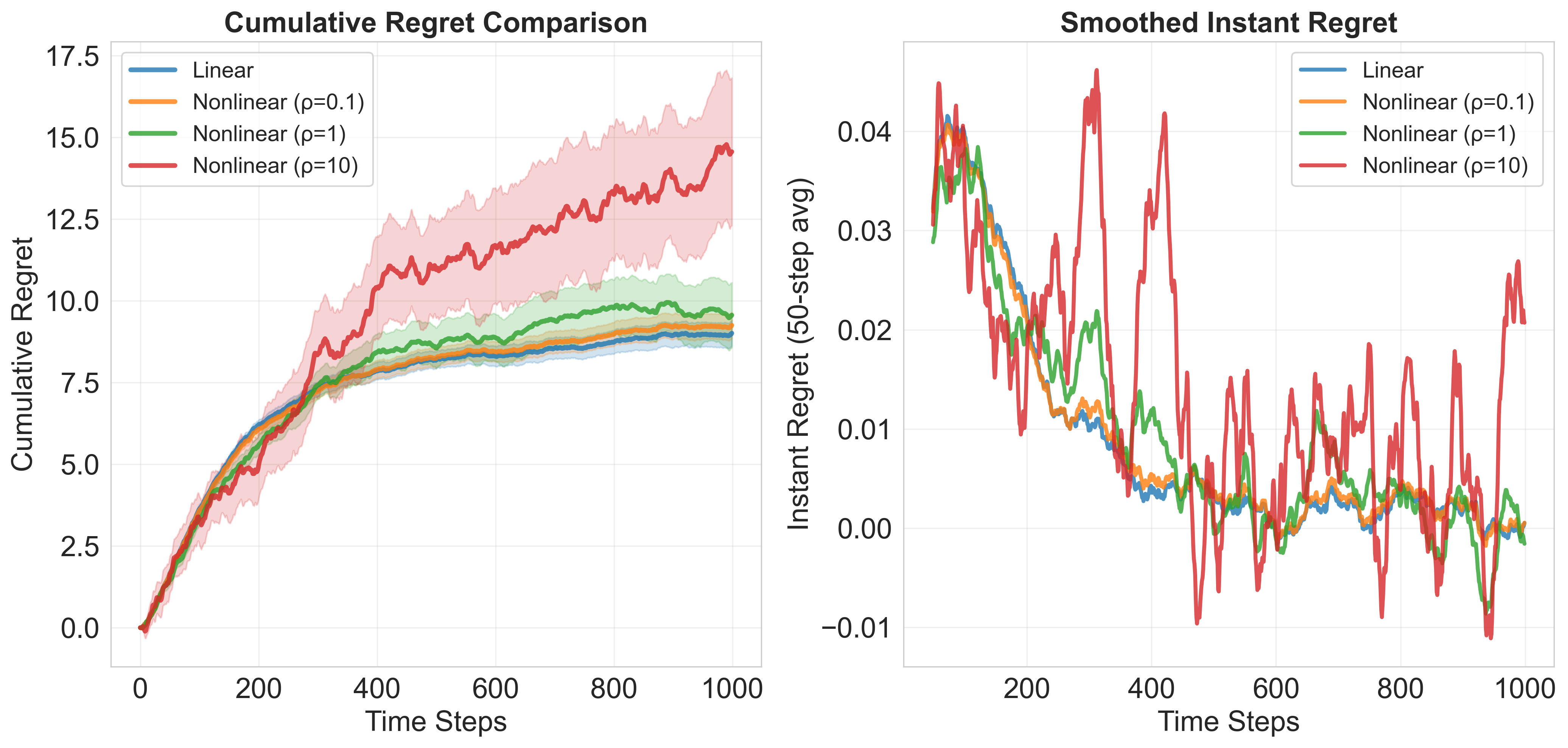}
   \caption{Comparison of PULSE-UCB agent learning results under different linearity settings}
   \label{fig:sim_rho}
\end{figure}

To test robustness, we evaluated \texttt{PULSE-UCB} in a linear environment and three nonlinear variants controlled by a parameter $\rho$. The results in Figure \ref{fig:sim_rho} show a clear correlation between performance and the degree of nonlinearity. The agent performs well in the linear and low-nonlinearity ($\rho=0.1$) cases, with final regrets around 9.3. As the model misspecification becomes more pronounced, the final regret increases to 9.7 for $\rho=1.0$ and further to 14.6 for the highly nonlinear case of $\rho=10.0$. The smoothed instant regret plot (Right) confirms this trend, showing larger and more volatile regret for higher $\rho$. This experiment demonstrates that while \texttt{PULSE-UCB} is robust to smooth deviations from linearity, its performance gracefully deteriorates as the environment becomes more complex.

\subsection{Robustness under Time-Varying Observation Patterns}
\label{app:random_feature_masking}

To empirically substantiate our theoretical claim that PULSE-UCB handles general, time-varying observation patterns, we conducted a Random Feature Masking experiment. In this setup, we adopt the high-dimensional setting described in Appendix~\ref{app:sensitivity_data_scale}, where the dimension of the underlying feature vector is fixed at $d=20$. However, at each time step $t$, the agent only observes a random subset of these features. Specifically, each feature coordinate is independently masked (set to zero) with a probability $p \in \{0.0, 0.2, 0.4, 0.6, 0.8\}$. This simulates a challenging regime where the available information changes dynamically across agents or time. We trained a robust version of the $\beta$-estimator using data augmentation (dropout) and compared PULSE-UCB against all baselines.

Table~\ref{tab:random_masking} summarizes the final cumulative regret under varying masking probabilities.

\begin{table}[htbp]
    \centering
    \caption{Robustness under Random Feature Masking with varying mask probabilities $p$.}
    \label{tab:random_masking}
    \begin{tabular}{r c c c c}
        \toprule
        Mask Prob ($p$) & PULSE-UCB & OFUL-Full & OFUL & CLBBF \\
        \midrule
        0.0 & $46.52 \pm 5.20$   & $2.73 \pm 1.77$ & $346.18 \pm 28.19$ & $144.95 \pm 7.19$ \\
        0.2 & $73.18 \pm 5.90$   & $2.64 \pm 1.76$ & $323.65 \pm 21.64$ & $138.15 \pm 10.06$ \\
        0.4 & $111.29 \pm 7.61$  & $2.04 \pm 2.51$ & $332.07 \pm 21.84$ & $141.86 \pm 12.60$ \\
        0.6 & $169.47 \pm 12.70$ & $3.12 \pm 1.27$ & $335.68 \pm 23.22$ & $139.63 \pm 12.02$ \\
        0.8 & $267.07 \pm 23.99$ & $2.66 \pm 1.39$ & $322.72 \pm 20.34$ & $146.55 \pm 10.72$ \\
        \bottomrule
    \end{tabular}
\end{table}

\textbf{Discussion on Algorithm Behavior and Robustness:}
The results summarized in Table~\ref{tab:random_masking} illustrate distinct behavioral patterns across the algorithms. As expected, OFUL-Full and the naive OFUL maintain their respective near-zero and high-regret baselines, as neither is influenced by history masking. Similarly, CLBBF exhibits stable but mediocre performance across all masking rates, indicating that it fails to leverage partial history to recover the latent context $W_t$. 

In contrast, PULSE-UCB demonstrates remarkable robustness. With full observations, it outperforms CLBBF by a substantial margin. Crucially, this advantage persists even when a significant portion of features is missing ($p=0.4$); it is only under conditions of extreme sparsity (e.g., $p \ge 0.6$) that the information loss causes PULSE-UCB to fall behind the conservative CLBBF baseline.

These results conclusively demonstrate that PULSE-UCB does not rely on a fixed, complete feature structure. As long as the observed subset of features retains predictive power regarding the latent context, our approach effectively bridges the information gap, validating its practical applicability in general regimes with time-varying observations.

\subsection{Robustness to Distributional and Covariate Shifts}
\label{app:covariate_shift}

Identifying the boundaries of our method under distribution mismatches is essential for understanding its practical applicability. In this section, we evaluate the robustness of PULSE-UCB empirically by introducing shifts in the marginal distribution of the observed covariates between the pre-training and online phases, while keeping the latent dynamics and reward mechanism fixed. To thoroughly assess this, we first examine a basic variance shift and then extend our evaluation to more complex, multi-dimensional geometric shifts in the feature space: Rotation, Translation, and Deformation.

\subsubsection{Impact of Covariate Variance Shift}
We first conducted a sensitivity analysis by introducing a \emph{variance shift}. The pre-training data are generated with a standard noise scale $\sigma = 0.1$, while the online environments use increasingly large noise levels $\sigma \in \{0.1, 0.5, 1.0, 1.5, 2.0\}$ (up to a $20\times$ increase in standard deviation), where $\sigma$ denotes the standard deviation of the white noise process $\varepsilon_t$ driving the ARMA generation of the observed context $S_t$. The base high-dimensional setting follows the description in Appendix~\ref{app:sensitivity_data_scale}.

Table~\ref{tab:variance_shift} summarizes the final cumulative regrets under varying noise scales.

\begin{table}[htbp]
    \centering
    \caption{Impact of Covariate Variance Shift (Training $\sigma = 0.1$).}
    \label{tab:variance_shift}
    \begin{tabular}{r c c c c}
        \toprule
        Noise Scale ($\sigma$) & PULSE-UCB & OFUL & CLBBF & OFUL-Full \\
        \midrule
        0.1 & $5.1 \pm 2.3$ & $504.9 \pm 48.0$ & $115.3 \pm 11.0$ & $1.8 \pm 1.7$ \\
        0.5 & $4.5 \pm 1.0$ & $510.5 \pm 84.5$ & $107.2 \pm 15.1$ & $2.9 \pm 2.5$ \\
        1.0 & $5.2 \pm 2.4$ & $519.9 \pm 50.1$ & $107.4 \pm 12.2$ & $2.6 \pm 1.3$ \\
        1.5 & $3.9 \pm 2.3$ & $498.0 \pm 44.2$ & $113.9 \pm 6.5$  & $2.7 \pm 1.9$ \\
        2.0 & $6.0 \pm 1.4$ & $513.6 \pm 29.8$ & $111.7 \pm 14.1$ & $4.0 \pm 1.5$ \\
        \bottomrule
    \end{tabular}
\end{table}

\textbf{Discussion on Variance Shift:} 
Table~\ref{tab:variance_shift} shows that PULSE-UCB remains remarkably stable under covariate variance shifts. Across all noise levels, its final cumulative regret stays in a narrow low-regret range and is consistently close to the oracle OFUL-Full agent. Importantly, despite pre-training only at the low-noise setting ($\sigma = 0.1$), the learned latent structure generalizes to much noisier online environments without noticeable degradation. This indicates that the transition model on the latent state space, rather than the raw covariates, is driving the performance of PULSE-UCB, ensuring robustness to substantial fluctuations in the scale of observed features. 

In contrast, the online-only baselines struggle throughout the entire range. OFUL incurs very large final regrets due to the severe information loss caused by the unobserved context. CLBBF performs better than OFUL but plateaus at a sub-optimal level, showing no capability to adapt to the latent structure. Across all noise scales, PULSE-UCB maintains a massive performance gap---roughly two orders of magnitude relative to OFUL and more than a factor of $20$ relative to CLBBF.

\subsubsection{Rotation: Structural Shift via AR Perturbation}
Varying the scalar noise variance $\sigma^2$ induces a one-dimensional scaling shared across all coordinates of the feature vector $\mathbf{x}_t$. To more thoroughly assess robustness to higher-dimensional covariate shifts, we introduce structural perturbations to the autoregressive (AR) coefficients that generate the scalar state process $S_t$. Since the feature vector $\mathbf{x}_t = (1, S_{t-1}, \ldots, S_{t-19})^\top$ collects 20 lags of this process, the AR parameters fully determine its covariance and correlation structure. Varying these parameters changes the principal axes (eigenvectors) of the 20-dimensional feature distribution, effectively rotating the feature manifold in the context space.

\textbf{Setup:} Historical data are generated under AR coefficients $\phi_{\text{train}}=[0.75, -0.25]$. Online data are generated with coefficients that interpolate between $\phi_{\text{train}}$ and a more oscillatory regime $\phi_{\text{target}} = [-0.80, -0.20]$, controlled by a shift intensity $\alpha$.

\textbf{Geometric Change:} We quantify the structural shift using \emph{PC Subspace Similarity} (the overlap between the top-5 principal components of the historical and online feature distributions). As shown in Table~\ref{tab:rotation}, this similarity drops from $1.000$ to $0.073$, indicating that the feature manifold has rotated to be nearly orthogonal to the training distribution.

\begin{table}[htbp]
    \centering
    \caption{Regret under Structural Shift (Rotation).}
    \label{tab:rotation}
    \begin{tabular}{c c c c c c c}
        \toprule
        $\alpha$ & AR Params & PC Subspace Sim. & PULSE-UCB & OFUL & CLBBF & OFUL-Full \\
        \midrule
        0.0 & $[0.75, -0.25]$  & $1.000$ & $5.24 \pm 0.43$ & $593.0 \pm 15.8$ & $110.5 \pm 2.5$ & $2.68 \pm 0.63$ \\
        0.3 & $[0.28, -0.23]$  & $0.878$ & $5.53 \pm 0.60$ & $418.8 \pm 8.97$ & $118.5 \pm 3.2$ & $1.90 \pm 0.60$ \\
        0.7 & $[-0.33, -0.21]$ & $0.858$ & $6.25 \pm 0.63$ & $367.3 \pm 8.71$ & $103.3 \pm 1.8$ & $1.77 \pm 0.55$ \\
        0.9 & $[-0.65, -0.21]$ & $0.452$ & $6.17 \pm 0.43$ & $249.2 \pm 33.7$ & $104.7 \pm 2.0$ & $1.81 \pm 0.57$ \\
        1.0 & $[-0.80, -0.20]$ & $\mathbf{0.073}$ & $\mathbf{5.94 \pm 0.47}$ & $222.0 \pm 35.9$ & $107.0 \pm 2.0$ & $1.53 \pm 0.61$ \\
        \bottomrule
    \end{tabular}
\end{table}

Notably, PULSE-UCB demonstrates superior robustness, maintaining low regret comparable to the Oracle even under severe structural shifts where standard baselines fail.

\subsubsection{Translation: Sensitivity to Location Shift}
To assess robustness to shifts in the domain's operating point, we introduce a global translation to the feature coordinates.

\textbf{Setup:} We shift the centroid of the state distribution by adding a constant $\mu \in [0, 5.0]$ to every generated state value $S_t$. 

\textbf{Geometric Change:} Since the feature vector $\mathbf{x}_t$ is constructed from lagged states, adding $\mu$ to the scalar process effectively adds $\mu$ to every coordinate of the feature vector (except the fixed intercept). Geometrically, this translates the entire feature cloud along the diagonal direction $\mathbf{v} = (0, 1, \ldots, 1)^\top$ in the multi-dimensional feature space.

\begin{table}[htbp]
    \centering
    \caption{Regret under Mean Shift (Translation).}
    \label{tab:translation}
    \begin{tabular}{c l c c c c}
        \toprule
        $\alpha$ & Shift Detail & PULSE-UCB & OFUL & CLBBF & OFUL-Full \\
        \midrule
        0.0 & Mean +0.0 & $4.22 \pm 0.67$ & $598.4 \pm 16.9$ & $106.1 \pm 3.3$ & $2.83 \pm 0.52$ \\
        0.4 & Mean +2.0 & $10.83 \pm 1.20$ & $10.48 \pm 1.00$ & $154.0 \pm 3.3$ & $10.86 \pm 0.91$ \\
        0.8 & Mean +4.0 & $23.35 \pm 0.88$ & $22.79 \pm 0.66$ & $92.56 \pm 1.3$ & $23.19 \pm 0.65$ \\
        1.0 & Mean +5.0 & $\mathbf{28.92 \pm 0.74}$ & $28.54 \pm 0.66$ & $86.34 \pm 0.8$ & $28.55 \pm 0.62$ \\
        \bottomrule
    \end{tabular}
\end{table}

As shown in Table~\ref{tab:translation}, PULSE-UCB tightly tracks the Oracle (OFUL-Full) performance throughout the entire range. This confirms that our method adapts effectively when the test domain is centered in a different region of the feature space compared to the training domain.

\subsubsection{Deformation: Sensitivity to Geometric Shape}
Finally, we evaluate robustness to changes in the shape of the distribution density, specifically moving from well-behaved Gaussian noise to heavy-tailed distributions.

\textbf{Setup:} We alter the noise process $\varepsilon_t$. The historical data retains standard Gaussian noise, while the online data switches to a heavy-tailed Student-t distribution ($\text{df}=3$). Simultaneously, we introduce a scaling factor that linearly increases the dispersion of this noise up to $3\times$ the original scale.

\textbf{Geometric Change:} This modification fundamentally changes the density geometry of the feature manifold. The transition to heavy tails introduces frequent outliers and extreme values in the multi-dimensional space, creating a spiky distribution shape that violates the sub-Gaussian assumptions typically required by linear bandits.

\begin{table}[htbp]
    \centering
    \caption{Regret under Geometric Shift (Deformation).}
    \label{tab:deformation}
    \begin{tabular}{c l c c c c}
        \toprule
        $\alpha$ & Shift Detail & PULSE-UCB & OFUL & CLBBF & OFUL-Full \\
        \midrule
        0.0 & Gaussian (0.1)     & $5.98 \pm 0.30$ & $591.9 \pm 15.2$ & $107.7 \pm 3.6$ & $2.55 \pm 0.61$ \\
        0.4 & T-dist $\times 1.8$ & $6.17 \pm 0.92$ & $1470.3 \pm 48.7$ & $69.54 \pm 8.8$ & $5.40 \pm 0.58$ \\
        0.8 & T-dist $\times 2.6$ & $4.57 \pm 0.97$ & $2117.7 \pm 104.0$ & $55.16 \pm 4.9$ & $3.60 \pm 1.03$ \\
        1.0 & T-dist $\times 3.0$ & $\mathbf{10.31 \pm 1.91}$ & $2930.3 \pm 169.7$ & $51.44 \pm 5.2$ & $11.11 \pm 1.89$ \\
        \bottomrule
    \end{tabular}
\end{table}

This setting poses a severe challenge. As shown in Table~\ref{tab:deformation}, the standard OFUL algorithm collapses under this geometric deformation, with regret exploding significantly. In stark contrast, PULSE-UCB remains remarkably stable (regret $\approx 10$), demonstrating superior resilience to non-Gaussian geometric distortions. 

\textbf{Conclusion:} Collectively, these experiments confirm that PULSE-UCB is highly robust across three critical dimensions of distributional shift: Rotation (Structure), Translation (Location), and Deformation (Geometry), significantly outperforming baselines in complex, multi-dimensional shift scenarios.

\section{Related Details about Real Dataset Experiments} \label{details_real}
This experiment evaluates the performance of the proposed PULSE-UCB algorithm against several baselines in a realistic setting using the public Taobao User Behavior dataset~\cite{taobao}.

\subsection{Dataset and Preprocessing}
We use the Taobao dataset, which contains user interaction data from Taobao's recommender system. The raw data consists of user profiles (\texttt{user\_profile.csv}), ad features (\texttt{ad\_feature.csv}), and user-ad interaction logs (\texttt{raw\_sample.csv}). Our preprocessing pipeline involves the following steps:
\begin{enumerate}
    \item \textbf{Filtering}: To manage the scale and focus on active user segments and ad categories, we filter the data. We retain only the interactions from users belonging to the top 10 most frequent user segments (\texttt{cms\_segid}) and ads belonging to the top 25 most frequent categories (\texttt{cate\_id}) and brands (\texttt{brand}).
    \item \textbf{Feature Encoding}: Categorical features for both users (e.g., age range, gender) and ads (e.g., category, brand) are converted into high-dimensional, sparse binary vectors using one-hot encoding. The numerical \texttt{price} feature for ads is logarithmically scaled and discretized.
    \item \textbf{Feature Combination}: For each user-ad interaction, the corresponding user feature vector and ad feature vector are concatenated to form a single high-dimensional feature vector.
    \item \textbf{Label Creation}: The \texttt{clk} column in the interaction log (1 for click, 0 for no-click) serves as the ground-truth reward signal for our online bandit simulation. The data is partitioned into two sets based on this label: $X_0$ for non-click events and $X_1$ for click events.
\end{enumerate}

\subsection{Dimensionality Reduction via Autoencoder}
The initial one-hot encoded feature vectors are extremely high-dimensional and sparse. To create a more manageable and dense feature representation, we train an \texttt{Autoencoder} with Batch Normalization.
\begin{itemize}
    \item \textbf{Architecture}: The model consists of an encoder that maps the raw feature dimension $d=83$ to a dense embedding of size $d=32$, and a decoder that reconstructs the original vector from this embedding. 
    \item \textbf{Training}: The autoencoder is trained on the shuffled combination of all available feature vectors ($X_0$ and $X_1$) for 500 epochs with an MSE loss function, a batch size of 10,000, and an Adam optimizer.
    \item \textbf{Output}: After training, we use the encoder to transform all high-dimensional feature vectors into dense 32-dimensional embeddings, which are used in all subsequent steps.
\end{itemize}

\subsection{Partially Observed Setting and Inference Model}
To simulate a realistic scenario where only a subset of features is immediately available, we define a partially observed setting.
\begin{itemize}
    \item \textbf{Feature Split}: Each 32-dimensional feature vector $Y_t$ is split into two halves. The first 16 dimensions, denoted as $S_t$, are considered as the observed features, while the remaining 16 dimensions, $S'_t$, are the unobserved features.
    \item \textbf{Inference Model}: For PULSE-UCB, we pre-train an inference model to predict $S'_t$ from $S_t$. This model is a Multi-Layer Perceptron (MLP) with two hidden layers of 128 neurons each, using ReLU activation functions.
    \item \textbf{Pre-training}: The MLP is trained on a dedicated pre-training set, which constitutes 20\% of the total shuffled data. The model is trained for 100 epochs using an MSE loss function and an Adam optimizer to minimize the reconstruction error of $S'_t$. The remaining 80\% of the data is reserved for the online evaluation phase.
\end{itemize}

\subsection{Online Evaluation Protocol}
The online simulation is performed on the held-out 80\% of the dataset. 
\begin{enumerate}
    \item The simulation runs for $T$ time steps, where $T$ is the size of the online dataset minus $K$.
    \item At each time step $t$, a set of $K=20$ candidate arms (ads) is randomly sampled without replacement from the online dataset.
    \item Each bandit agent selects one arm from the $K$ candidates based on its internal policy.
    \item The agent observes the reward (click or no-click) associated with the chosen arm.
    \item The agent updates its internal parameters using the feature vector of the chosen arm and the observed reward.
    \item This process is repeated over independent runs with different random seeds to ensure robust results, and the average cumulative click-through rate (CTR) is reported.
\end{enumerate}

\end{document}